\documentclass[]{fairmeta}
\setcitestyle{authoryear,round}

\usepackage{adjustbox}
\usepackage{colortbl}
\usepackage[shortlabels]{enumitem}
\usepackage{float}
\usepackage{pifont}
\usepackage{tabularx}
\usepackage{xspace}
\usepackage{xcolor}
\let\cite\citep
\makeatletter
\renewcommand\NAT@biblabel[1]{\NAT@biblabelnum{#1}}
\makeatother
\renewcommand\authorformat[2][]{%
  {\sffamily \bfseries #2%
   \if\relax\detokenize{#1}\relax\else$^{#1}$\fi}}
\renewcommand\affiliationformat[2][]{%
  {\normalsize
   \if\relax\detokenize{#1}\relax\else$^{#1}$\fi
   #2}}
\definecolor{affone}{RGB}{0,102,204}       
\definecolor{afftwo}{RGB}{128,0,128}       
\definecolor{affcorr}{RGB}{0,100,0}        
\newcommand{\cmark}{\textcolor{green!60!black}{\ding{51}}}
\newcommand{\xmark}{\textcolor{red!70!black}{\ding{55}}}

\newcommand{\ourmethod}{HippoCamp\xspace}
\newcommand{\hippologo}{%
  \raisebox{-0.2ex}{\includegraphics[height=1.1em]{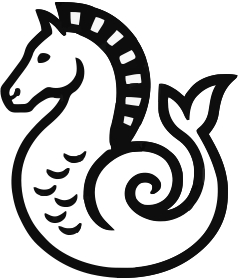}}%
}

\title{\texorpdfstring{\hippologo\ HippoCamp: Benchmarking Contextual Agents on Personal Computers}{HippoCamp: Benchmarking Contextual Agents on Personal Computers}}

\author[\textcolor{affone}{1}]{Zhe Yang}
\author[\textcolor{affone}{1}]{Shulin Tian}
\author[\textcolor{affone}{1}\textcolor{black}{,}\textcolor{afftwo}{2}]{Kairui Hu}
\author[\textcolor{affone}{1}\textcolor{black}{,}\textcolor{afftwo}{2}]{Shuai Liu}
\author[\textcolor{affone}{1}]{Hoang-Nhat Nguyen}
\author[\textcolor{affone}{1}]{Yichi Zhang}
\author[\textcolor{affone}{1}]{Zujin Guo}
\author[\textcolor{affone}{1}]{Mengying Yu}
\author[\textcolor{affone}{1}]{Zinan Zhang}
\author[\textcolor{affone}{1}]{Jingkang Yang}
\author[\textcolor{affone}{1}\textcolor{black}{,}\textcolor{afftwo}{2}\textcolor{black}{,}\textcolor{affcorr}{\dagger}]{Chen Change Loy}
\author[\textcolor{affone}{1}\textcolor{black}{,}\textcolor{affcorr}{\dagger}]{Ziwei Liu}

\affiliation[\textcolor{affone}{1}]{\textcolor{affone}{S-Lab, Nanyang Technological University, Singapore}}
\affiliation[\textcolor{afftwo}{2}]{\textcolor{afftwo}{Synvo AI}}
\contribution[\textcolor{affcorr}{\dagger}]{\textcolor{affcorr}{Corresponding authors}}

\date{April 1, 2026}
\project{\href{https://hippocamp-ai.github.io}{\texttt{https://hippocamp-ai.github.io}}}
\metadata[Data Visualization]{\href{https://hippocamp-ai.github.io/hippocamp/}{\texttt{https://hippocamp-ai.github.io/hippocamp/}}}

\abstract{
We present \ourmethod{}, a new benchmark designed to evaluate agents' capabilities on multimodal file management. Unlike existing agent benchmarks that focus on tasks like web interaction, tool-use, or software automation in generic settings, \ourmethod{} evaluates agents in user-centric environments to model individual user profiles and search from massive personal files for context-aware reasoning. Our benchmark instantiates device-scale file systems over real-world profiles spanning diverse modalities, comprising 42.4 GB of data across over 2K real-world files. Building upon the raw files, we construct 581 QA pairs to assess agents' capabilities in search, evidence perception, and multi-step reasoning. To facilitate fine-grained analysis, we provide 46.1K densely annotated structured trajectories for step-wise failure diagnosis. We evaluate a wide range of state-of-the-art multimodal large language models (MLLMs) and agentic methods on \ourmethod{}. Our comprehensive experiments reveal a significant performance gap: even the most advanced commercial models achieve merely a 48.3\% accuracy in user profiling, struggling particularly with long-horizon retrieval and cross-modal reasoning within dense personal file systems. Furthermore, our step-wise failure diagnosis identifies multimodal perception and evidence grounding as the primary bottlenecks. Ultimately, \ourmethod{} exposes the critical limitations of current agents in realistic, user-centric environments and provides a robust foundation for developing next-generation personal AI assistants.

}

\begin{document}

\maketitle

\noindent
\begin{tcolorbox}[
    colback=blue!3,
    colframe=blue!60!black,
    arc=4pt,
    boxrule=0.5pt,
    boxsep=2pt,
    left=5pt,
    right=5pt,
    top=3pt,
    bottom=2pt,
    enhanced jigsaw,
    drop shadow=gray!40!white,
]
{\noindent\sffamily\bfseries HippoCamp at a Glance}\\[-2pt]
{\color{blue!30!black}\hrule height 0.3pt}\vspace{4pt}
\small
\begin{enumerate}[leftmargin=1.5em, itemsep=4pt, parsep=0pt, topsep=0pt, label={\small\textbf{S\arabic*.}}]
    \item \textbf{Model realistic personal file systems:} HippoCamp instantiates three archetypal, long-lived personal computing environments with deep folder hierarchies, heterogeneous long-tail file types, and cross-modal evidence distributed across more than 2K files.\unskip\nobreak\hfill\mbox{\footnotesize\textit{See~\S\ref{sec:method}}}
    \item \textbf{Benchmark two complementary personalized-memory tasks:} HippoCamp evaluates \textit{factual retention} and \textit{profiling}, both of which require tightly coupled search, multimodal perception, and reasoning over user-specific evidence.\unskip\nobreak\hfill\mbox{\footnotesize\textit{See~\S\ref{ssec:tasks}~and~\S\ref{sec:experiment}}}
    \item \textbf{Enable interpretable failure diagnosis:} HippoCamp pairs 581 QA tasks with 46.1K structured annotations, allowing errors to be localized to search, perception, and reasoning stages.\unskip\nobreak\hfill\mbox{\footnotesize\textit{See~\S\ref{subsec:capability_analysis}~and~\S\ref{subsec:failure_modes}}}
    \item \textbf{Expose a large gap to robust personal-file agents:} even the strongest evaluated system reaches only 48.3\% profiling accuracy overall, with persistent failures in entity disambiguation, multimodal grounding, and iterative evidence synthesis.\unskip\nobreak\hfill\mbox{\footnotesize\textit{See~\S\ref{sec:experiment}~and~\S\ref{subsec:agent_design}}}
\end{enumerate}
\end{tcolorbox}

\clearpage
\section{Introduction}
\label{sec:intro}

\begin{figure}[ht]
  \centering
  \includegraphics[width=\textwidth]{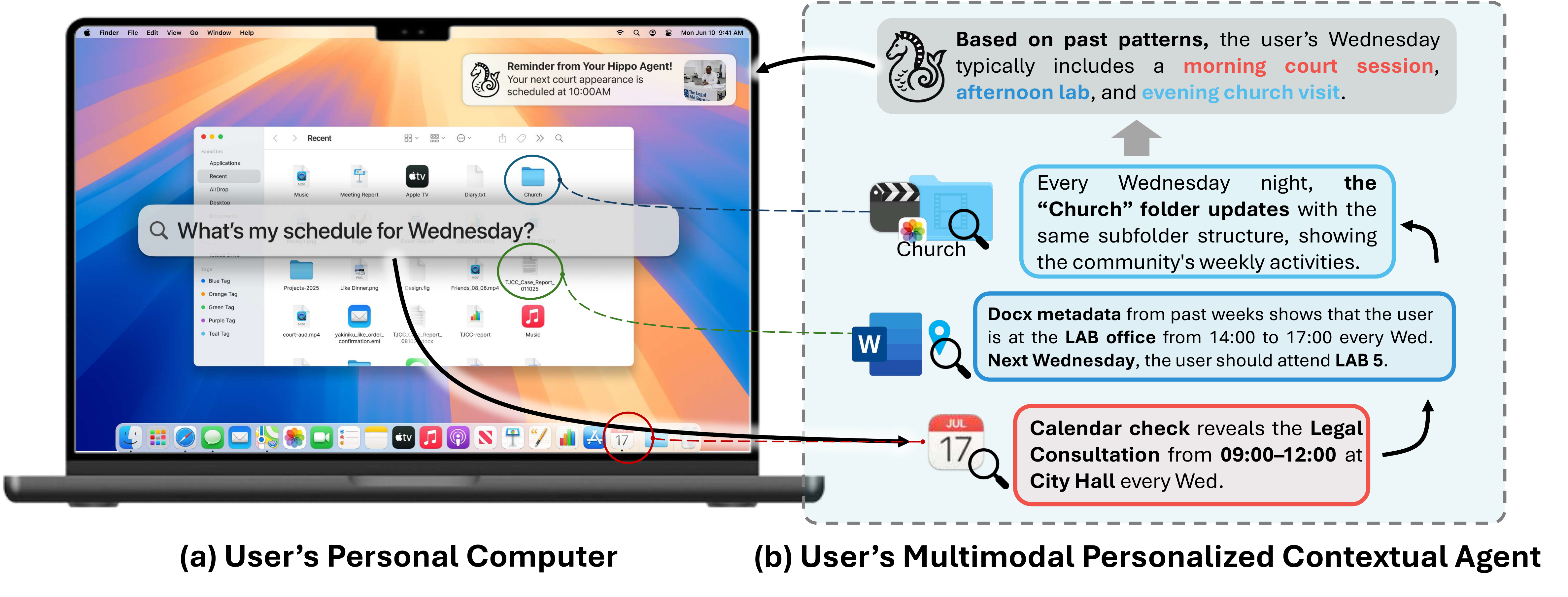}
  \captionsetup{skip=2pt}
  \caption{\textbf{Overview of Task in HippoCamp Benchmark.} HippoCamp is a benchmark designed to evaluate agents' ability to search, perceive, and reason over long-term, realistic, large-scale personal file systems. It reproduces multimodal digital environments containing tens of gigabytes of user-specific assets across diverse modalities. Through tasks that demand grounded retrieval, cross-file pattern inference, and long-horizon reasoning, the benchmark rigorously evaluates the personalized multimodal memory of contextual agents.}
  \label{fig:overview}
\end{figure}

The rise of large multimodal models (LMMs) and agentic systems has accelerated progress toward intelligent assistants capable of operating across perception, reasoning, and action~\cite{xi2023risepotentiallargelanguage, sumers2024cognitivearchitectureslanguageagents, Yao2025SurveyAgenticMLLMs}. A natural and impactful application of such systems lies in \textbf{personalized digital environments}~\cite{Wang2024PersonalizedAgents}, where agents manage, retrieve, and reason over users' heterogeneous digital assets~\cite{Li2024PersonalLLMAgents}.
Unlike web-based or task-specific agents~\cite{Wei2025BrowseComp, Mialon2023GAIA, Yao2023WebShop}, personalized agents operating in local digital environments must navigate continuously evolving devices composed of multimodal files, diverse formats, and long-term behavioral traces~\cite{Vianna2019SearchingHeterogeneous, wu2024oscopilotgeneralistcomputeragents}.
Reasoning in such contexts requires integrating perception across modalities with retrieval and reflection over past interactions, paralleling the role of the human hippocampus in contextual memory consolidation and recall.

However, despite recent progress in long-context reasoning and multimodal retrieval~\cite{Liu2025LongContextSurvey}, there remains no standardized benchmark for evaluating an agent's ability to \textbf{understand, recall, and reason} over massive, personalized, multimodal file systems. While recent efforts address document-level multimodal retrieval~\cite{Dong2025MMDocIR}
or personalized tool-use planning~\cite{xiu2026astrabenchevaluatingtooluseagent},
none target the scale and heterogeneity of personal computing environments. Existing evaluations predominantly target general domains such as web automation~\cite{Gur2024WebAgent, Song2025BearCUBS, Koh2024VisualWebArena}, code generation~\cite{Yu2024CoderEval}, document understanding~\cite{Ouyang2025OmniDocBench}, or embodied planning~\cite{Sadhu2025VestaBench}, while neglecting the rich information encoded in personalized file systems. Although these benchmarks have advanced perception and tool-use capabilities, they operate in isolated, goal-specific scenarios detached from the user's personal context. Consequently, they fail to capture the long-term continuity, complex cross-referencing and validation across interrelated multimodal files over time, and the personalized reasoning required for realistic personal computing.

To fill this gap, we introduce \textbf{\ourmethod{}}, a benchmark for evaluating memory-augmented agents in realistic personal computing environments, as illustrated in \cref{fig:overview}. HippoCamp constructs \emph{three} representative personal computing environments derived from real-world user data. Rather than a literal three-way categorization, each environment is an \emph{archetypal} instantiation of a high-dimensional profile space. Together, they expose device-scale file-system challenges such as deep hierarchical organization, heterogeneous long-tail file formats, and cross-modal evidence dependencies. The benchmark encompasses two representative task categories that mirror authentic computer usage: \textit{factual retention} and \textit{profiling}. Each task demands agentic behaviors integrating \textbf{search, perception, and reasoning}, enabling systematic evaluation of personalized multimodal intelligence. In summary, HippoCamp makes three main contributions:

\begin{enumerate}
\item \textbf{Realistic personal computing environments.} We faithfully simulate user-level digital ecosystems by constructing three distinct, file-intensive profiles. This design captures long-term continuity, idiosyncratic folder structures, and the interconnected nature of real-world personal file systems.

\item \textbf{Device-scale corpus with dense supervision.} We provide a massive dataset comprising over 2000 heterogeneous files (totaling 42.4 GB), accompanied by 581 user-need-driven, evidence-grounded queries. To support a rigorous evaluation of agent capabilities across varying depths and perspectives, we include 46.1K fine-grained annotations at multiple levels of granularity.
\item \textbf{Comprehensive agent capability evaluation.} We formulate a rigorous question-answering benchmark built on two core task categories: \textit{factual retention}: retrieving specific information, and \textit{profiling}: inferring user preferences. These tasks go beyond simple retrieval, requiring agents to execute multi-step behaviors that seamlessly integrate file-system search, multimodal evidence perception, and personalized reasoning across interrelated files over time.

\end{enumerate}

\section{Related Work}
\label{sec:related}

\begin{table}[ht]
\centering
\caption[Comparison of Benchmarks Related to Multimodal Context and Agentic Reasoning.]{
Comparison of Benchmarks Related to Multimodal Context and Agentic Reasoning.
For modalities,
\includegraphics[height=0.9em]{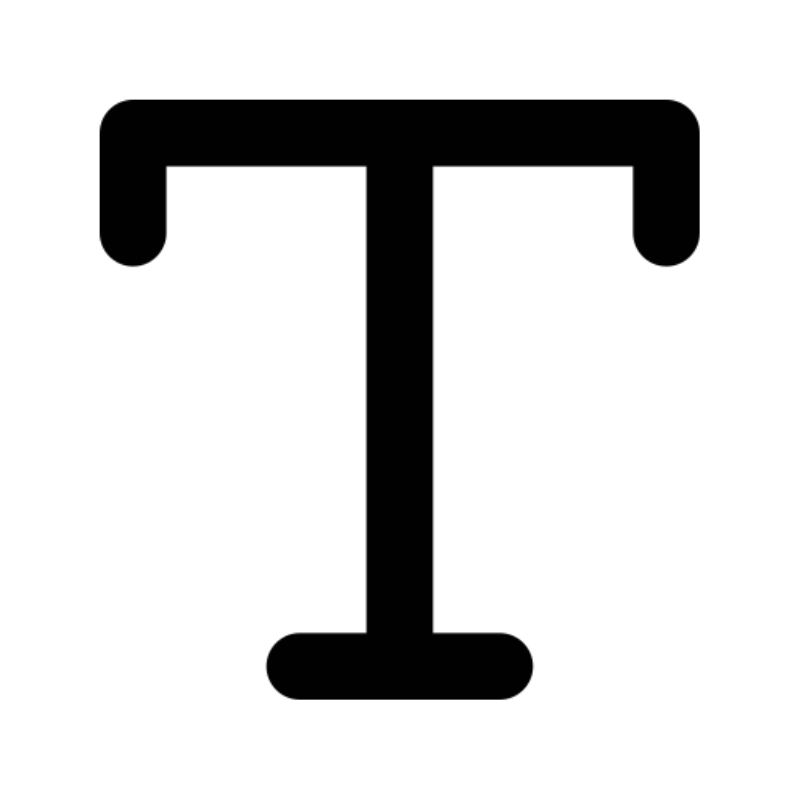} denotes text,
\includegraphics[height=0.9em]{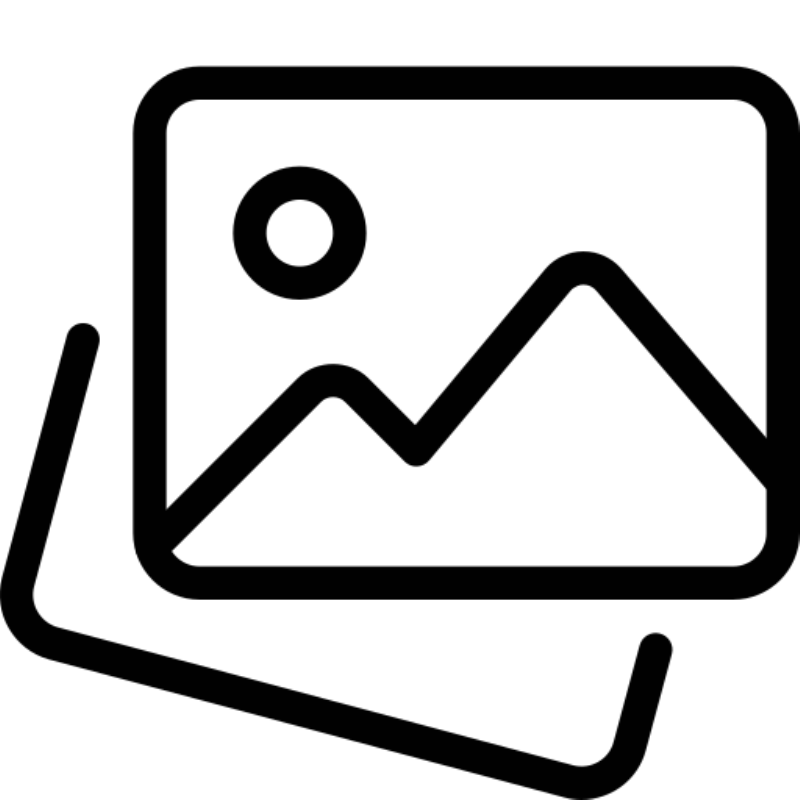} denotes images,
\includegraphics[height=0.9em]{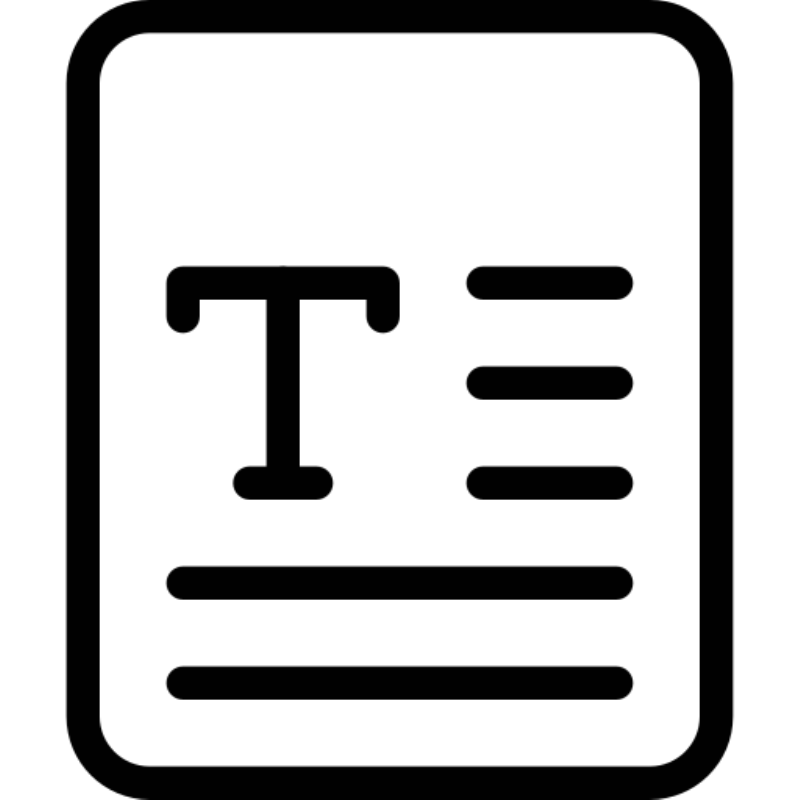} denotes documents,
\includegraphics[height=0.9em]{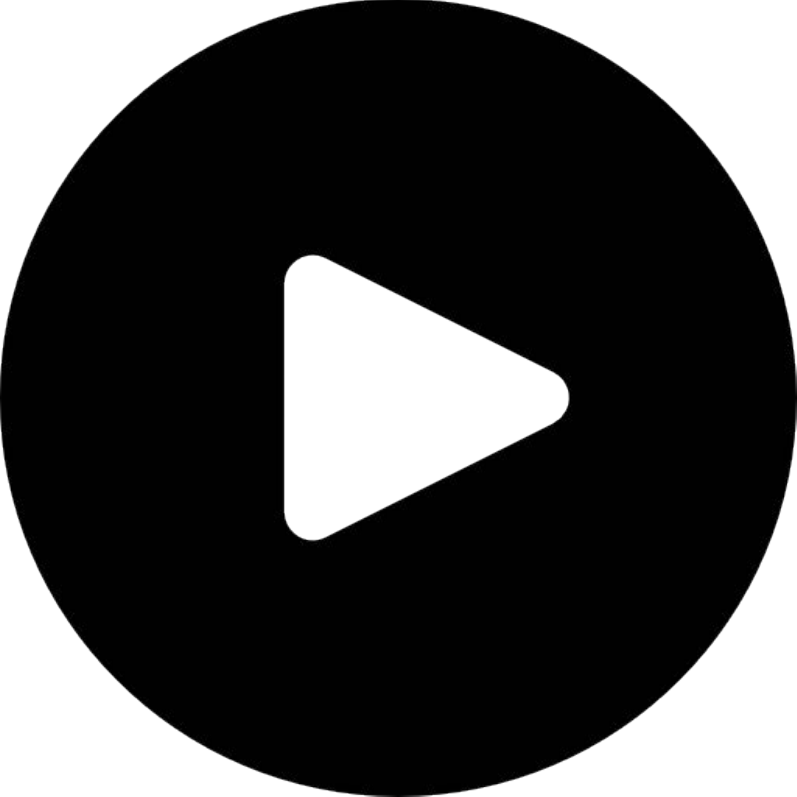} denotes videos, and
\includegraphics[height=0.9em]{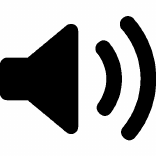} denotes audio.
HippoCamp introduces a personalized multimodal file-system environment that spans all these modalities, enabling cross-file and cross-modal reasoning tasks at real-world scale.}
\label{tab:compare_benchmark}
\scriptsize
\setlength\tabcolsep{3pt}
\renewcommand{\arraystretch}{0.95}
\resizebox{\linewidth}{!}{
\begin{tabular}{@{}lllccccc@{}}
\toprule[1pt]
\textbf{Benchmark}
& \textbf{Source}
& \textbf{Modalities}
& \textbf{\#Samples}
& \textbf{Multimodal}
& \textbf{User Profile}
& \textbf{File-System} \\
\midrule

HotpotQA~\cite{Yang2018HotpotQA} / KILT~\cite{Petroni2021KILT}
& Wikipedia
& \includegraphics[height=0.9em]{figs/icons/text.pdf}
& ${\sim}$100k
& \xmark
& \xmark & \xmark \\

BrowseComp~\cite{Wei2025BrowseComp}
& Web
& \includegraphics[height=0.9em]{figs/icons/text.pdf}
& 1,266
& \xmark
& \xmark & \xmark \\

MetaTool~\cite{Huang2024MetaTool} / MINT~\cite{Wang2024MINT}
& Web + Tools
& \includegraphics[height=0.9em]{figs/icons/text.pdf}\;\text{\scriptsize (w/ code)}
& ${\sim}$3k--20k
& \xmark
& \xmark & \xmark \\

WebQA~\cite{Chang2022WebQA}
& Web
& \includegraphics[height=0.9em]{figs/icons/text.pdf}
  \includegraphics[height=0.9em]{figs/icons/image.pdf}
& 7.5k
& \cmark
& \xmark & \xmark \\

GAIA~\cite{Mialon2023GAIA} / WebShop~\cite{Yao2023WebShop} / PaperBench~\cite{Starace2025PaperBench}
& Web
& \includegraphics[height=0.9em]{figs/icons/text.pdf}
  \includegraphics[height=0.9em]{figs/icons/image.pdf}
& ${\sim}$1k--10k
& \cmark
& \xmark & \xmark \\

MultiModalQA~\cite{Talmor2021MultiModalQA}
& Web
& \includegraphics[height=0.9em]{figs/icons/text.pdf}
  \includegraphics[height=0.9em]{figs/icons/image.pdf}
  \includegraphics[height=0.9em]{figs/icons/document.pdf}\;\text{\scriptsize (Table-Only)}
& 29k
& \cmark
& \xmark & \xmark \\

MMDocRAG~\cite{Dong2025RAGDocQA} / M3DocRAG~\cite{Cho2024M3DocRAG}
& Documents
& \includegraphics[height=0.9em]{figs/icons/text.pdf}
  \includegraphics[height=0.9em]{figs/icons/image.pdf}
  \includegraphics[height=0.9em]{figs/icons/document.pdf}
& ${\sim}$4k--10k
& \cmark
& \xmark & \xmark \\

LoCoMo~\cite{Maharana2024LongTermMemory}\textsuperscript{\raisebox{-0.2ex}{$\dagger$}}
& Personal Lifelog
& \includegraphics[height=0.9em]{figs/icons/text.pdf}
& 300
& \xmark
& \cmark & \xmark \\
EgoLifeQA~\cite{Yang2025EgoLife} / Ego-R1-Bench~\cite{Tian2025EgoR1}\textsuperscript{\raisebox{-0.2ex}{$\ddagger$}}
& Personal Lifelog
& \includegraphics[height=0.9em]{figs/icons/text.pdf}
  \includegraphics[height=0.9em]{figs/icons/video.pdf}
  \includegraphics[height=0.9em]{figs/icons/audio.pdf}
& ${\sim}$5k
& \cmark
& \cmark & \xmark \\

\midrule
\textbf{HippoCamp (Ours)}
& \textbf{Personal File Systems}
& \includegraphics[height=0.9em]{figs/icons/text.pdf}
  \includegraphics[height=0.9em]{figs/icons/image.pdf}
  \includegraphics[height=0.9em]{figs/icons/document.pdf}
  \includegraphics[height=0.9em]{figs/icons/video.pdf}
  \includegraphics[height=0.9em]{figs/icons/audio.pdf}
& \textbf{581 QAs}
& \textbf{\cmark}
& \textbf{\cmark} & \textbf{\cmark} \\

\bottomrule[1pt]
\end{tabular}}

{\setlength{\parindent}{0pt}\raggedright
\par\raisebox{-0ex}{$\dagger$} LoCoMo contains 300 text-only personal questions;
\par\raisebox{-0ex}{$\ddagger$} EgoLifeQA/Ego-R1-Bench provides approximately 20\% personalized samples, and is limited to videos/audio single-modal context.\par}
\end{table}

\noindent\textbf{Benchmarks for Multimodal Contextual Agents.} Contextual retrieval benchmarks span from text-centric datasets~\cite{Yang2018HotpotQA,Petroni2021KILT,Tang2024MultiHopRAG,Friel2025RAGBench} to multimodal and agentic settings driven by richer evidence needs. MultimodalQA~\cite{Talmor2021MultiModalQA} supports cross-modal grounding over text, images, and tables, while WebQA~\cite{Chang2022WebQA} studies retrieval from distractor-containing candidate pools. Document benchmarks such as M3DocRAG~\cite{Cho2024M3DocRAG} and MMDocRAG~\cite{Dong2025RAGDocQA} emphasize fine-grained selection under noise. Agentic benchmarks move from static corpora to interactive environments: WebShop~\cite{Yao2023WebShop} and PaperBench~\cite{Starace2025PaperBench} evaluate goal-driven action sequences, and MetaTool~\cite{Huang2024MetaTool}, MINT~\cite{Wang2024MINT}, InfoDeepSeek~\cite{Xi2025InfoDeepSeek}, and BrowseComp~\cite{Wei2025BrowseComp} probe tool use, multi-turn reasoning, and web exploration. However, these benchmarks largely assume public data and fully observable states, and thus do not evaluate long-lived personalized context or heterogeneous multimodal evidence distributed across user devices. Real personal environments contain identity cues, behavioral histories, and longitudinal records across diverse file types---signals largely absent from current evaluations.

\noindent\textbf{Agentic Systems with Memory and Personalization.} Agentic systems perform multi-step workflows in interactive environments and thus require context integration; memory is central to long-horizon coherence. Prior work explores (i) \emph{trajectory-based} memory, e.g., internalizing experience via finetuning or reinforcement learning~\cite{Zhang2024AgentOhana,Fu2025AgentRefine}; (ii) \emph{retrieval-based} memory, e.g., storing and retrieving past episodes or tool traces~\cite{Zhao2024ExpeL,Luo2025AgentAuditor}; and (iii) \emph{skill distillation}, compressing reusable skills into inference-time modules~\cite{Wang2025MobileAgentE,Zheng2025SkillWeaver}. Other lines build structured episodic/semantic memories to organize interaction histories and improve multi-step reliability~\cite{Zhang2025GMemory,Yang2025EgoLife}. Building on these mechanisms, personalization operationalizes memory at the user level: PersonaAgent~\cite{Zhang2025PersonaAgent} maintains user-specific episodic/semantic records, while recent systems learn user knowledge graphs, preference embeddings, or long-term histories to adapt without parameter updates~\cite{Wang2024PersonalizedAgents,Lee2025MAP}. Complementarily, Telemem~\cite{Chen2026TeleMem} consolidates user-grounded interactions into narrative and multimodal episodic memories, enabling personalized retrieval over dialogue and visual experience. However, existing evaluations are small-scale or synthetic and typically restricted to narrow modalities (e.g., text/webpages), failing to reflect the heterogeneous and evolving context of personal computing where evidence spans all five modalities, motivating benchmarks that match the scale and diversity of user file ecosystems.

\noindent\textbf{Multimodal Contextual Retrieval.} Retrieval-augmented generation (RAG) uses retrieval as external memory to surface evidence beyond the context window, while reasoning-centric models such as DeepSeek-R1~\cite{Guo2025DeepSeekR1} refine intermediate traces and invoke tools during multi-step inference. Multimodal RAG generalizes text retrieval to heterogeneous evidence: document systems~\cite{Cho2024M3DocRAG,Dong2025MMDocIR} encode layout and mixed text--image content for fine-grained grounding; image--text frameworks~\cite{Zhan2025MMRAG,Guo2025RAGAnything} fuse visual and textual features for cross-modal alignment; and video methods such as VideoRAG~\cite{Ren2025VideoRAG} index temporal clips for long-range visual retrieval. Despite broader coverage, top-$k$ retrieval remains brittle when cues are dispersed across many files or long time spans. Reasoning models partially mitigate this by generating intermediate traces to coordinate search: Search-R1~\cite{Jin2025SearchR1} and MMSearch-R1~\cite{Wu2025MMSearchR1} iteratively refine queries, and Ego-R1~\cite{Tian2025EgoR1} extends this paradigm to egocentric streams. However, both RAG and reasoning models primarily assume public or task-bounded retrieval spaces, rather than the personalized, long-lived multimodal context of real user ecosystems. Our benchmark targets this setting directly (see \cref{tab:compare_benchmark}), evaluating models where relevant signals accumulate across all modalities, and where both retrieval and reasoning must operate within authentic, user-specific digital environments.

\section{The HippoCamp Benchmark}
\label{sec:method}

\begin{figure}[ht]
  \centering
  \includegraphics[width=\textwidth]{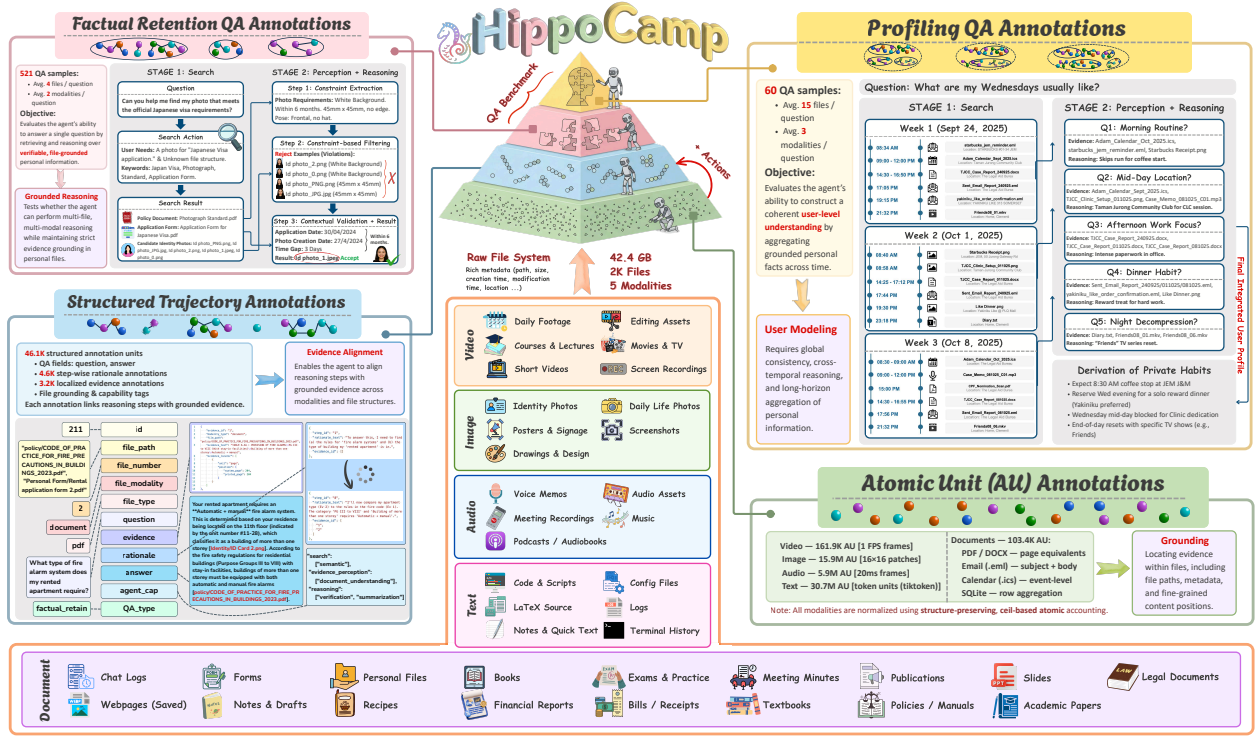}
  \caption{\textbf{\ourmethod{} hierarchical annotation schema.} The pyramid organizes benchmark supervision from low-level atomic grounding and action traces to structured trajectories and task-level QA, with increasing \textbf{abstraction and aggregation} toward user-level memory. It highlights the benchmark's multi-granularity diagnostic structure: \textit{factual retention} is grounded in localized evidence and intermediate reasoning steps, while \textit{profiling} sits at the top, requiring long-horizon, cross-modal synthesis of multiple grounded facts into coherent user-level inference.}
  \label{fig:hippocamp_overview}
\end{figure}

\ourmethod{} is a benchmark for evaluating memory-augmented agents in realistic, device-resident personal file systems. It models personal computing as a multimodal, long-tail information space in which everyday artifacts and user-centric records are organized by file-system structure and temporal metadata. Within this setting, we formulate personalized file understanding as open-ended, evidence-grounded question answering. Rather than serving as a dataset overview, \cref{fig:hippocamp_overview} specifies the benchmark's \emph{supervision hierarchy}: it organizes annotation from low-level, localized evidence and action traces to structured trajectories and task-level evaluation, with increasing abstraction toward user-level memory. This hierarchy clarifies how HippoCamp supports diagnosis at multiple granularities, from whether an agent can localize evidence in individual files to whether it can compose grounded intermediate steps into coherent answers over long-horizon contexts. Within this structure, we define two task types with increasing aggregation demands. \textbf{Factual retention} requires retrieving and reasoning over verifiable file-grounded facts, whereas \textbf{profiling} sits at the top of the hierarchy, requiring agents to synthesize multiple grounded facts across time into coherent user-level inferences such as preferences, behavioral patterns, scheduling information, retrospective reflections, and workflows. Solving both tasks depends on three coupled capabilities: \textbf{search} to locate relevant files in a large heterogeneous system, \textbf{perception} to interpret multimodal contents, and \textbf{reasoning} to integrate cross-file, cross-temporal evidence into accurate, context-aware answers.

\subsection{Dataset Construction}
\label{ssec:dataset_construct}
\begin{figure}[ht]
  \centering
  \includegraphics[width=\linewidth]{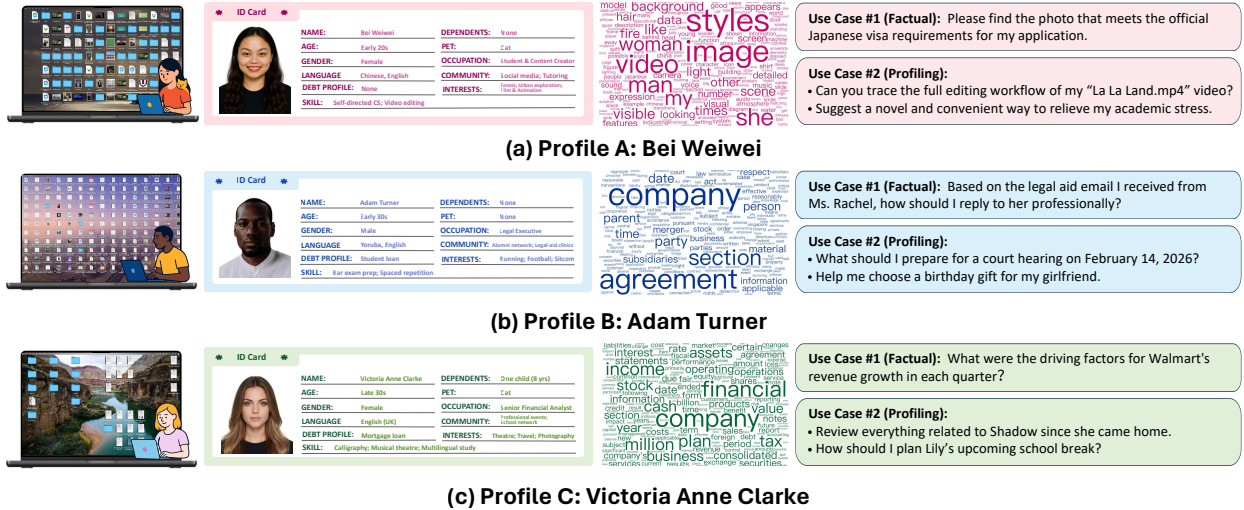}
  \caption{\textbf{Archetypal user profiles in HippoCamp.} Each profile instantiates a distinct personal-device environment characterized along multiple dimensions, and is paired with a representative \textit{word-cloud summary of text-converted file contents}, together with \textit{factual retention} and \textit{profiling} QA examples for (a) Bei, (b) Adam, and (c) Victoria.}
  \label{fig:profiles}
\end{figure}

\noindent\textbf{Source.}

\ourmethod{} is derived from interviews with 100+ participants sampled to reflect general personal-computing settings. During recruitment, we apply strict, multi-stage source selection, retaining only candidates whose file systems exhibit stable behavioral regularities and \emph{evidence-complete} long-horizon personal traces that support auditable user-level inference via cross-file references. We then aggregate the selected participants' files into coherent archetypal profiles by matching file-type/modality distributions and high-level organizational patterns while spanning diverse demographic, socio-economic, professional, and lifestyle dimensions. These aggregated collections are then condensed into three distinct and representative profiles, each assigned semantic attributes such as name and age. The resulting profiles--(a) Bei Weiwei, (b) Adam Turner and (c) Victoria Anne Clarke, each highlight different facets of personal computing: Profile A represents a student and content-creator context, Profile B a legal-executive environment, and Profile C a senior-financial-analyst setting (see~\cref{fig:profiles} for detailed specifications). To preserve coherence, we screen for temporal and semantic conflicts and apply minimal edits only when unavoidable; we further remove system-generated, non-user artifacts and anonymize sensitive identifiers using consistent pseudonyms, producing an unindexed ``haystack'' file system suitable for stress-testing personalized multimodal agents.

\begin{figure}[ht]
  \centering
  \includegraphics[width=\linewidth]{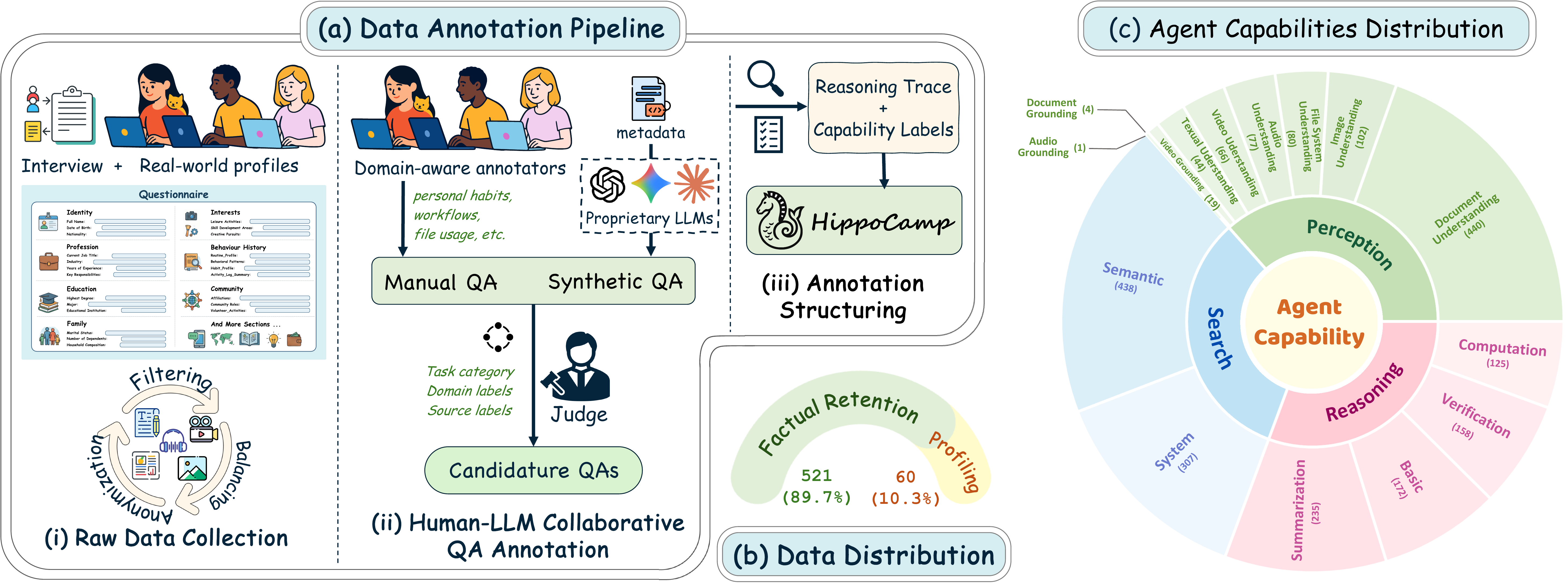}
  \caption{\textbf{Benchmark overview.} (a) Data collection and human--LLM collaborative annotation pipeline for grounded trajectory construction. (b) Distribution of \textit{factual retention} and \textit{profiling} tasks. (c) Distribution of annotated agent capability labels over \textit{search}, \textit{perception}, and \textit{reasoning}.}
  \label{fig:bench_stat}
\end{figure}

\noindent\textbf{Annotation.}

As shown in \cref{fig:bench_stat}(a), we annotate QA with a hybrid pipeline that combines \textbf{expert-driven manual authoring} and \textbf{LLM-assisted synthesis}. Domain-aware annotators, drawn from the contributor groups underlying each archetypal profile, create \emph{manual} questions that are explicitly user-driven, grounded in their own files and routines and reflective of authentic personal-computing needs. In parallel, proprietary LLMs~\cite{Comanici2025Gemini25, anthropic2025claude-sonnet-4-5, openai2025gpt5} generate \emph{synthetic} candidates conditioned on contextual metadata (file paths, timestamps, and directory hierarchies) to improve coverage and balance modality and evidence distributions. All candidates are then consolidated by human annotators, who rigorously review, edit, and filter them into a final \emph{curated} question set that is meaningful, factually correct, and faithfully grounded in the personal context. We further enforce diversity through intent-level deduplication and pattern-level de-duplication, while balancing modality combinations and evidence-set sizes across the collection. For each retained question, annotators construct intermediate supervision by structuring it into a grounded trajectory, including a step-wise reasoning trace, explicit file-grounded evidence, and agent capability labels.

\subsection{Tasks}
\label{ssec:tasks}

Each QA pair is annotated as a structured trajectory stored as a JSON record, including the question and ground-truth answer, a step-wise rationale, and fine-grained localized evidence with atomic pointers into file content (e.g., page indices, table cells, or textual spans), together with file-level grounding metadata. Each trajectory is further labeled with \textbf{agent capability} tags (\cref{fig:bench_stat}(c)) that decompose the required behaviors into three stages: \textbf{search} (system-level navigation, semantic retrieval), \textbf{perception} (file-system understanding, modality-specific understanding and grounding for text, documents, images, videos, and audio), and \textbf{reasoning} (basic inference, computation, summarization, and verification). All trajectories are authored by human annotators following a minimalist annotation principle that explicitly links these stages, enabling interpretable diagnosis of multimodal agent behavior. This schema supports fine-grained capability analysis, evidence-level benchmarking, and evaluation of long-horizon, multi-step reasoning in realistic personalized file ecosystems.
\noindent\textbf{Profiling.}
The profiling tasks evaluate whether an agent can construct a coherent user-level model from device-resident files by aggregating grounded personal facts across time. Each query targets high-level persona attributes, including preferences and routines, scheduling constraints, retrospective accounts, and workflow patterns spanning life, work, and study. Unlike single-fact queries, profiling requires synthesizing heterogeneous cues from multimodal content, file-system organization, and temporal regularities, and producing a globally consistent, actionable response. For example, ``For the afternoon on October 27, 2025, schedule a good plan for me.'' tests whether the agent can integrate evidence from relevant files (e.g., calendars and communications) with historical routines and stated preferences to produce a feasible, personalized plan consistent with existing commitments.
\noindent\textbf{Factual retention.}
The factual retention tasks focus on evaluating an agent's capability to retrieve, comprehend, and reason over factual information distributed across multimodal files within the user's device. Unlike conventional open-domain QA, these tasks are grounded in user-specific, device-scale contexts, requiring the agent to accurately locate relevant files, interpret heterogeneous content types, and integrate information to produce precise, context-aware answers. For example, a query such as ``I have notes on the maximum flow problem. Which class is the notes record from, and what is the course duration?'' tests whether the agent can identify, understand, and reason over content stored in diverse formats. Solving such tasks does not only require fine-grained retrieval and multimodal comprehension but also long-horizon contextual reasoning across temporally and semantically related files.

\section{Experiment}
\label{sec:experiment}

\subsection{Experimental Setup}
All evaluations are conducted in a controlled, profile-isolated setting. For each test case, a method is given access only to the corresponding profile's file system in \ourmethod{}, with no external retrieval, web access, or auxiliary side-channel metadata. Agents receive the natural-language query as input and may freely explore the environment under full access permissions, using their native mechanisms to search, perceive, and reason over the complete multimodal file corpus.

\begin{table}[!ht]
\caption{\textbf{Main results on HippoCamp across user profiles.}
We evaluate representative MLLMs and agent methods on \textit{profiling} and \textit{factual retention}, reporting F1 and accuracy (Acc) for each archetypal profile and the overall average. Values are percentages (one decimal; \% omitted). Best is highlighted; second-best is underlined.}
\label{tab:main_exp}
\centering
\small
\setlength{\tabcolsep}{4.8pt}
\renewcommand{\arraystretch}{1.10}
\begin{adjustbox}{max width=\textwidth}
\begin{tabular}{@{}l|cccccccc|cccccccc@{}}
\toprule
\multirow{3}{*}{\textbf{Method}} &
\multicolumn{8}{c|}{\textbf{Profiling}} &
\multicolumn{8}{c}{\textbf{Factual Retention}} \\
\cmidrule(lr){2-9} \cmidrule(lr){10-17}
& \multicolumn{2}{c}{(a) Bei} & \multicolumn{2}{c}{(b) Adam} & \multicolumn{2}{c}{(c) Victoria} & \multicolumn{2}{c|}{Overall}
& \multicolumn{2}{c}{(a) Bei} & \multicolumn{2}{c}{(b) Adam} & \multicolumn{2}{c}{(c) Victoria} & \multicolumn{2}{c}{Overall} \\
\cmidrule(lr){2-3} \cmidrule(lr){4-5} \cmidrule(lr){6-7} \cmidrule(lr){8-9}
\cmidrule(lr){10-11} \cmidrule(lr){12-13} \cmidrule(lr){14-15} \cmidrule(lr){16-17}
& F1 & Acc & F1 & Acc & F1 & Acc & F1 & Acc
& F1 & Acc & F1 & Acc & F1 & Acc & F1 & Acc \\
\midrule
\multicolumn{17}{c}{\textbf{\textit{RAG Methods}}} \\ \midrule
Standard RAG~\cite{Lewis2020RAG}
& 13.7 & 10.0 & 20.8 & 35.0 & \cellcolor{yellow!25}\textbf{20.6} & \underline{35.0} & 18.4 & 26.7 & 29.7 & 24.2 & 39.7 & 42.7 & 20.5 & 23.6 & 30.0 & 30.2 \\
Self RAG~\cite{Asai2024SelfRAG} &
\underline{13.8} & 5.0 & 16.0 & 25.0 & 15.9 & 0.0 & 15.2 & 10.0 & 33.9 & 26.1 & 41.5 & 38.8 & 20.2 & 17.7 & 31.9 & 27.5 \\
\midrule
\multicolumn{17}{c}{\textbf{\textit{Search Agent Methods}}} \\ \midrule
ReAct~\cite{Yao2023ReAct} (Qwen3-30B-A3B~\cite{Qwen3VL2025}) & 5.5 & 5.6 & 17.8 & 25.0 & 12.2 & 10.0 & 11.8 & 13.5 & \cellcolor{yellow!25}\textbf{42.4} & 26.1 & \cellcolor{yellow!25}\textbf{60.4} & 37.9 & 26.5 & 21.7 & \cellcolor{yellow!25}\textbf{43.1} & 28.5 \\
ReAct~\cite{Yao2023ReAct} (Gemini-2.5-flash~\cite{Comanici2025Gemini25})
& 13.7 & 10.0 & \underline{21.4} & 25.0 & \underline{20.5} & 25.0 & \underline{18.5} & 20.0 & 26.9 & 24.2 & 35.7 & 55.3 & 17.0 & 36.4 & 26.5 & 38.7 \\
Search-R1~\cite{Jin2025SearchR1} & 6.6 & 0.0 & 16.5 & 15.0 & 9.4 & 0.0 & 10.8 & 5.0 & \underline{38.7} & 23.7 & \underline{58.0} & 28.2 & 26.4 & 24.1 & \underline{41.0} & 25.3 \\
\midrule
\multicolumn{17}{c}{\textbf{\textit{Autonomous Agent Systems}}} \\ \midrule
Terminal Agent (Qwen3-VL-8B-Instruct~\cite{Qwen3VL2025})
& 5.4 & 0.0 & 16.3 & 25.0 & 13.2 & 25.0 & 11.6 & 16.7 & 14.6 & 10.7 & 21.6 & 13.6 & 15.7 & 10.3 & 17.3 & 11.5 \\
Terminal Agent (Gemini-2.5-flash~\cite{Comanici2025Gemini25})  & 9.0 & 5.0 & 17.0 & \underline{45.0} & 19.0 & 25.0 & 15.0 & 25.0 & 21.8 & 18.1 & 33.1 & 31.1 & 24.4 & 20.7 & 26.4 & 23.3 \\
Terminal Agent (GPT-5.2~\cite{openai2025gpt5_2_system_card}) & 8.1 & \underline{15.0} & 14.9 & \underline{45.0} & 10.5 & 30.0 & 11.1 & \underline{30.0} & 13.0 & \underline{29.8} & 31.6 & \underline{59.2} & \underline{29.0} & \underline{55.7} & 24.6 & \underline{48.2} \\
ChatGPT Agent Mode~\cite{openai2024gpt4technicalreport, openai2025gpt5} &
\cellcolor{yellow!25}\textbf{23.8} & \cellcolor{yellow!25}\textbf{35.0} &
\cellcolor{yellow!25}\textbf{22.7} & \cellcolor{yellow!25}\textbf{55.0} &
16.7 & \cellcolor{yellow!25}\textbf{55.0} &
\cellcolor{yellow!25}\textbf{21.0} & \cellcolor{yellow!25}\textbf{48.3} &
20.4 & \cellcolor{yellow!25}\textbf{31.2} &
56.2 & \cellcolor{yellow!25}\textbf{90.3} &
\cellcolor{yellow!25}\textbf{29.3} & \cellcolor{yellow!25}\textbf{67.0} &
35.3 & \cellcolor{yellow!25}\textbf{62.8} \\
\bottomrule
\end{tabular}
\end{adjustbox}
\end{table}

All results in~\cref{tab:main_exp} follow a \textbf{max-budget protocol}, allowing each method to run up to its predefined step or token without artificial wall-clock constraints. Retrieval-augmented (RAG) baselines index the full file corpus using their built-in encoders, followed by retrieval and a single-pass generation step. All files are reformatted as required by each method to ensure fair comparison. Search agent baselines integrate retrieval with iterative reasoning and tool use. They alternate between search actions and evidence-conditioned generation over multiple steps. Autonomous agent baselines are evaluated either in a Dockerized Ubuntu environment that replicates the full file system or in non-vacuum product-grade agent modes. The Docker setup exposes native system capabilities (e.g., Linux terminal and Python execution), while model-based methods access files through standardized APIs with internal modality conversion. In contrast, product-grade agent modes access and search the connected file source through their standard user interfaces. No additional supervision, external plug-ins, or manually curated signals are introduced. To ensure comparability, all systems operate over the same underlying file structures and use their default configurations.

\subsection{Baseline Methods}

\noindent\textbf{RAG methods.}
Standard RAG~\cite{Lewis2020RAG}, Self-RAG~\cite{Asai2024SelfRAG} represent classical retrieval and generation pipelines. They depend on shallow semantic retrieval over multimodal embeddings and assume the relevant evidence will be surfaced in a candidate set. These methods lack multi-step inspection, adaptive search, cross-file synthesis, or persistent memory-capabilities that are essential for \ourmethod{}.
\noindent\textbf{Search Agent methods.}
Search-agent baselines embed retrieval within an explicit multi-turn search--reason loop. Following the ReAct framework~\cite{Yao2023ReAct}, we instantiate ReAct with both Gemini 2.5 Flash~\cite{Comanici2025Gemini25} and Qwen3-30B-A3B~\cite{Qwen3VL2025} models, where the system alternates between reasoning steps and explicit search actions, conditioning subsequent generation on retrieved observations. Search-R1~\cite{Jin2025SearchR1} similarly interleaves reasoning and search through structured tags, enabling dynamic evidence acquisition across multiple steps.
\noindent\textbf{Autonomous Agent methods.}
We evaluate Docker-based autonomous agents instantiated with multiple model backends, including Gemini 2.5 Flash~\cite{Comanici2025Gemini25}, ChatGPT 5.2~\cite{openai2025gpt5_2_system_card}, Claude Sonnet 4.5~\cite{anthropic2025claude-sonnet-4-5}, and Qwen3-VL-8B-Instruct~\cite{Qwen3VL2025}, alongside proprietary product-grade agent modes. In the main table, we report ChatGPT Agent Mode~\cite{openai2025gpt5, openai2024gpt4technicalreport}; Claude-based agent~\cite{anthropic2025claude-sonnet-4-5} settings are discussed separately due to severe file system and long-document processing failures. These systems operate through recursive tool-use: issuing queries, browsing file previews, interpreting intermediate results, and updating hypotheses. They represent the most advanced publicly accessible paradigms for grounded reasoning, though their tool policies and architectural assumptions differ significantly.

\subsection{Evaluation Protocol}
\label{ssec:evaluation}

\begin{figure}[!htb]
    \centering
    \includegraphics[width=0.86\textwidth,keepaspectratio]{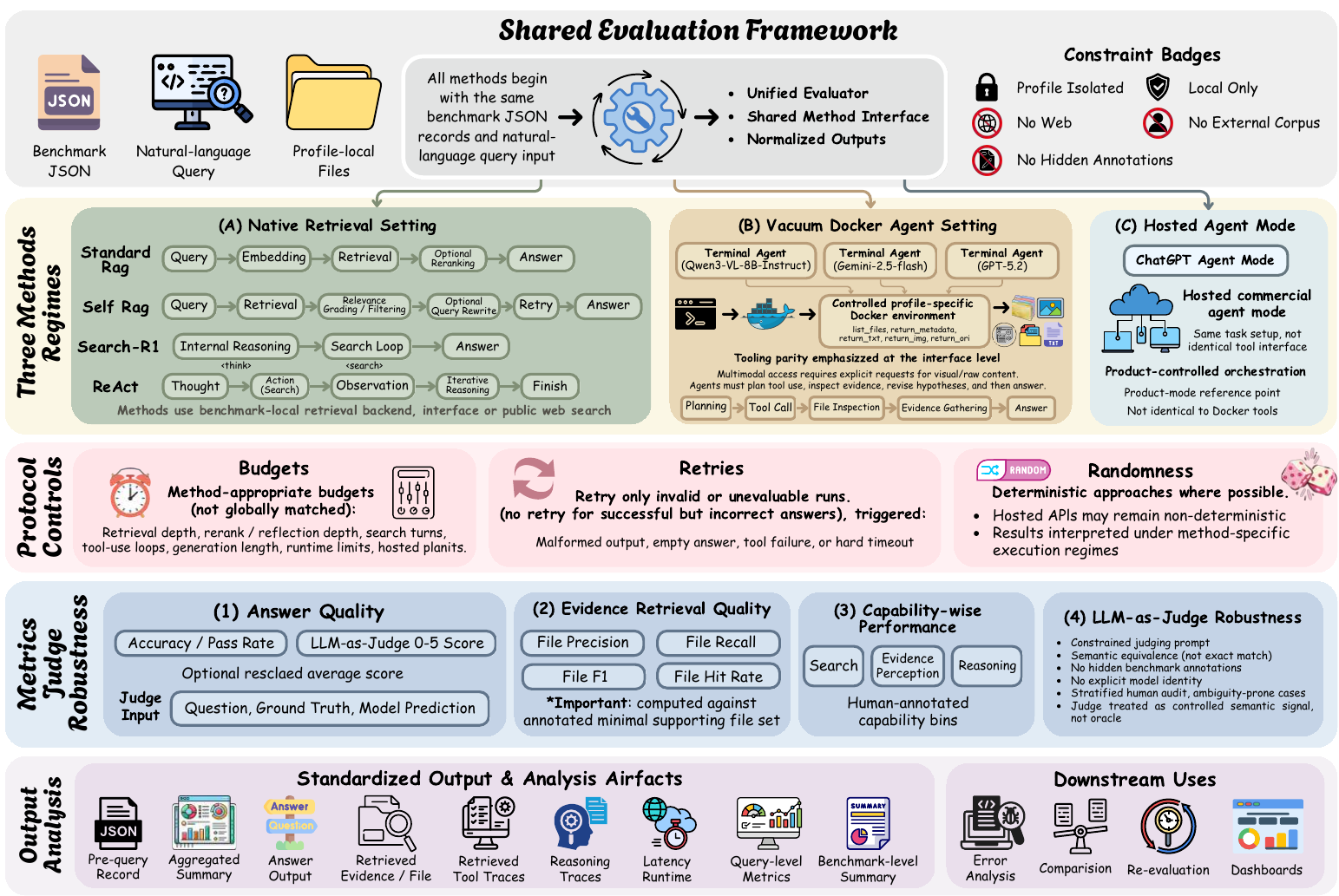}
    \caption{\textbf{Shared evaluation framework for \ourmethod{}.} All methods are evaluated under common profile-local constraints and a shared evaluator, supporting three execution regimes: (A) native retrieval methods, (B) vacuum-Docker terminal agents, and (C) hosted commercial agent modes. The framework standardizes control policies, metric computation, and output artifacts for downstream analysis.}
    \label{fig:evaluation_pipeline}
\end{figure}

To rigorously evaluate agent performance on personalized, multimodal file management, \ourmethod{} employs a unified evaluation protocol covering two core dimensions: \textbf{question answering quality} and \textbf{evidence retrieval accuracy}. Each dimension is assessed using both automatic metrics and large language model (LLM)-based judgment to ensure reliability across all tasks. As illustrated in \cref{fig:evaluation_pipeline}, all methods are dispatched through a common evaluator interface under profile-local access constraints, supporting three execution regimes: native retrieval, vacuum-Docker terminal agents, and hosted commercial agent modes. Full details on execution regimes, budgets, and robustness checks are provided in \cref{app:evaluation_robustness}.

\noindent\textbf{Question answering evaluation.}
For all tasks, we adopt an \textbf{LLM-as-a-judge} ~\cite{Li2025LLMJudge} paradigm. A strong reference LLM \footnote{In this benchmark, we use GPT-4o as the reference LM} is prompted with the question, ground-truth answer, and the model's generated response, and instructed to produce a binary correctness judgment (yes/no) together with a quality score on a standardized 0--5 scale. The prompt explicitly instructs the judge to consider factual alignment, reasoning soundness, and contextual personalization. We report overall accuracy, measured as the fraction of responses judged correct.
\noindent\textbf{Evidence retrieval evaluation.}
For tasks requiring document or file retrieval, we assess retrieval quality using {recall hit rate} and {F1 score} based on the ground-truth evidence file set. F1 captures the balance between retrieving relevant files and avoiding spurious ones, while recall measures the agent's ability to identify all necessary evidence supporting correct reasoning.

\subsection{Experimental Results}

\cref{tab:main_exp} reports results across all profiles and both task types. Overall, a large gap persists between current methods and human experts, exposing fundamental weaknesses in both RAG pipelines and agentic systems for long-horizon personalized file reasoning. These results motivate \ourmethod{} as a diagnostic benchmark that surfaces failures in multimodal retrieval, hierarchical file understanding, and evidence-grounded multi-step reasoning, and they suggest the need for stronger indexing, persistent memory, and verification-centric agent architectures. We next analyze performance by method family.

\noindent\textbf{RAG methods.} RAG pipelines perform poorly overall, especially on profiling, where performance remains low (e.g., Standard RAG: 18.4 F1 / 26.7 Acc overall; Self-RAG: 15.2 F1 / 10.0 Acc overall). Retrieval is brittle and frequently returns shallow or irrelevant files, while generation lacks robust cross-file aggregation. This is most evident on Profile (c) Victoria, where Self-RAG fails to produce any correct profiling answer (0.0 Acc). Factual retention improves only modestly (30.0--31.9 overall F1), largely limited to cases resembling direct lookup; models still often overfit to filenames or directory strings rather than grounding in file content.

\noindent\textbf{Search agent methods.} Iterative search agents improve factual retention by performing multi-step exploration and file inspection, achieving up to 55.3 Acc on Profile (b) Adam (ReAct with Gemini-2.5-flash). Search-R1 attains strong factual-retention F1 on document-heavy profiles (e.g., 58.0 on Profile (b) Adam), suggesting advantages in dense document environments. However, these gains do not transfer to profiling: Search-R1 achieves only 10.8 overall profiling F1 with 5.0 Acc and yields 0.0 Acc on Profiles (a) and (c), indicating that locating candidate files alone is insufficient for synthesizing longitudinal user-level inferences.

\noindent\textbf{Autonomous Agent Systems.}
We further evaluate autonomous agents in a controlled vacuum Docker environment and in native (non-vacuum) agent modes. Overall, these systems yield only moderate gains over search-based agents and remain far below human performance. ChatGPT Agent Mode performs best, reaching 55.0\% profiling accuracy on Profiles (b) Adam and (c) Victoria and achieving the strongest overall scores (profiling: 21.0 F1 / 48.3 Acc; factual retention: 35.3 F1 / 62.8 Acc in \cref{tab:main_exp}). Despite these improvements, substantial errors persist. In practice, the system is computationally expensive (often requiring 10--15 minutes per query) and operationally unstable, frequently producing incomplete outputs or missing file references that necessitate re-execution, thereby increasing run-to-run variance and making observed performance sensitive to execution instability. Claude Sonnet 4.5 is omitted in the vacuum setting due to unreliable long-document processing under its API constraints; in native agent mode it cannot consistently interface with the local file system, leading to near-zero performance. Common failures include imprecise file localization and inconsistent metadata interpretation, with more frequent hallucinated file references or unsupported metadata in the vacuum setting, indicating brittle grounding under constrained access.

\section{Analysis}
\label{sec:analysis}
Our main results (\cref{tab:main_exp}) and capability decomposition (\cref{tab:analysis_exp}) converge on a single overarching finding: \textit{the dominant failure source in \ourmethod{} lies not in evidence retrieval per se, but in the post-retrieval pipeline}---methods frequently locate partially relevant files yet fail to discriminate, ground, integrate, and verify them under profile-local, cross-modal, and temporally extended conditions. We unpack this finding along three axes: metric decoupling and bottleneck localization (\cref{subsec:capability_analysis}), a canonical failure pipeline that unifies observed error modes (\cref{subsec:failure_modes}), and concrete design principles for next-generation file-system agents (\cref{subsec:agent_design}).

\subsection{Capability-wise Decomposition}
\label{subsec:capability_analysis}

\begin{table}[!ht]
  \caption{\textbf{Agent capability-wise analysis on HippoCamp.}
  For the methods in~\cref{tab:main_exp}, we report F1 and LLM-judge accuracy (Acc) aggregated by agent capability labels, decomposed into \textit{search}, \textit{perception}, and \textit{reasoning}, for \textit{profiling} and \textit{factual retention} as well as the overall average. Values are percentages (one decimal; \% omitted). Best is highlighted; second-best is underlined.}
  \label{tab:analysis_exp}
  \centering
  \small
  \setlength{\tabcolsep}{4.4pt}
  \renewcommand{\arraystretch}{1.10}
  \begin{adjustbox}{max width=\textwidth}
  \begin{tabular}{@{}l|cccccccc|cccccccc@{}}
  \toprule
  \multirow{3}{*}{\textbf{Method}} &
  \multicolumn{8}{c|}{\textbf{Profiling}} &
  \multicolumn{8}{c}{\textbf{Factual Retention}} \\
  \cmidrule(lr){2-9} \cmidrule(lr){10-17}
  & \multicolumn{2}{c}{Search} & \multicolumn{2}{c}{Perception} & \multicolumn{2}{c}{Reasoning} & \multicolumn{2}{c|}{Overall}
  & \multicolumn{2}{c}{Search} & \multicolumn{2}{c}{Perception} & \multicolumn{2}{c}{Reasoning} & \multicolumn{2}{c}{Overall} \\
  \cmidrule(lr){2-3} \cmidrule(lr){4-5} \cmidrule(lr){6-7} \cmidrule(lr){8-9}
  \cmidrule(lr){10-11} \cmidrule(lr){12-13} \cmidrule(lr){14-15} \cmidrule(lr){16-17}
  & F1 & Acc & F1 & Acc & F1 & Acc & F1 & Acc
  & F1 & Acc & F1 & Acc & F1 & Acc & F1 & Acc \\
  \midrule
  \multicolumn{17}{c}{\textbf{\textit{RAG Methods}}} \\ \midrule
  Standard RAG~\cite{Lewis2020RAG}
  & 26.4 & 26.2 & 21.8 & 13.8 & 26.7 & 25.5 & 25.0 & 21.8
  & 19.0 & 26.2 & \underline{21.1} & 28.7 & 13.0 & 19.1 & 17.7 & 24.7 \\
  Self RAG~\cite{Asai2024SelfRAG}
  & 27.8 & 23.1 & 23.1 & 13.2 & 27.7 & 22.4 & 26.2 & 19.6
  & 15.2 & 8.9 & 16.4 & 12.2 & 10.7 & 7.0 & 14.1 & 9.4 \\
  \midrule
  \multicolumn{17}{c}{\textbf{\textit{Search Agent Methods}}} \\ \midrule
  ReAct~\cite{Yao2023ReAct} (Qwen3-30B-A3B~\cite{Qwen3VL2025})
  & \cellcolor{yellow!25}\textbf{36.3} & 26.1 & \underline{27.9} & 16.7
  & \cellcolor{yellow!25}\textbf{36.3} & 23.9 & \underline{33.5} & 22.2
  & 11.9 & 13.4 & 14.5 & 18.7 & 8.6 & 10.6 & 11.7 & 14.2 \\
  ReAct~\cite{Yao2023ReAct} (Gemini-2.5-flash~\cite{Comanici2025Gemini25})
  & 23.5 & 34.9 & 20.5 & 20.4 & 23.4 & 33.0 & 22.5 & 29.4
  & \underline{19.4} & 19.1 & \cellcolor{yellow!25}\textbf{21.3} & 23.4 & 13.2 & 14.8 & \underline{18.0} & 19.1 \\
  Search-R1~\cite{Jin2025SearchR1}
  & \underline{34.8} & 24.9 & \cellcolor{yellow!25}\textbf{32.8} & 15.7
  & \underline{35.1} & 25.8 & \cellcolor{yellow!25}\textbf{34.2} & 22.1
  & 10.2 & 3.8 & 12.3 & 7.2 & 7.8 & 3.9 & 10.1 & 5.0 \\
  \midrule
  \multicolumn{17}{c}{\textbf{\textit{Autonomous Agent Systems}}} \\ \midrule
  Terminal Agent (Qwen3-VL-8B-Instruct~\cite{Qwen3VL2025})
  & 15.8 & 11.1 & 14.6 & 5.7 & 16.9 & 11.4 & 15.8 & 9.4
  & 12.2 & 18.6 & 14.0 & 24.2 & 8.1 & 12.2 & 11.4 & 18.3 \\
  Terminal Agent (Gemini-2.5-flash~\cite{Comanici2025Gemini25})
  & 24.5 & 21.0 & 26.9 & 13.7 & 25.2 & 21.1 & 25.5 & 18.6
  & 14.9 & 23.3 & 14.4 & 24.7 & 12.2 & 17.2 & 13.8 & 21.7 \\
  Terminal Agent (GPT-5.2~\cite{openai2025gpt5_2_system_card})
  & 23.6 & \underline{46.3} & 15.4 & \underline{27.3} & 22.6 & \underline{44.1} & 20.5 & \underline{39.2}
  & 10.0 & \underline{27.2} & 10.2 & \underline{29.7} & \underline{15.5} & \cellcolor{yellow!25}\textbf{36.4} & 11.9 & \underline{31.1} \\
  ChatGPT Agent Mode~\cite{openai2024gpt4technicalreport, openai2025gpt5}
  & 28.9 & \cellcolor{yellow!25}\textbf{56.5} & 15.1 & \cellcolor{yellow!25}\textbf{28.5}
  & 30.2 & \cellcolor{yellow!25}\textbf{55.8} & 24.7 & \cellcolor{yellow!25}\textbf{46.9}
  & \cellcolor{yellow!25}\textbf{19.5} & \cellcolor{yellow!25}\textbf{49.1} & 17.3 & \cellcolor{yellow!25}\textbf{55.5}
  & \cellcolor{yellow!25}\textbf{30.4} & \underline{33.8} & \cellcolor{yellow!25}\textbf{22.4} & \cellcolor{yellow!25}\textbf{46.1} \\
  \bottomrule
  \end{tabular}
  \end{adjustbox}
  \end{table}


\noindent\textbf{Profiling requires a different capability composition than factual retention.}
Across all methods in~\cref{tab:main_exp}, profiling accuracy is systematically lower than factual retention---and the gap widens for search-centric methods. Search-R1 drops from 25.3\% factual accuracy to 5.0\% on profiling, a five-fold decline. ReAct (Qwen3) shows a similar pattern: 28.5\% factual versus 13.5\% profiling. ChatGPT Agent Mode, by contrast, narrows the gap to 62.8\% versus 48.3\%. Factual retention rewards localized, file-grounded lookup; profiling demands weak-signal aggregation across temporally extended traces, referent disambiguation among co-occurring entities, and longitudinal abstraction into stable user-level patterns. Methods built around retrieval strength alone lack this capability composition.

\noindent\textbf{Profile difficulty is governed by structural clarity and entity ambiguity, not domain content.}
Performance varies consistently across profiles (\cref{tab:main_exp}). Adam's document-centric legal environment yields the highest scores: ChatGPT Agent Mode reaches 90.3\% factual accuracy and 55.0\% profiling accuracy. Victoria's finance environment falls in the middle. Bei's socially entangled, media-rich college environment is hardest: the same system drops to 31.2\% factual and 35.0\% profiling. This gradient holds across method families. The underlying factor is not domain complexity but how explicitly personal structure is externalized. Structured environments with clear role boundaries and formal documents are easier; informal, multi-entity environments amplify referential ambiguity and weaken the signal-to-noise ratio of individual traces.

\noindent\textbf{Retrieval quality and answer quality decouple in two distinct modes.}
\cref{tab:main_exp} reveals a systematic divergence between F1 and accuracy. When F1 exceeds accuracy---as with ReAct (Qwen3) on factual retention (43.1\% F1 vs.\ 28.5\% Acc)---the method retrieves relevant files but fails to convert them into correct answers, indicating a breakdown in evidence discrimination or synthesis. When accuracy exceeds F1---as with Terminal Agent GPT-5.2 on factual retention (24.6\% F1 vs.\ 48.2\% Acc)---the method produces correct answers without retrieving the ground-truth evidence files, suggesting reliance on parametric knowledge rather than grounded file evidence. Both modes confirm that search competence and answer quality are separable.


To localize where in the agent pipeline these gaps originate, we turn to the capability-wise decomposition in~\cref{tab:analysis_exp}.


\noindent\textbf{Search is necessary but not decisive.}
Search-centric agents achieve the highest retrieval F1 on profiling: ReAct (Qwen3) reaches 36.3\% and Search-R1 reaches 34.8\%. Their overall profiling accuracy, however, lags far behind---22.2\% and 22.1\%, respectively. ChatGPT Agent Mode inverts this pattern: a lower search F1 of 28.9\% but the highest accuracy of 56.5\%. Locating candidate files is a prerequisite, not a solution.

\noindent\textbf{Perception is the most universal bottleneck.}
Across all method families, perception accuracy on profiling is uniformly low, ranging from 13.2\% (Self-RAG) to 28.5\% (ChatGPT Agent Mode). Even the strongest system achieves perception accuracy roughly half of its search accuracy (28.5\% vs.\ 56.5\%). This gap reflects more than OCR or vision failures. It captures the broader difficulty of converting heterogeneous file content---PDFs, calendar entries, photos, voice memos---into evidence units that can support downstream reasoning.

\noindent\textbf{Reasoning quality is contingent on prior evidence discrimination.}
Reasoning is not an independent capability; it inherits errors from earlier stages. Terminal Agent (GPT-5.2) achieves 44.1\% profiling reasoning accuracy despite only 27.3\% perception accuracy, raising the possibility that some correct answers stem from parametric knowledge rather than grounded evidence---consistent with the Acc-exceeds-F1 pattern observed in~\cref{tab:main_exp}. Search-R1 shows the converse: strong retrieval signals (35.1\% reasoning F1) but weak answer commitment (25.8\% accuracy). Strong reasoning cannot compensate for weak evidence selection, and high retrieval F1 does not guarantee that the evidence reaching the reasoning stage is correct.

\subsection{Systematic Failure Analysis}
\label{subsec:failure_modes}

\begin{figure}[ht]
  \centering
  \includegraphics[width=0.88\linewidth]{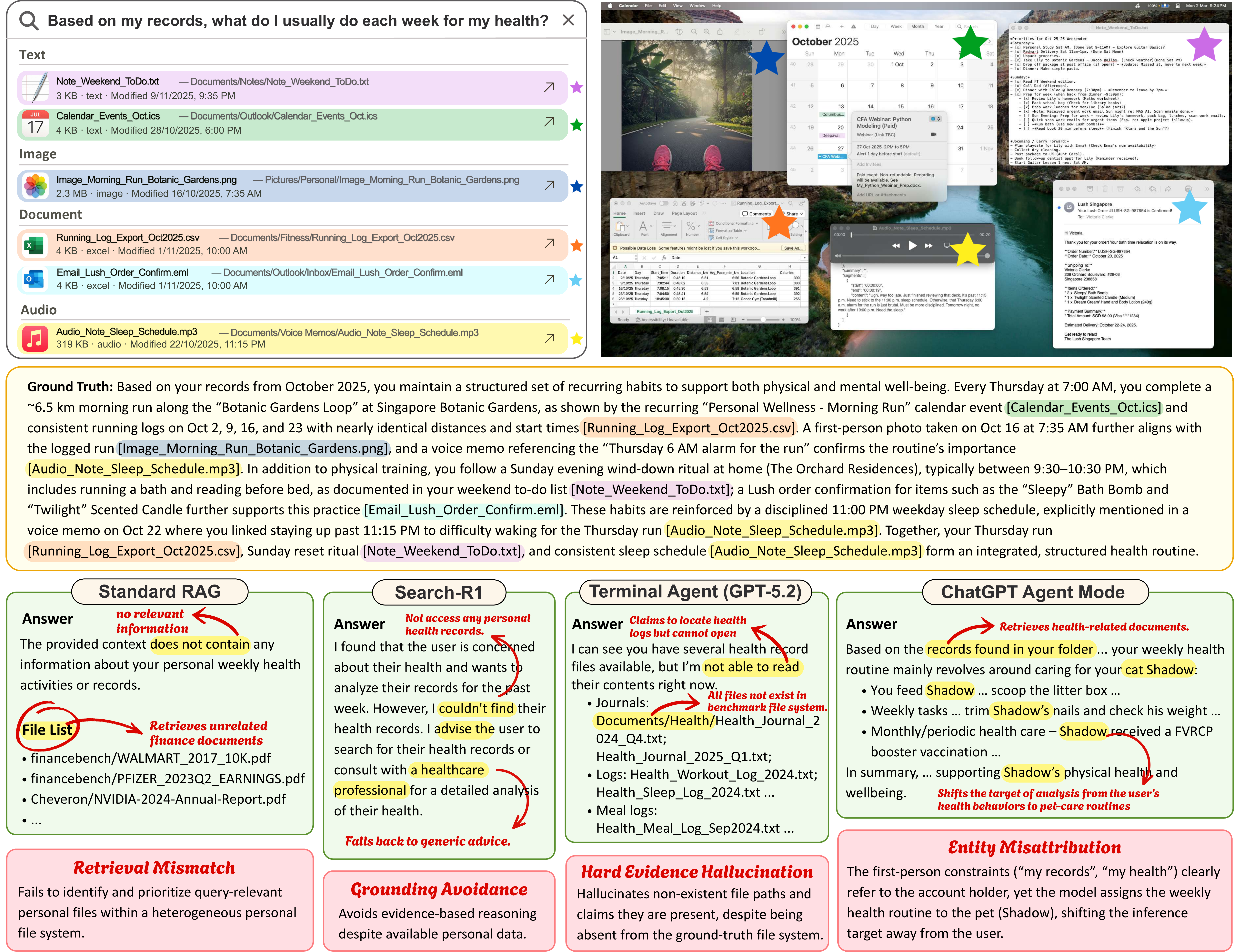}
  \caption{\textbf{Representative failure and success patterns on a cross-modal profiling query.} The query requires aligning evidence across heterogeneous personal files and modalities. Standard RAG exhibits \emph{retrieval mismatch} (retrieving unrelated finance documents), Search-R1 shows \emph{grounding avoidance} (generic advice despite available evidence), Terminal Agent (GPT-5.2) triggers \emph{hard evidence hallucination} (fabricated file paths/metadata), and ChatGPT Agent Mode makes an \emph{entity misattribution} error (shifting the referent from the user to the pet), while iterative exploration yields the strongest overall behavior.}
  \label{fig:4_error_analysis}

\end{figure}

The capability gaps identified above manifest as concrete error patterns in agent outputs. \cref{fig:4_error_analysis} illustrates these on a representative profiling query---``Based on my records, what do I usually do each week for my health?''---whose ground truth requires cross-modal synthesis over six file types (calendar events, running logs, photos, voice memos, text notes, and emails). These cases are not isolated anecdotes. They form a recurring failure pipeline that progresses through five stages: off-target retrieval, grounding avoidance, evidence hallucination, entity misbinding, and missing verification. Different methods break at different stages, which explains why their capability profiles diverge even when overall accuracy is similar.

\noindent\textbf{Retrieval mismatch.}
RAG-based systems often fail to distinguish user-relevant personal files from semantically related but contextually irrelevant documents. In the case study, Standard RAG has retrieved financial reports, in our case, \texttt{WALMART\_2017\_10K.pdf}, and concludes that no relevant evidence exists, consistent with embedding-level ambiguity from keyword overlap (e.g., ``health'' in disclosures). This mismatch cascades to weak downstream grounding: Standard RAG attains 26.4\% search F1 but only 13.8\% perception accuracy on profiling as shown in \cref{tab:analysis_exp}. Self-RAG collapses more sharply, with 0.0\% profiling accuracy on Victoria shown in~\cref{tab:main_exp}, indicating that self-reflection cannot recover from off-target initial retrieval. This is the earliest failure point in the pipeline: once the retrieval direction is wrong, all downstream stages inherit the error.

\noindent\textbf{Grounding avoidance.}
Even when retrieval surfaces partially relevant candidates, the pipeline can fail at the evidence commitment stage. Reasoning-centric agents locate candidates yet avoid committing to evidence-grounded answers. In the representative example, Search-R1 defaults to generic advice despite relevant health files being present. Quantitatively, this appears as a pronounced F1--accuracy gap: Search-R1 achieves 34.2\% profiling F1 in \cref{tab:analysis_exp} but only 5.0\% profiling accuracy in \cref{tab:main_exp}, including 0.0\% accuracy on Bei and Victoria despite 7--17\% profiling F1. A plausible explanation is that, under distribution shift in personal files, the model favors ``safe'' parametric responses over risky grounding in unfamiliar evidence.

\noindent\textbf{Hard evidence hallucination.}
A more dangerous grounding failure occurs when agents fabricate evidence rather than abstaining. Terminal-based agents operating in sandboxed environments frequently invent file paths or metadata. The case study shows Terminal Agent with GPT-5.2 as backbone inventing health files, in our case, \texttt{Health\_Journal\_2024\_Q4.txt}, and then claiming it cannot open them, producing an internally consistent but fabricated evidence chain. Capability scores support this interpretation: the agent shows high search accuracy (46.3\%) but low perception F1 (15.4\%) on profiling in~\cref{tab:analysis_exp}, suggesting actions proceed without verifiable grounding. Its profiling accuracy exceeding F1 (30.0\% vs.\ 11.1\% in~\cref{tab:main_exp}) further indicates occasional correct outputs may be driven by parametric knowledge rather than grounded file evidence.

\noindent\textbf{Entity misattribution.}
Even when evidence is real and relevant, binding it to the correct referent remains a distinct failure stage. The strongest system can retrieve genuine health records yet assign them to the wrong entity. ChatGPT Agent Mode locates genuine health-related records but attributes the routine to the user's pet rather than the user, exposing difficulty in first-person constraint resolution when multiple entities share the same file system. Consistently, its retrieval outpaces comprehension (56.5\% search accuracy vs.\ 28.5\% perception accuracy on profiling in~\cref{tab:analysis_exp}). The effect is profile-dependent: it reaches 90.3\% factual accuracy on Adam's document-centric legal profile but only 31.2\% on Bei, where richer interpersonal context increases entity ambiguity in \cref{tab:main_exp}.

\noindent\textbf{Verification deficit.}
None of the evaluated methods include an explicit final-stage check that re-examines whether the generated answer is traceable to a minimal, coherent evidence set. The consequence is visible in the F1--accuracy gaps throughout~\cref{tab:analysis_exp}: retrieval and reasoning F1 consistently exceed answer accuracy, indicating that errors introduced at earlier stages---misbinding, hallucination, or insufficient discrimination---propagate unchecked to the final output.

\noindent\textbf{Success pattern: iterative discovery.}
ChatGPT Agent Mode exhibits a qualitatively distinct strategy of iterative file-system exploration. By repeatedly listing directories and reading candidate files, it progressively refines hypotheses and can recover from early missteps, yielding the best overall performance (48.3\% profiling accuracy and 62.8\% factual accuracy in \cref{tab:main_exp}) and the most balanced capability profile (\cref{tab:analysis_exp}). The gains are strongest in structured domains, reaching 55.0\% profiling accuracy and 90.3\% factual accuracy on Adam (\cref{tab:main_exp}), but remain limited for cross-modal profiling where signals are weaker and more heterogeneous, reaching only 35.0\% profiling accuracy on Bei. Its advantage is not stronger one-shot retrieval but the ability to revisit and correct errors at multiple stages of the pipeline---recovering from off-target queries, refining evidence hypotheses, and progressively narrowing the support set.

\subsection{Prospect: Designing the Next-Generation File-system Agent}
\label{subsec:agent_design}

\noindent\textbf{Evaluating true personalization beyond static personas.} Unlike prior benchmarks that rely on explicit, static personas, HippoCamp demonstrates that true personalization in local digital environments is fundamentally a multimodal, cross-file reasoning challenge. To authentically model an individual user's digital life, agents must synthesize implicit behavioral signals scattered across heterogeneous file types rather than simply retrieving explicit statements. Furthermore, they must successfully disambiguate the account holder from other surrounding entities to avoid critical misattributions, while simultaneously reasoning over the long-term temporal continuity of the user's digital footprint. By requiring the longitudinal synthesis of these complex, evolving constraints, HippoCamp moves beyond generic retrieval to provide a rigorous evaluation of how well an agent can genuinely adapt to a personalized computing ecosystem.

\noindent\textbf{How to design a good file-system agent?}
The gap between human capability and agent performance (\cref{tab:analysis_exp}) suggests that robust file-system agents require a tighter coupling of structure, interaction, and verification. Concretely, agents should treat file-system hierarchies and cross-file relations as inductive biases for search, replace one-shot retrieval with iterative information foraging guided by metadata and lightweight checks, and integrate evidence-centric verification loops that bind intermediate inferences to concrete file paths and localized evidence to mitigate hallucinations (\cref{fig:4_error_analysis}).

\noindent\textbf{Structure-aware search.}
File hierarchies, temporal regularities, attachment relations, and version histories encode organizational intent that pure embedding similarity discards. Our results show that methods relying solely on semantic retrieval frequently confuse topically similar but contextually irrelevant files (\cref{subsec:failure_modes}). Treating file-system structure as an inductive bias for search---rather than flattening all files into a single vector index---can reduce off-target retrieval at the pipeline's earliest stage.

\noindent\textbf{Evidence narrowing before answering.}
The consistent gap between retrieval F1 and answer accuracy across \cref{tab:main_exp,tab:analysis_exp} indicates that surfacing relevant files does not automatically yield correct answers. Search-centric methods achieve the highest retrieval F1 on profiling yet the lowest answer accuracy (\cref{tab:analysis_exp}), suggesting that broad candidate pools without subsequent evidence selection introduce noise that downstream reasoning cannot filter. Agents should form a minimal, sufficient support set before generating an answer, rather than summarizing from all retrieved candidates.

\noindent\textbf{Profile-local entity modeling.}
Personal file systems contain multiple co-occurring entities---the account holder, family members, pets, colleagues---that share files and contexts. The entity misattribution failures in \cref{fig:4_error_analysis} indicate that agents must explicitly maintain and update referent models rather than assuming all first-person references map to a single user. Without such modeling, correct retrieval and perception still yield wrong answers.

\noindent\textbf{Verification as an explicit final stage.}
No evaluated method includes a verification loop that re-binds the final answer to localized, file-level evidence and checks for internal contradictions. Adding such a stage would directly address the verification deficit identified in \cref{subsec:failure_modes} and help close the gap between retrieval quality and answer quality. A robust file-system agent must not only retrieve broadly and synthesize across files, but also confirm that its response is traceable to a minimal, coherent support set.

\section{Conclusion}
\label{sec:conclusion}
This work introduces \ourmethod{}, a benchmark evaluating agents' ability to search, perceive, and reason over realistic, multimodal personal file systems. Our analysis reveals that the dominant bottleneck lies not in evidence retrieval but in the post-retrieval pipeline---evidence discrimination, multimodal grounding, entity binding, and final verification---and that profiling demands a qualitatively different capability composition from factual retention. Personalized multimodal memory remains an unsolved challenge. \ourmethod{} provides a rigorous foundation to diagnose these shortcomings and guide the development of next-generation personal file-system agents.

\clearpage
\appendix

\phantomsection
\addcontentsline{toc}{section}{Appendix}

\section*{Appendix}
\label{app:appendix}

This appendix provides supplementary benchmark construction, annotation, task, and evaluation details:
\begin{itemize}[noitemsep,topsep=0pt,parsep=0pt,partopsep=0pt,leftmargin=1.5em]
    \item \S\ref{app:dataset_construction} details participant selection, interview protocol, archetype aggregation, privacy filtering, external augmentation, and file-system statistics.
    \item \S\ref{app:annotation_schema} presents the trajectory schema, evidence-unit design, human-in-the-loop QA pipeline, and agreement procedures.
    \item \S\ref{app:tasks_difficulty} expands the task taxonomy, difficulty definitions, and representative profile examples.
    \item \S\ref{app:evaluation_robustness} describes evaluation settings, budgets, metrics, robustness checks, and extended result summaries.
\end{itemize}

\section{Dataset Construction and Profile Aggregation}
\label{app:dataset_construction}
\subsection{Participant Pool and Source Selection}
\label{sec:sm_pool}

HippoCamp is constructed from interviews with \textbf{100+} personal-device users. This section specifies the participant screening and source-selection protocol used to form the candidate pool prior to profile aggregation. Recruitment spans varied ages, living situations, and technical backgrounds, with stratified screening over \emph{demographic, socio-economic, professional, and lifestyle} dimensions to preserve diversity while maintaining comparable device-resident usage intensity. We retain only participants whose devices satisfy the following reproducible criteria:

\textbf{(C1) File-system richness and modality coverage.} The device must contain a dense, heterogeneous corpus with at least \textbf{500} user files, covering \textbf{at least 4} of the five modalities (text, document, image, video, audio) and \textbf{at least 10} distinct file extensions.

\textbf{(C2) Longitudinal depth.} The corpus must cover at least a 3-month span of creation or modification activity, and the participant must report sustained full-time study or professional practice with a personal workstation or laptop as the primary device, yielding stable routines and recurring workflows.

\textbf{(C3) Evidence completeness and auditability.} Candidate sources must support auditable user-level inference with a minimal evidence checklist: (i) cross-file corroboration for key personal facts (at least two independent supporting artifacts), (ii) consistent temporal anchors (timestamps or dated records) for long-horizon behaviors, and (iii) interpretable organizational traces (directory structure, naming conventions, or recurring records) that enable reconstructing schedules, routines, and workflows without relying on unverifiable narrative. This criterion excludes ``profile holes'' in which profiling claims cannot be grounded in verifiable evidence.

We further exclude sources dominated by rigid corporate IT templates or centrally managed directory schemes that obscure user-driven organization. Candidates failing any criterion are removed prior to aggregation, and the remaining contributors form the screened candidate pool used in Appendix~A.3. Finally, before any release-facing processing, we obtain explicit consent and apply privacy safeguards: participants provide anonymized directory trees and representative files under controlled handling, and sensitive identifiers are redacted or pseudonymized while preserving evidence-bearing cues (\cref{sec:sm_privacy}).

\subsection{In-Depth Interview Protocol}
\label{sec:sm_interview}

To support profile reconstruction and downstream validation, we conduct protocol-guided interviews (60--90 minutes per participant) prior to data extraction. The interview protocol is designed to elicit reproducible information about device scope, organizational habits, recurring workflows, ambiguity-prone regions, and representative information needs, which are later used for source filtering, aggregation checks, and QA validation.

\noindent\textbf{Environment scoping.}
Participants describe primary devices, synchronization practices, and recurring task cycles. These responses define the operational scope of the personal environment used for subsequent extraction and validation.

\noindent\textbf{File-system mental model.}
We elicit top-level folder semantics, naming conventions, temporal or project-based grouping rules, and the metadata cues participants rely on when locating historical materials. These responses are used to check whether reconstructed directory structure and retrieval-oriented organization remain behaviorally plausible after aggregation.

\noindent\textbf{Workflow reconstruction.}
Participants reconstruct concrete task episodes end-to-end (e.g., preparing deliverables, tracking versions, revisiting ongoing projects), exposing cross-file dependencies and multi-step retrieval patterns. These episodes serve as validation targets when constructing grounded trajectories and representative task instances.

\noindent\textbf{Boundary and breakdown analysis.}
We probe irregular regions (downloads, desktop dumps, temporary folders) and time-pressured failure cases to identify systematic noise sources and ambiguity patterns. These observations inform exclusion rules, privacy filtering, and later error-oriented QA design.

\noindent\textbf{Intent elicitation for task design.}
Using domain-relevant prompts, we elicit recurring personal-computing needs in participants' own terms, especially those involving verification, multi-source aggregation, and longitudinal synthesis. In Appendix A, these intents function as reference constraints for later QA authoring rather than as direct benchmark instances.

\noindent\textbf{Post-extraction validation.}
After the directory crawl, participants review the extracted structure to confirm representativeness, flag irrelevant or missing elements, and approve privacy-preserving reconstruction. This step serves as a final consistency check before profile aggregation.

\subsection{Archetype Aggregation}
\label{sec:sm_aggregation}

We construct three archetypal profiles by aggregating files from multiple contributors. In this appendix, \emph{coherence} means that each resulting profile forms a logically closed and behaviorally plausible environment in which timestamps, entities, and cross-file references remain mutually consistent, so that both profiling and factual-retention queries admit verifiable file-grounded evidence. To enforce this requirement, aggregation proceeds through four stages: distribution-preserving partitioning, coherence checks, minimal-edit repair, and final human validation.

\textbf{Distribution-preserving aggregation.}
Contributors are partitioned into three archetypal profiles under constraints on modality composition, file-type frequencies, and high-level organizational patterns. The partitioning procedure is designed to preserve long-tail format coverage, directory-structure diversity, and the relative balance between personal, academic, and professional materials, rather than concentrating specific formats or workflows in a single profile. When multiple allocations satisfy these constraints, we prefer assignments that maintain stronger internal compatibility in temporal scope, project structure, and recurring usage patterns.

\textbf{Coherence checks.}
After contributor partitioning and before any minimal repair, we apply automated and manual validation to verify three forms of profile-level consistency: (i) \emph{temporal consistency} across timestamped artifacts (e.g., calendars, emails, logs, and media metadata) to avoid incompatible timelines; (ii) \emph{entity and identity consistency} to prevent collisions among recurring names, identifiers, and persistent entities that would create contradictory narratives; and (iii) \emph{project and workflow consistency} to ensure that multi-file threads (e.g., course materials, case folders, and analyses) remain internally coherent, with valid cross-file references and dependencies. Detected violations are resolved by source re-assignment or, when necessary, removal of the minimal number of conflicting items.

\textbf{Minimal-edit policy.} 
Edits are introduced only when required to repair contradictions that would otherwise break profile-level consistency. Typical interventions include pseudonym alignment, removal of duplicated or conflicting identifiers, and limited normalization of metadata or filenames when these are necessary to restore valid cross-file references. We do not introduce edits to enrich, stylize, or artificially diversify the reconstructed environments; when a conflict cannot be repaired locally, we instead remove the minimal number of offending items.

\textbf{Human validation.}
Finally, the aggregated profiles are reviewed as final candidate environments to verify cross-file consistency, representative structure, and the absence of unresolved privacy or coherence violations before benchmark release. Contributing annotators then provide end-to-end sign-off that the assembled environments preserve native organizational patterns and habitual usage signals. These final checks produce three coherent archetypal profiles suitable for subsequent QA construction and evidence-grounded evaluation.

\subsection{Privacy, Filtering, and Anonymization}
\label{sec:sm_privacy}

We adopt a privacy-first data governance protocol designed to eliminate leakage of personally identifying information (PII) while preserving the evidence-bearing cues required for benchmark evaluation. Participation is strictly opt-in with explicit consent, and contributors retain the right to withdraw their data at any time; upon withdrawal, the corresponding files are removed from the candidate pool and from any aggregated profile derived from them. All handling is performed under controlled access, and no raw data are used for annotation or analysis until privacy processing and participant verification are completed.

\textbf{Filtering of system-generated artifacts.}
To avoid skewing file-system statistics with non-user content, we remove \emph{system-generated, non-user artifacts} that do not reflect user intent (e.g., OS caches, application temporary files, indexing databases, recycle-bin remnants, and other background transient files). Filtering is implemented via a reproducible rule set combining path-based patterns (e.g., OS cache directories), file-type and filename heuristics (e.g., known cache extensions), and duplicate/near-duplicate detection. This step is conservative with respect to user-generated content: we retain ordinary files even when small or sparse, and only exclude artifacts that are clearly OS- or application-generated and semantically non-content-bearing.

\textbf{Anonymization with evidence preservation.}
All privacy-relevant content in the benchmark is either \emph{synthetically generated} using a proprietary image-generation model (Nano Banana~\cite{Comanici2025Gemini25}) from the fictional identity profiles or \emph{reproduced} to preserve task structure while removing identifiers. Any externally sourced non-sensitive assets are included only under licenses permitting redistribution and commercial use; no raw personal media containing real-world identifiers are retained. We anonymize sensitive identifiers through redaction and consistent pseudonymization while preserving the minimal cues needed for evidence grounding. Specifically, we (i) replace personally identifying names, emails, phone numbers, addresses, IDs, account handles, and organization names with stable pseudonyms that are consistent within a profile, (ii) remove or sanitize embedded metadata fields that may reveal identity (e.g., author fields, device identifiers, GPS coordinates when present), and (iii) preserve structural and temporal signals necessary for evaluation, such as directory hierarchy, relative timestamps and ordering, and cross-file references. Where redaction is applied, we maintain format fidelity (e.g., keeping date and numeric patterns) to avoid breaking downstream verification tasks, and we ensure that anonymization does not create spurious evidence.

\textbf{Participant verification and final approval.}
Crucially, anonymization is followed by a participant review step. Contributors inspect the processed outputs (directory tree and redacted files) and approve that no sensitive information remains and that the remaining content is an acceptable representation of their device context. Only data that pass this participant check are admitted to downstream aggregation and annotation. Together, filtering, anonymization, and participant verification ensure that HippoCamp contains no identifiable personal information while retaining the evidence cues needed for grounded evaluation.

\subsection{External Benchmark used for Data Augmentation}
\label{app:external-augmentation}

After profile aggregation (\cref{sec:sm_aggregation}), we observed an ecological asymmetry: certain specialized document \emph{forms} that frequently arise in professional workflows are naturally underrepresented in personal devices due to confidentiality constraints (e.g., regulatory disclosures, client-facing contracts, and financial statements). This sparsity is most salient for the document-centric profiles (b) Adam (Law) and (c) Victoria (Finance), and it can reduce coverage of realistic task patterns that depend on such materials. To improve coverage without compromising privacy or coherence, we incorporate a limited amount of curated public-domain source material for these two profiles. Specifically, we use FinanceBench-derived documents for Finance and LegalBench-RAG-derived documents for Law as document-form enrichment rather than direct benchmark merging. Any imported content is rewritten/sanitized to match HippoCamp's fictional identity system (\cref{sec:sm_privacy}), and all resulting QA and trajectories are re-annotated under our schema with complete contextual metadata (e.g., file paths, timestamps, and directory hierarchy signals). These materials are used only to supplement underrepresented document forms; the benchmark semantics, task construction, and final annotations remain governed by the HippoCamp pipeline.

\textbf{FinanceBench integration~\cite{islam2023financebench}.}
FinanceBench provides analyst-relevant document types. We incorporate its documents into the Finance profile as additional evidence-bearing artifacts. FinanceBench QA items, when used, are treated strictly as screened candidate material and are rewritten or adapted into user-need-driven questions consistent with the fictional persona and HippoCamp task taxonomy, rather than reused verbatim. All retained questions are paired with newly constructed step-wise rationales, localized evidence, and file grounding, with metadata normalized to our trajectory format.

\textbf{LegalBench-RAG integration~\cite{pipitone2024legalbench}.}
LegalBench-RAG requires heavier adaptation because its original entities and QA are tied to real jurisdictions and case identifiers. We therefore use its QA only as inspiration or seed candidates after screening, and systematically rewrite questions and sanitize document content to remove real-world entities while preserving task-relevant legal structure. The goal is to retain realistic legal document structure, not to preserve the original benchmark instance as such. Documents irrelevant to the rewritten tasks are removed, and the remaining corpus is brought into full temporal/identity consistency with the fictional Law persona (\cref{sec:sm_privacy}). All retained questions are re-annotated from scratch under HippoCamp's trajectory schema, including complete metadata, evidence localization, and step-wise reasoning traces.

\textbf{Distribution and coherence safeguards.}
To avoid biasing the benchmark toward augmented sources, we cap the contribution of externally sourced documents such that profile-level modality and file-type statistics remain within the distributional constraints enforced during aggregation (\cref{sec:sm_aggregation}). All augmented items undergo the same coherence checks and human validation as native-source materials before inclusion. These externally sourced documents constitute only a minority supplement to the overall corpus and are introduced solely to improve coverage and structural completeness in underrepresented document regimes. Accordingly, task semantics, annotation protocols, and evaluation targets remain defined by HippoCamp rather than inherited from the source benchmarks or their original label spaces.

\subsection{File-System Statistics}
\subsubsection{Modality Composition by Profile}
\begin{figure}[ht]
  \centering
  \includegraphics[width=\linewidth]{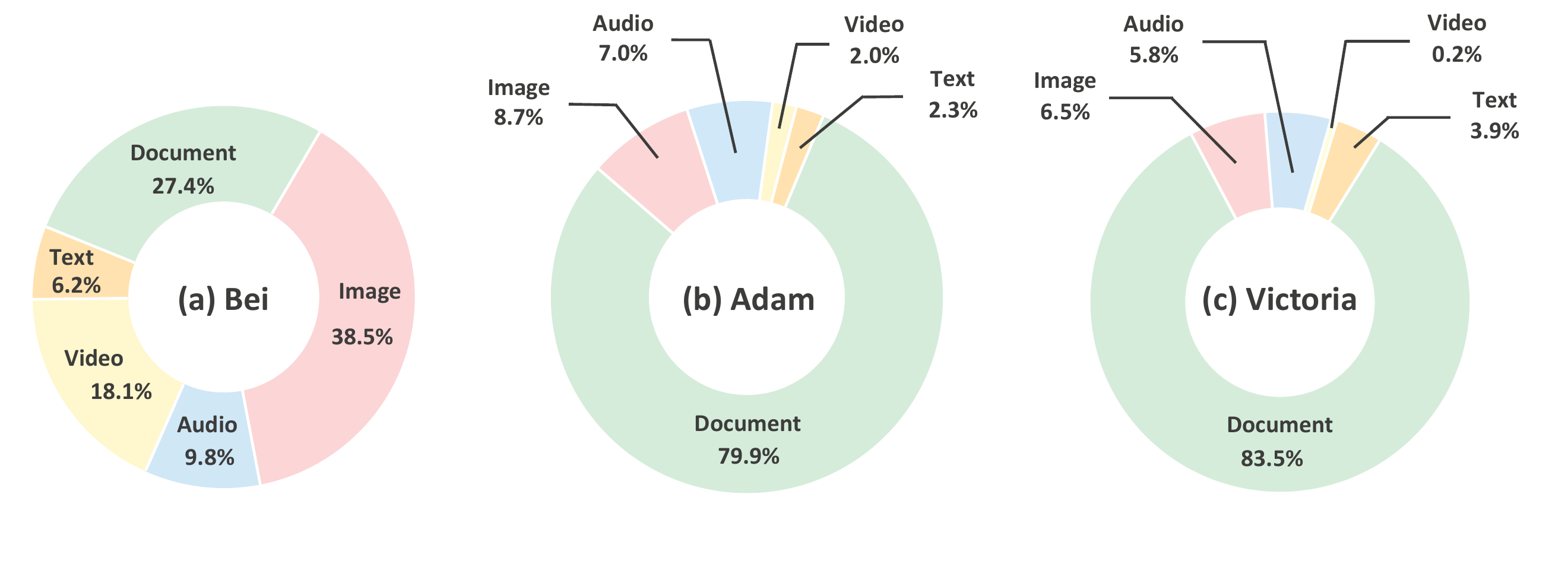}
  \caption{\textbf{Modality composition across HippoCamp profiles (by file count).} Percentage breakdown over five modalities for (a) Bei Weiwei, (b) Adam Turner, and (c) Victoria Anne Clarke.}
  \label{fig:5_modality_breakdown}
  \vspace{-10pt}
\end{figure}

\cref{fig:5_modality_breakdown} summarizes profile-specific modality composition by file count and should be read primarily as a description of evidence distribution rather than storage burden. The three archetypes are intentionally heterogeneous rather than uniformly balanced. Profile~(a) Bei Weiwei exhibits the broadest modality spread, with images (38.5\%), videos (18.1\%), and audio (9.8\%) contributing substantial evidence alongside documents (27.4\%). By contrast, Profiles~(b) Adam Turner and (c) Victoria Anne Clarke are strongly document-dominant (79.9\% and 83.5\% documents, respectively), although both retain smaller but nontrivial shares of non-document modalities. This variation matters because benchmark questions rarely depend on a single modality in isolation: broader modality spread increases pressure on multimodal perception and cross-modal grounding, whereas document-dominant regimes place more weight on selective retrieval and evidence consolidation across many related artifacts. Preserving such heterogeneous modality mixtures prevents the benchmark from collapsing into an artificially balanced setting and exposes agents to realistic shifts in evidence composition across users.

\subsubsection{Storage Footprint and File-Type Burden}

\begin{table}[ht]
\centering
\caption{\textbf{File type and storage footprint by profile.} Extension-level size statistics across Bei, Adam, and Victoria, highlighting storage asymmetry and long-tail format coverage.}
\label{tab:domain-file-stats}
\vspace{0.35em}
\scriptsize
\setlength{\tabcolsep}{2.8pt}
\renewcommand{\arraystretch}{1.04}
\begin{tabular}{@{}llrrr@{}}
\toprule
\textbf{Ext.} & \textbf{Modality} & \textbf{Bei} & \textbf{Adam} & \textbf{Victoria} \\
\midrule
mp3    & Audio     & 384.80 MB & 21.16 MB  & 1.20 GB \\
csv    & Documents & --        & 11.92 KB  & 435 B \\
docx   & Documents & 2.26 MB   & 44.98 MB  & 4.15 MB \\
eml    & Documents & 14.47 MB  & 21.84 MB  & 107.91 KB \\
ics    & Documents & 1.08 KB   & 13.28 KB  & 25.99 KB \\
pdf    & Documents & 1.08 GB   & 82.07 MB  & 960.08 MB \\
pptx   & Documents & 237.88 MB & --        & -- \\
xlsx   & Documents & --        & 53.69 KB  & 11.98 KB \\
sqlite & Documents & --        & 76.00 KB  & -- \\
gif    & Images    & 35.77 MB  & --        & -- \\
jpeg   & Images    & 207.18 MB & --        & -- \\
jpg    & Images    & 152.97 MB & --        & -- \\
png    & Images    & 94.38 MB  & 49.74 MB  & 64.89 MB \\
bin    & Text      & 4.26 MB   & --        & -- \\
ipynb  & Text      & 389.56 KB & --        & -- \\
json   & Text      & --        & --        & 6.95 KB \\
log    & Text      & 2.02 KB   & --        & -- \\
npy    & Text      & 68.89 MB  & --        & -- \\
pkl    & Text      & 1.37 KB   & --        & -- \\
pt     & Text      & 12.31 MB  & --        & -- \\
pth    & Text      & 28.55 MB  & --        & -- \\
py     & Text      & 27.95 KB  & --        & -- \\
txt    & Text      & 2.23 MB   & 57.70 KB  & 14.95 KB \\
mkv    & Video     & 7.95 GB   & 2.00 GB   & -- \\
mp4    & Video     & 27.63 GB  & 116.99 MB & 42.05 MB \\
md     & Documents & 22.71 KB  & --        & 65.56 KB \\
\bottomrule
\end{tabular}
\end{table}

\cref{tab:domain-file-stats} summarizes per-extension storage footprints and long-tail format coverage across the three profiles. Unlike \cref{fig:5_modality_breakdown}, which characterizes modality composition by file count, this table captures storage asymmetry, extension diversity, and processing burden. The key observation is that file-size footprint is not reducible to modality proportions alone: a relatively small number of large video, audio, or long-form document files can dominate storage, whereas dense collections of smaller text and document artifacts can create heavy retrieval surfaces without comparable total volume.

This distinction is visible across the three profiles. Bei contributes a disproportionately large storage footprint through high-volume video containers (\texttt{.mp4}, \texttt{.mkv}) and sizable image/audio assets, making the profile expensive to index, convert, and search over at full fidelity. Adam exhibits a more compact but extension-diverse footprint, where long-form documents (\texttt{.pdf}, \texttt{.docx}) coexist with communication and scheduling traces (\texttt{.eml}, \texttt{.ics}), increasing the need for selective retrieval across many semantically adjacent artifacts. Victoria is dominated by large PDFs and structured artifacts (e.g., \texttt{.pdf}, \texttt{.xlsx}) alongside substantial audio (\texttt{.mp3}), illustrating how storage burden can arise from heterogeneous professional and administrative materials even without extreme video volume. Overall, these statistics are most informative as a measure of retrieval cost, conversion overhead, extension-level processing complexity, and evidence-localization burden rather than as a restatement of profile semantics.

\subsubsection{Temporal Coverage of Files}
\begin{figure}[ht]
    \centering
    \includegraphics[width=0.86\linewidth]{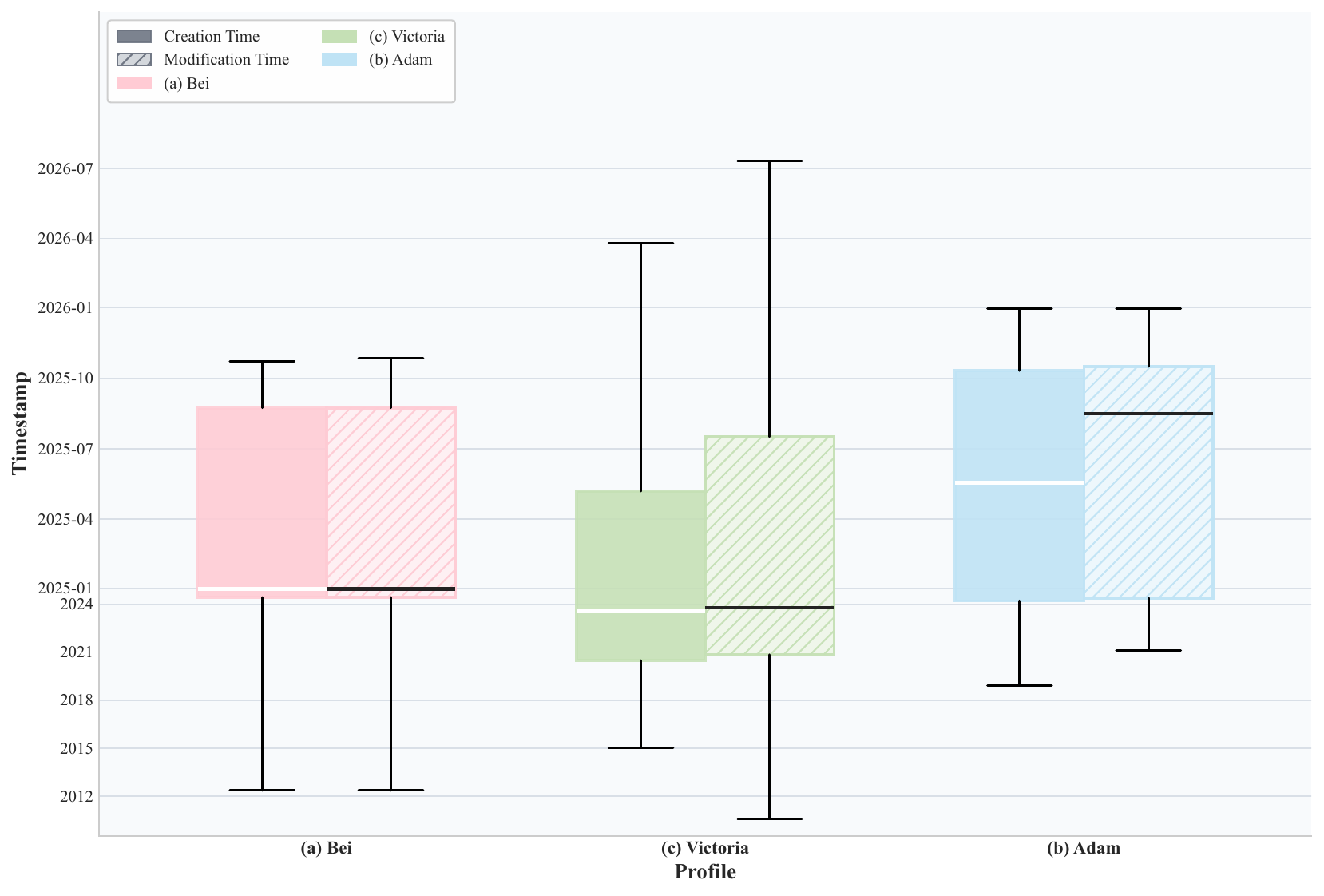}
    \caption{\textbf{Temporal coverage of file activity.} Creation and modification timestamp distributions for (a) Bei, (b) Adam, and (c) Victoria.}
    \label{fig:file-temporal-boxplot}
\end{figure}

\cref{fig:file-temporal-boxplot} summarizes the temporal footprint of the three profiles using both creation time (artifact origin) and modification time (most recent interaction), which jointly capture long-term accumulation and recent activity. Across profiles, file activity concentrates in 2024--2025 while retaining a realistic long tail of legacy artifacts extending back to 2012. This hybrid temporal structure is critical for HippoCamp: it enables (i) verification-oriented factual retention via temporal cross-checks among dated records (e.g., emails, calendars, and media timestamps), and (ii) profiling over longitudinal behavioral signals that require aggregating evidence across time rather than relying on a single snapshot. At the same time, the presence of both recent dense activity and older sparse traces prevents the benchmark from degenerating into either a purely recency-biased setting or an unrealistically archival-only corpus, thereby supporting evaluation of temporal generalization under realistic personal-device conditions.

\section{Annotation Schema and Quality Control}
\label{app:annotation_schema}

HippoCamp is annotated not as isolated QA pairs but as \emph{grounded trajectories} that couple questions and answers with file-system grounding, localized evidence, and stepwise reasoning traces; in this section, we detail the trajectory schema, evidence localization and atomic units, the human-in-the-loop QA construction pipeline, and inter-annotator agreement with quality control.

\subsection{Trajectory JSON Schema}
\label{sec:s4_trajectory_json_schema}

This section specifies the released trajectory JSON schema at a level intended to support direct parser reimplementation. Beyond documenting storage format, the schema also explains how our downstream analyses are derived, including capability-wise breakdowns, difficulty statistics, and evidence-level evaluation. In particular, the top-level record distinguishes: (i) the task specification itself, (ii) the minimal supporting file set, (iii) localized evidence supervision, (iv) stepwise rationale traces, and (v) capability labels. We additionally expose several difficulty-related attributes either as native fields or as deterministic derived quantities. A schematic overview of the hierarchical annotation schema is shown in \cref{fig:hippocamp_overview}, and we provide the detailed specification in the following subsections.

\subsection{JSON Record Overview}
\label{sec:s4_json_record_overview}

Each trajectory instance is represented as one JSON object; each released persona file stores an array of such objects. \cref{tab:json_record_overview} summarizes the top-level fields used for parsing, evaluation, and analysis. At the record level, the schema separates: (i) task specification fields, including the query and normalized answer; (ii) file-level support fields, which define the minimal supporting file set; (iii) capability annotations over search, evidence perception, and reasoning; and (iv) process supervision through localized evidence items and stepwise rationale traces.

\begin{table*}[t]
\centering
\caption{\textbf{Top-level fields in one HippoCamp trajectory record.} Stored fields are explicitly present in the released JSON.}
\label{tab:json_record_overview}
\scriptsize
\setlength{\tabcolsep}{2.8pt}
\renewcommand{\arraystretch}{1.03}
\begin{tabularx}{\textwidth}{@{}
>{\raggedright\arraybackslash}p{0.17\textwidth}
>{\raggedright\arraybackslash}p{0.13\textwidth}
>{\raggedright\arraybackslash}p{0.2\textwidth}
>{\raggedright\arraybackslash}X
>{\raggedright\arraybackslash}X
@{}}
\toprule
\textbf{Field} & \textbf{Type} & \textbf{Allowed values / format} & \textbf{Role} & \textbf{Primary use} \\
\midrule
\texttt{id} & string & String identifier & Record identifier for one QA instance. & Per-instance alignment; error analysis. \\

\texttt{question} & string & Free-form natural language & User query defining the information need. & LLM-judge input; token statistics; difficulty. \\

\texttt{answer} & string & Normalized textual answer; optional file refs in \texttt{[...]} & Gold answer target in normalized form. & LLM-judge reference; token statistics; difficulty. \\

\texttt{QA\_type} & string & \texttt{factual\_retention}, \texttt{profiling} & Top-level task family label. & Main results grouping; task-type breakdowns. \\

\texttt{profiling\_type} & string & Profiling subtype labels & Profiling subtask label when applicable. & Profiling-subtype breakdowns. \\

\texttt{data\_source} & string & Source tag from annotation ontology & Provenance/source tag for the instance. & Source audit; source-based analyses. \\

\texttt{file\_path} & array[string] & Valid file paths in the environment & Ground-truth minimal supporting file set. & File retrieval evaluation; difficulty; error analysis. \\

\texttt{file\_number} & number & Non-negative integer & Cardinality of the supporting file set. & Evidence-breadth statistics; difficulty. \\

\texttt{file\_modality} & array[string] & \texttt{document}, \texttt{video}, \texttt{audio}, \texttt{text}, \texttt{image} & Modality set of the supporting files. & Modality-breadth statistics; modality-wise analysis. \\

\texttt{file\_type} & array[string] & Supporting-file extensions & Extension set of the supporting files. & File-type analysis; difficulty. \\

\texttt{agent\_cap} & object & Capability labels over search / evidence perception / reasoning & Required capability annotations. & Capability-wise evaluation; breakdown analysis. \\

\texttt{evidence} & array[object] & Localized evidence objects & Localized answer-supporting evidence items. & Grounding evaluation; evidence statistics; evidence-level F1. \\

\texttt{rationale} & array[object] & Stepwise rationale objects & Stepwise reasoning trace linked to evidence. & Reasoning-depth analysis; process diagnosis. \\
\bottomrule
\end{tabularx}
\end{table*}

Beyond the stored fields listed above, several analysis attributes are computed deterministically from the released content. Specifically, \texttt{question\_tokens} and \texttt{answer\_tokens} are the token counts of the corresponding text fields under the tokenizer used in analysis; \texttt{evidence\_items} is defined as \texttt{len(evidence)}; and \texttt{time\_span\_days} is computed from the temporal extent of the annotated support, using the earliest and latest relevant evidence times when such timestamps are available, and defaulting to zero otherwise.

\subsection{Field Semantics and Interpretation Rules}
\label{sec:s4_field_dictionary}

We clarify several field-level conventions that are important for correctly interpreting the released schema and reproducing downstream analyses. 
The field \texttt{file\_path} records the \emph{ground-truth minimal file set} required to answer a query, rather than the full set of files traversed by annotators during data construction. 
It therefore represents the smallest annotated support set sufficient for deriving the gold answer, while \texttt{file\_number} serves as its cardinality indicator.

The fields \texttt{file\_modality} and evidence-level \texttt{modality\_type} are defined at different levels of granularity. 
\texttt{file\_modality} is a set-valued summary over the supporting files listed in \texttt{file\_path}, whereas \texttt{modality\_type} is assigned to each localized evidence item after normalization and characterizes the evidence unit itself. 
Accordingly, the former should be interpreted as a file-level aggregate, while the latter provides item-level modality information and may capture finer-grained structure than that expressed by the supporting file as a whole.

The field \texttt{answer} is defined as a \emph{normalized answer target} rather than a rationale trace. 
It is written in concise, directly judgeable natural language, may contain one or multiple sentences when required for completeness, and may optionally include file references in square-bracket form (e.g., \texttt{[meeting\_notes.pdf]}) when such references are useful for disambiguation. 
By design, however, \texttt{answer} should not contain chain-of-thought, stepwise derivation, or unnecessary explanatory narrative.

Finally, \texttt{agent\_cap} encodes the capabilities required to solve an instance, rather than model predictions or observed system behaviors. 
Within this field, \texttt{agent\_cap.reasoning} follows a constrained label system: \texttt{basic} denotes the minimal reasoning regime and is mutually exclusive with all stronger reasoning categories, whereas \texttt{computation}, \texttt{verification}, and \texttt{summarization} may be assigned individually or jointly. 
Thus, the presence of \texttt{basic} precludes the co-occurrence of any higher-order reasoning label, while any instance requiring one or more higher-order reasoning operations must exclude \texttt{basic}.

\subsection{Evidence Object Specification}
\label{sec:s4_evidence_object_spec}

Each entry in \texttt{evidence[]} is a localized support object intended to be directly checkable against the source artifact. At minimum, an evidence object contains an \texttt{evidence\_id}, a \texttt{file\_path}, a normalized item-level \texttt{modality\_type}, an \texttt{evidence\_text} field, and one or more localization entries under \texttt{evidence\_locator}.

The field \texttt{evidence\_id} is unique within a single trajectory record and serves as the key used by rationale steps to cite supporting evidence. It is not required to be globally unique across the entire benchmark release. The field \texttt{evidence\_text} is intended to preserve answer-supporting content in a verification-friendly form rather than to summarize the whole file. For text or document evidence, it is typically a direct excerpt. For audio or video, it is a transcription or tightly faithful rendering of the localized segment. For image-like evidence, it is a concise content description anchored to the localized region rather than a free-form interpretation.

Localization is stored in \texttt{evidence\_locator}. Conceptually, each locator is represented as a \texttt{\{unit, position\}} pair, where \texttt{unit} defines the measurement basis and \texttt{position} specifies the actual range or index. Typical document evidence uses page-based units, while temporal media uses timestamp-based ranges. For page-based documents, the position may include both \texttt{system\_page} and \texttt{printed\_page}; for audio or video, the position is typically a start--end timestamp range.

\subsection{Rationale Trace Specification}
\label{sec:s4_rationale_trace_spec}

Each entry in \texttt{rationale[]} represents one annotated reasoning step. A rationale step minimally contains a \texttt{step\_id}, a natural-language \texttt{rationale\_text}, and an \texttt{evidence\_id} list pointing to the evidence items that ground the step. The \texttt{step\_id} is unique within the enclosing rationale trace; \texttt{rationale\_text} describes the intermediate operation, observation, or conclusion at that step; and \texttt{evidence\_id} provides explicit grounding links back to localized support.

The released rationale traces follow a normalized three-stage structure. The first stage is \textbf{planning}, which decomposes the task into subgoals or search intentions. The second stage is \textbf{navigation + reading}, which covers file discovery, region localization, and evidence acquisition. The third stage is \textbf{integration + verification}, which synthesizes evidence across sources, performs any necessary checking or computation, and assembles the final grounded answer. These stages are conceptual normalization categories rather than rigid user-visible tags, but they provide a consistent process template for reasoning analysis.

Groundedness is enforced at the step level. We allow \texttt{evidence\_id = []} only for abstract planning steps that frame the task before evidence has been located. Any step that reports file content, extracts factual support, performs cross-source integration, verifies a claim, executes a computation based on source material, or contributes directly to the final answer must cite one or more valid evidence IDs. In other words, empty evidence references are permitted only for high-level planning, not for reading, integration, verification, or answer-bearing steps.

\subsection{Schema Validation Rules}
\label{sec:s4_schema_validation_rules}

To ensure reproducibility, we apply a lightweight schema-validation protocol to the released records. The purpose of this protocol is not to enforce an overly restrictive normalization layer, but to guarantee that the released JSON objects remain consistently parseable and that the bookkeeping underlying downstream evaluation is stable and reproducible across implementations.

Our validation procedure checks the following conditions:

\begin{itemize}
    \item \textbf{Evidence-reference consistency.} Every non-empty \texttt{evidence\_id} referenced by a rationale step must resolve to a valid \texttt{evidence\_id} defined in the corresponding \texttt{evidence[]} array of the same trajectory record.
    \item \textbf{File-system-resolvable support paths.} Every top-level \texttt{file\_path} entry, together with every evidence-level \texttt{file\_path}, must be resolvable to an existing file in the released file system.
    \item \textbf{Locator-range validity.} Each evidence locator must be well-formed under its declared unit. For page-based evidence, page indices must lie within the bounds of the corresponding document; for timestamp-based evidence, start and end times must define a valid interval within the duration of the associated media file.
    \item \textbf{Answer--evidence support consistency.} The annotated answer must be supportable by the linked evidence set. We assess this through a combination of manual audit and script-level sanity checks over evidence references and their localized contents.
\end{itemize}

Given these constraints, a parser can reconstruct a trajectory record by reading the top-level task and support fields, resolving \texttt{evidence[]} into an evidence map indexed by \texttt{evidence\_id}, grounding each rationale step through its cited evidence references, and verifying the resulting structure against the validation rules above. This procedure is sufficient to reproduce the bookkeeping required for capability-wise analysis, difficulty computation, and evidence-level matching metrics.

\begin{figure}[ht]
\centering
\fbox{%
\begin{minipage}{0.965\textwidth}
\vspace{0.25em}
\centering {\textbf{Illustrative JSON Record with Multimodal Evidence Variants}}\par
\vspace{0.35em}
\hrule
\vspace{0.55em}

\hspace{0.8em}
\begin{minipage}[t]{0.445\textwidth}
\tiny
\ttfamily
\raggedright

"id": "92", \\

"question": "What is the exact timestamp of the first food close-up after the fairy waves her wand?", \\

"answer": "The first food close-up appears at 00:01:24.", \\[0.35em]

"QA\_type": "factual\_retention", \\

"data\_source": "Mixed-Multimodal", \\[0.35em]

"file\_path": [ \\
\hspace*{1.2em}"docs/case\_notes.pdf", \\
\hspace*{1.2em}"slides/briefing.pptx", \\
\hspace*{1.2em}"tables/ssic\_code.xlsx", \\
\hspace*{1.2em}"video/zootopia.mp4" \\
], \\

"file\_number": 4, \\

"file\_modality": ["document", "video"], \\

"file\_type": ["pdf", "pptx", "xlsx", "mp4"], \\[0.45em]

"agent\_cap": \{ \\
\hspace*{1.2em}"search": ["semantic", "metadata"], \\
\hspace*{1.2em}"evidence\_perception": [ \\
\hspace*{2.4em}"document\_reading", \\
\hspace*{2.4em}"slide\_reading", \\
\hspace*{2.4em}"spreadsheet\_reading", \\
\hspace*{2.4em}"video\_grounding" \\
\hspace*{1.2em}], \\
\hspace*{1.2em}"reasoning": ["verification", "summarization"] \\
\}, \\[0.45em]

"rationale": [ \\
\hspace*{1.2em}\{ \\
\hspace*{2.4em}"step\_id": "1", \\
\hspace*{2.4em}"rationale\_text": "Identify candidate files and collect localized evidence.", \\
\hspace*{2.4em}"evidence\_id": ["1"] \\
\hspace*{1.2em}\}, \\[0.2em]
\hspace*{1.2em}\{ \\
\hspace*{2.4em}"step\_id": "2", \\
\hspace*{2.4em}"rationale\_text": "Verify the retrieved evidence and synthesize the answer.", \\
\hspace*{2.4em}"evidence\_id": ["1", "3"] \\
\hspace*{1.2em}\}, \\
\hspace*{1.2em}\{ \\
\hspace*{2.4em}"step\_id": "3", \\
\hspace*{2.4em}"rationale\_text": "Verify the retrieved evidence and synthesize the answer.", \\
\hspace*{2.4em}"evidence\_id": ["2", "4"] \\
\hspace*{1.2em}\} \\
],
\end{minipage}
\hfill
\begin{minipage}[t]{0.475\textwidth}
\tiny
\ttfamily
\raggedright

"evidence": [ \\

\hspace*{1.2em}\{ \\
\hspace*{2.4em}"evidence\_id": "1", \\
\hspace*{2.4em}"file\_path": "docs/case\_notes.pdf", \\
\hspace*{2.4em}"modality\_type": "document", \\
\hspace*{2.4em}"evidence\_text": "The hearing is scheduled for February 14, 2026.", \\
\hspace*{2.4em}"evidence\_locator": [\{ \\
\hspace*{3.6em}"unit": "page", \\
\hspace*{3.6em}"position": \{"system\_page": 3, "printed\_page": 2\} \\
\hspace*{2.4em}\}] \\
\hspace*{1.2em}\}, \\[0.2em]

\hspace*{1.2em}\{ \\
\hspace*{2.4em}"evidence\_id": "2", \\
\hspace*{2.4em}"file\_path": "slides/briefing.pptx", \\
\hspace*{2.4em}"modality\_type": "document", \\
\hspace*{2.4em}"evidence\_text": "UN Peacekeeping: First Combat Troops to Mali.", \\
\hspace*{2.4em}"evidence\_locator": [\{ \\
\hspace*{3.6em}"unit": "slide", \\
\hspace*{3.6em}"position": \{"system\_page": 23, "printed\_page": 23\} \\
\hspace*{2.4em}\}] \\
\hspace*{1.2em}\}, \\[0.2em]

\hspace*{1.2em}\{ \\
\hspace*{2.4em}"evidence\_id": "3", \\
\hspace*{2.4em}"file\_path": "tables/ssic\_code.xlsx", \\
\hspace*{2.4em}"modality\_type": "document", \\
\hspace*{2.4em}"evidence\_text": "UEN 202227796H has secondary SSIC 46100.", \\
\hspace*{2.4em}"evidence\_locator": [\{ \\
\hspace*{3.6em}"unit": "sheet", \\
\hspace*{3.6em}"position": \{"sheet\_name": "Company", "row\_number": 9\} \\
\hspace*{2.4em}\}] \\
\hspace*{1.2em}\}, \\[0.2em]

\hspace*{1.2em}\{ \\
\hspace*{2.4em}"evidence\_id": "4", \\
\hspace*{2.4em}"file\_path": "video/zootopia.mp4", \\
\hspace*{2.4em}"modality\_type": "video", \\
\hspace*{2.4em}"evidence\_text": "First food close-up appears after the wand wave.", \\
\hspace*{2.4em}"evidence\_locator": [\{ \\
\hspace*{3.6em}"unit": "timestamp", \\
\hspace*{3.6em}"position": "00:01:24--00:01:26" \\
\hspace*{2.4em}\}] \\
\hspace*{1.2em}\} \\

]
\end{minipage}

\vspace{0.25em}
\end{minipage}%
}
\caption{\textbf{Illustrative JSON schema example following the released HippoCamp format.} The example shows top-level metadata, multimodal evidence objects, and evidence-linked rationale steps.}
\label{fig:json_record_example}
\end{figure}

\subsection{Designing Atomic Units for Evidence}
\label{sec:s4_au}

\subsubsection{Motivation}
A central challenge in personal-file QA is that evidence is naturally localized at \emph{different granularities} across modalities. For example, documents are typically referenced at the page level, audio and video at the timestamp level, while images and embedded visual regions may require spatial localization. Without a modality-aligned notion of evidence granularity, evaluation becomes difficult to compare across modalities and may overestimate retrieval quality when a method identifies the correct file but fails to localize the decisive evidence. To address this issue, we introduce \emph{atomic units} (AUs), a modality-normalized abstraction of the smallest evidence-bearing region used primarily for fine-grained grounding analysis and error diagnosis. Intuitively, AUs provide a diagnostic interface for asking not only whether an agent found the right file, but also whether it grounded its answer in the right part of that file.

\subsubsection{Definition}
We define an atomic unit as the minimal modality-aligned segment at which evidence can be localized and, in principle, independently verified. AUs do not replace the human-readable \texttt{evidence\_locator} field in the released JSON; rather, they normalize heterogeneous locator types into a common analytical representation. In the current release, the schema explicitly instantiates page-level and timestamp-level locators, while the AU formulation additionally specifies how finer-grained spatial/structural regions can be represented when needed for diagnosis or future extensions.

\begin{table}[ht]
\centering
\caption{\textbf{Atomic-unit (AU) definition by modality.} AUs normalize heterogeneous evidence locators into modality-aligned minimal units.}
\label{tab:au_definition}
\small
\setlength{\tabcolsep}{3pt}
\renewcommand{\arraystretch}{1.10}
\begin{tabularx}{\linewidth}{@{}
>{\raggedright\arraybackslash}p{0.16\linewidth}
>{\raggedright\arraybackslash}p{0.26\linewidth}
>{\raggedright\arraybackslash}p{0.30\linewidth}
>{\raggedright\arraybackslash}X
@{}}
\toprule
\textbf{Modality} & \textbf{AU unit} & \textbf{Typical resolution} & \textbf{Example} \\
\midrule
Text & token span / sentence span & contiguous text span & a clause in a note or email body \\
Document & page-equivalent; optional region & page; optional table/figure region & page 4 of a PDF; a row in a regulatory table \\
Image & spatial patch / region & patch grid or bounding box & the face region or background region in a photo \\
Video & frame or temporal chunk & frame or timestamp window & 00:07--00:08 in an advertisement clip \\
Audio & temporal segment & timestamp window / frame span & 00:42:56--00:43:26 in a recording \\
\bottomrule
\end{tabularx}
\end{table}
\subsection{AU Generation Procedures}
AUs are generated by modality-specific procedures that trade off precision and reproducibility.

\paragraph{Text.}
For free-form text, we treat contiguous token or sentence spans as AUs. In the released data, evidence is stored as \texttt{evidence\_text}; AU mapping is obtained by aligning the annotated span back to the corresponding textual region in the source file.

\paragraph{Documents.}
For paginated documents, the default AU is a \emph{page-equivalent} unit, since page indices are the native locators recorded in \texttt{evidence\_locator}. When finer structure is required (e.g., a table row, a figure caption, or an embedded visual region), the page can be further refined into document regions using page coordinates and optional table-cell or figure-region subdivision. This page-first design matches the current release while remaining extensible to more granular document grounding.

\paragraph{Images.}
For images, AUs are defined over a spatial coordinate system. A practical implementation is a regular patch grid (e.g., $16\times16$ patches over the image plane), with optional aggregation into bounding boxes when a human-meaningful region is available. In the current release, image evidence is often represented textually via \texttt{evidence\_text}; AU assignment therefore functions primarily as an analysis-time abstraction unless explicit spatial regions are additionally annotated.

\paragraph{Video.}
For video, AUs are temporal chunks or frames derived from a fixed sampling rule. A reproducible default is uniform frame sampling at a predefined frame rate (e.g., 1 fps for analysis) or timestamp-window segmentation with a fixed stride. The current release uses timestamp-based locators, so the AU mapping is directly induced by the referenced temporal interval.

\paragraph{Audio.}
For audio, AUs are temporal segments induced by timestamp windows. A standard realization uses fixed-length frames or short segments (e.g., analysis windows with constant hop size), but the released benchmark records evidence at the timestamp level; the corresponding AU is therefore the segment spanned by the annotated time interval.

\subsection{AU--Evidence Mapping}
Each evidence item in HippoCamp contains a human-readable \texttt{evidence\_locator}, and AU mapping deterministically converts that locator into one or more modality-aligned units:
\begin{itemize}
    \item \textbf{Document page locator} $\rightarrow$ page-equivalent AU;
    \item \textbf{Document region / table / figure locator} $\rightarrow$ region-level AU within the page;
    \item \textbf{Image region} $\rightarrow$ spatial patch set or bounding-box AU;
    \item \textbf{Timestamp or timestamp range} $\rightarrow$ audio/video temporal AU span.
\end{itemize}
To support robust evaluation, AU matching is defined with tolerance where appropriate. For temporal media, matching can allow a small timestamp slack around the annotated interval; for spatial regions, overlap can be measured by standard region criteria (e.g., IoU-style overlap when bounding boxes are available). These tolerances ensure that AU-level grounding remains reproducible without being brittle to negligible alignment differences.

\subsection{Usage in Evaluation, Diagnosis, and Training}
AUs serve three purposes in HippoCamp.

\paragraph{Evaluation.}
At evaluation time, AU-normalized grounding provides a modality-consistent way to assess whether a method has localized the decisive evidence rather than merely retrieving the correct file. This is especially important for multimodal files where file-level hits can be misleading. AU-aware comparison therefore serves as a diagnostic complement to file-level retrieval metrics, providing finer-grained evidence overlap and grounding analysis where localization quality is important.

\paragraph{Diagnosis.}
At diagnosis time, AUs help distinguish between different failure modes. For example, a model may identify the correct file but ground on the wrong page, wrong timestamp, or wrong visual region. Such errors would be invisible under file-level scoring alone but are exposed by AU-level analysis. In this sense, AUs are introduced primarily to support fine-grained perception and grounding error analysis rather than to define a standalone headline score.

\paragraph{Training and release considerations.}
In the current benchmark release, AUs primarily support evaluation and diagnosis rather than being exposed as a standalone training supervision target. Accordingly, AU-level signals are intended mainly for fine-grained analysis and diagnostic breakdowns, while the benchmark’s primary reported retrieval metrics remain at the file level. This design is motivated by privacy, copyright, and annotation-cost considerations, especially for spatially localized personal media. At the same time, the AU formulation remains compatible with future privacy-preserving derived supervision schemes, such as masked crops, region descriptors, or AU-only representations that preserve grounding structure without exposing raw personal content.

\subsection{Human-in-the-loop QA Construction}
\label{sec:s4_human_loop}

As outlined in \cref{fig:bench_stat}, HippoCamp uses a human-in-the-loop QA construction pipeline to transform candidate information needs into finalized, trajectory-annotated benchmark instances. This pipeline proceeds in five stages: two-source question proposal, candidate consolidation, de-duplication and coverage balancing, trajectory structuring, and bounded model assistance. The role of this section is to describe how benchmark items are constructed; agreement, adjudication, and release-time quality control are deferred to Appendix~\ref{sec:s4_iaa_qc}.

\subsection{Two-Source Question Proposal}
\label{sec:s4_two_source}

\paragraph{Manual proposals.}
The primary source of HippoCamp questions is manual authoring by participants and expert annotators who are familiar with the profile-specific file systems. These questions are explicitly \emph{user-driven}: they originate from concrete information needs that a participant could plausibly encounter in day-to-day personal computing, such as recalling a past fact, reconstructing a workflow, summarizing a prior episode, or planning under current constraints. Because contributors understand their own habits, organization strategies, and recurring tasks, manual proposals capture realistic intents that are difficult to recover from files alone.

\paragraph{Synthetic proposals.}
To complement manual authoring, we use LLMs to generate candidate questions conditioned on restricted contextual metadata, such as file paths, timestamps, directory structure, and selected seed examples. The role of synthetic proposals is not to define benchmark semantics, but to improve \emph{coverage} along underrepresented dimensions, including modality combinations, evidence-set size, and long-tail task patterns. These proposals are therefore treated strictly as candidates and are passed to the consolidation stage for review, revision, merging, or rejection.

\subsection{Candidate Consolidation and Screening}
\label{sec:s4_judging}

All manually authored and LLM-suggested candidates are consolidated through a human screening stage. At this stage, annotators retain only candidates that satisfy four construction-time requirements: (i) \textbf{groundedness}, meaning that a plausible answer can be supported by actual files in the profile; (ii) \textbf{non-triviality}, meaning that the query requires meaningful retrieval, perception, or reasoning rather than direct filename or metadata lookup; (iii) \textbf{low redundancy}, meaning that the candidate does not duplicate existing items at the intent or reasoning-pattern level; and (iv) \textbf{privacy safety}, meaning that the wording and anticipated evidence requirements do not reintroduce sensitive information beyond the anonymized profile. The retained set forms the candidate pool passed to de-duplication, balancing, and trajectory structuring; final release-time acceptance criteria are described separately in Appendix~\ref{sec:s4_qc_checks}.

\subsection{De-duplication and Coverage Constraints}
\label{sec:s4_dedup}

To avoid over-concentration in a small number of query forms, we apply de-duplication and balancing at two levels.

\paragraph{Intent-level de-duplication.}
Questions that express the same underlying information need with only superficial rewording are merged, and only one representative formulation is kept.

\paragraph{Pattern-level de-duplication.}
We also suppress repeated questions that rely on essentially the same solution structure or evidence configuration, even when the wording differs. For example, two questions requiring the same file set, the same cross-file joins, and the same reasoning pattern are not treated as distinct items.

\paragraph{Coverage balancing.}
After de-duplication, we rebalance the retained candidate set to improve coverage over modality combinations, evidence-set sizes, and task families. In particular, we monitor the relative representation of \textit{factual retention} versus \textit{profiling}, as well as the breadth of multimodal and multi-file instances, so that the final benchmark contains both common personal-computing questions and harder long-tail cases.

\subsection{Trajectory Structuring}
\label{sec:s4_structuring}

For each retained QA candidate, human annotators construct the grounded trajectory record used for benchmark release. This record includes (i) a minimal supporting file set and file-level metadata, (ii) localized evidence objects with explicit locators, (iii) a stepwise rationale trace, and (iv) capability labels spanning search, evidence perception, and reasoning. The objective of this stage is to convert an accepted question into a compact, schema-compliant support structure that can be used for evaluation and downstream diagnosis.

A key design principle is the \textbf{minimalist gold trajectory}. The goal is not to enumerate every plausible solution path, but to record the smallest set of evidence nodes and reasoning transitions sufficient to justify the answer. The resulting trajectory therefore defines a compact support structure for evaluation while leaving room for agents to discover longer or alternative valid paths during inference.

\subsection{Model Assistance Protocol}
\label{sec:s4_model_assist}

To keep LLM assistance reproducible and bounded, we restrict the information visible to the model. LLMs are allowed to observe only limited contextual metadata and selected seed examples, including file paths, timestamps, and directory hierarchy cues; they do \emph{not} receive unrestricted access to the full file system, raw personal identifiers, or any external knowledge source. Models are explicitly instructed not to invent PII, not to introduce unsupported facts, and not to propose questions whose answers depend on information outside the provided corpus. The typical output is either a candidate question or a lightweight JSON skeleton that is subsequently edited and completed by human annotators. Under this protocol, the model is used only for bounded proposal or lightweight structuring support; all finalized benchmark content remains subject to human revision and approval.

\subsection{Prompt Families for LLM-assisted Proposal}
\label{sec:s4_prompt_families}

We use a small family of prompt templates to support bounded LLM-assisted proposal generation under different task families and profile contexts. These prompts are implementation details of the proposal stage rather than part of the benchmark definition itself: all outputs remain provisional and are subsequently screened, revised, or discarded by human annotators. Across all prompt families, the model only observes a curated local batch of related files together with limited contextual metadata and seed examples. It is explicitly prohibited from introducing external knowledge, unreleased personal identifiers, or filename/path-based shortcuts, and all retained items undergo human review and trajectory structuring before inclusion. The prompt families vary mainly along two axes. First, they differ by \emph{task family}: factual-retention prompts emphasize explicit fact extraction and verification from bounded evidence, whereas profiling prompts emphasize cross-file aggregation, temporal regularity, and user-level synthesis. Second, they differ by \emph{profile context}: prompts for Bei emphasize academic, creative, and lifestyle traces; prompts for Adam emphasize legal workflow, professional correspondence, and structured routines; and prompts for Victoria emphasize financial analysis, reporting cycles, and numerically grounded research behavior. These profile-specific constraints help ensure that LLM proposals remain aligned with the intended personal-computing environment rather than drifting toward generic QA.

\paragraph{Representative prompt example.}
Below we show a representative condensed prompt used for College Profiling. It conditions the model on an anonymized persona description, a curated local file batch, a profiling subtask definition, and several seed questions, and asks for candidate profiling questions that require grounded cross-file or cross-time synthesis.

\begin{figure}[ht]
\centering
\fbox{%
\begin{minipage}{0.965\textwidth}
\vspace{0.25em}
\centering {\textbf{Representative Prompt: College Profiling (Bei Weiwei)}}\par
\vspace{0.35em}
\hrule
\vspace{0.55em}

\hspace{0.8em}
\begin{minipage}[t]{0.94\textwidth}
\tiny
\ttfamily
\raggedright
Generate candidate profiling questions for the HippoCamp benchmark.\\[0.3em]

\textbf{Persona.}\\
Bei Weiwei is a graduate student whose digital ecosystem includes coursework materials, tutoring records, schedules, travel plans, creative media projects, and personal notes. Her behavior exhibits repeated study routines, tutoring workflows, and seasonal content-creation patterns.\\[0.3em]

\textbf{Inputs.}\\
You will be given:\\
- a small batch of related files curated by a human annotator,\\
- a profiling subtask label and definition,\\
- several seed questions illustrating the desired style.\\[0.3em]

\textbf{Task.}\\
Propose natural profiling questions that:\\
- require synthesizing information across multiple files or time periods,\\
- reflect habits, routines, preferences, or long-term patterns,\\
- align with the provided profiling subtask,\\
- do not reference filenames, paths, or system metadata,\\
- remain grounded strictly in the provided files.\\[0.3em]

\textbf{Output.}\\
Produce a small set of candidate profiling questions with brief answer sketches and indicative evidence hints. These outputs are proposals only and will be reviewed and rewritten by human annotators.
\end{minipage}

\vspace{0.45em}
\hrule
\vspace{0.2em}
\end{minipage}%
}
\caption{\textbf{Representative prompt for LLM-assisted proposal generation.} A condensed College-Profiling prompt conditioned on persona description, local file batch, profiling subtask, and seed examples. The model proposes candidate questions only; all retained items are subsequently reviewed and restructured by human annotators.}
\label{fig:prompt_example_college_profiling}
\end{figure}

The remaining prompt families follow the same overall structure but differ in domain-specific constraints. Adam-oriented prompts emphasize legal workflow, correspondence, and verification-heavy assistance, whereas Victoria-oriented prompts emphasize analytical routines, reporting cycles, and numerically grounded task structure. In the most sensitive financial cases, human annotators author the question, answer, and evidence directly, and the model is used only for schema normalization or rationale drafting. Across all prompt families, LLM outputs function as bounded proposals rather than authoritative annotations.

\subsection{Inter-Annotator Agreement and Quality Control}
\label{sec:s4_iaa_qc}

This section describes the agreement, adjudication, and release-time quality-control procedures applied after candidate construction and trajectory authoring. The focus here is not how benchmark items are proposed or structured, but how accepted records are reviewed for consistency, privacy safety, grounding fidelity, and schema compliance before release.

\subsubsection{Annotator Setup and Sampling Protocol}
\label{sec:s4_iaa_setup}

HippoCamp is annotated by domain-aware human annotators drawn from the contributor groups underlying the three archetypal profiles. This dual-role setup is deliberate: because annotators are intimately familiar with the organizational logic, recurring workflows, and realistic information needs of their own digital environments, they can formulate and verify grounded tasks with profile-specific contextual knowledge that would be difficult to recover through external annotation alone. To mitigate profile-specific bias, however, all records are subsequently subjected to cross-checking, secondary review, and adjudication.

For agreement and validation analysis, we adopt a stratified sampling protocol spanning all three profiles, both task families (\textit{factual retention} and \textit{profiling}), diverse modality configurations, and multiple difficulty bands. This ensures that reliability is assessed not only on simple text-dominant cases, but also on multi-file, multimodal, and long-horizon instances that are most diagnostic for agent evaluation.

\subsubsection{Unified Review Protocol}
\label{sec:s4_protocols}

All annotators follow a unified review protocol during secondary review and release preparation. For each accepted record, reviewers verify that the question is interpretable under the released anonymized environment, that the cited supporting files and evidence locators are sufficient to justify the gold answer, and that capability labels, task labels, and rationale fields remain internally consistent. Reviewers do not seek to enumerate all possible valid reasoning paths; instead, they verify that the released record is coherent, minimally sufficient, and schema-compliant.

For every accepted question, annotators identify the shortest plausible solution path that uses only the files and evidence spans strictly necessary to justify the gold answer. The resulting trajectory records the most direct, human-plausible reasoning route a domain expert would take, including only essential waypoints such as key files, relevant pages or timestamps, and minimal supporting evidence. This minimalist design preserves trajectory diversity during evaluation: agents may follow richer or longer valid paths, but any successful solution must recover at least the essential evidence nodes recorded in the gold trace. The released trajectory should therefore be interpreted as a compact support structure rather than an exhaustive execution trace.

\subsubsection{Adjudication Protocol}
\label{sec:s4_adjudication}

Annotation disagreements are resolved through adjudication rather than automatic majority voting. When two annotations conflict, the item is forwarded to an additional reviewer, who re-examines the source files, evidence links, and task specification. If the disagreement reflects ambiguity in the underlying files, annotators resolve it through focused review and, when necessary, revise the evidence spans or question wording to restore interpretability and grounding. The final released record is therefore an \emph{adjudicated gold annotation}, not a raw vote aggregate.

Adjudication is especially important for profiling tasks, where disagreements more often arise from longitudinal interpretation rather than explicit fact extraction, and for specialized legal or financial items, where superficially plausible annotations may still be unacceptable if they omit a controlling clause, numerical constraint, or domain-specific exception.

\subsubsection{Quality Control and Automated Sanity Checks}
\label{sec:s4_qc_checks}

We adopt a multi-stage release-time quality-control pipeline that combines rule-based checks, human review, persona consistency validation, and dataset-level balancing.

\paragraph{Acceptance criteria.}
The following acceptance criteria apply to records considered for final release, rather than to the earlier candidate-screening stage described in Appendix~\ref{sec:s4_judging}. Each candidate instance must satisfy five criteria:
(i) \textbf{groundedness}, meaning that the answer is fully supported by the provided files and does not rely on unverifiable external knowledge;
(ii) \textbf{clarity}, meaning that the question is interpretable to a third party with access only to the released files;
(iii) \textbf{task realism}, meaning that it reflects a plausible information need for the corresponding persona;
(iv) \textbf{non-triviality}, meaning that it cannot be solved by superficial metadata lookup or trivial copying; and
(v) \textbf{privacy safety}, meaning that no textual component reintroduces identifiers beyond the approved anonymized profile.

\paragraph{Automated validation.}
In addition to human review, we run automated sanity checks over all records. These checks verify file existence and path validity, locator validity (e.g., page ranges and timestamp bounds), closure of \texttt{evidence\_id} references, near-duplicate question detection via similarity thresholds, and a second privacy audit based on rule-based scans for names, emails, phone numbers, IDs, and other sensitive identifiers.

\paragraph{Filtering outcomes.}
Based on this pipeline, each candidate is either accepted, revised, or rejected. Accepted items satisfy all criteria; revised items contain correctable issues such as ambiguous wording or incomplete grounding; rejected items contain structural flaws requiring substantive rewriting, including hallucinated evidence, privacy leakage, or excessive redundancy.

\subsubsection{Post-hoc Audit and Final Validation}
\label{sec:s4_posthoc}

After annotation and adjudication, we perform an additional stratified human audit across profile, task type, modality composition, and difficulty level. The purpose of this audit is to verify that the released benchmark remains balanced and coherent under a fresh review pass and to identify any residual annotation artifacts that escaped earlier filtering. When failures are identified, the corresponding records are revised and versioned so that all post-hoc corrections remain traceable.

Finally, before release, all accepted records must pass schema-level validation: evidence spans must be correctly referenced, rationale steps must conform to the trajectory specification, answer format must match the intended task, and task/profiling labels must be internally consistent. Together, these procedures ensure that released HippoCamp records remain auditable, internally coherent, and suitable for fine-grained evaluation of long-horizon multimodal agents.

\section{Tasks and Difficulty}
\label{app:tasks_difficulty}
\subsection{Task Taxonomy}
\label{sec:sm_taxonomy}

This appendix expands the two task families introduced in the main text---\textbf{Factual Retention} and \textbf{Profiling}---by detailing their recurrent subtypes and providing representative grounded examples. Rather than restating the benchmark-level framing, we focus here on the distinct evidence structures, abstraction patterns, and representative case forms associated with each task family.

\subsubsection{Factual Retention}
\paragraph{Motivation.}
Factual-retention queries require agents to recover precise, verifiable facts from a device-resident corpus under realistic file-system ``haystack'' conditions. In Appendix~C.1, we focus on the main evidence regimes these queries instantiate and on representative grounded examples, rather than repeating the task-family motivation given in the main text.

\paragraph{Definition.}
Factual retention covers several recurring evidence regimes, including (i) \emph{atomic fact retrieval} (e.g., dates, quantities, entity attributes), (ii) \emph{document-level localization} (e.g., identifying the correct file, path, or storage location), (iii) \emph{temporal or comparative fact recovery} (e.g., trends or changes across dated records), and (iv) \emph{normative clause extraction} (e.g., obligations, permissions, or exceptions stated in contracts and policies). The common requirement across these cases is that the answer remain fully traceable to explicit file-grounded evidence.

\begin{figure*}[ht]
    \centering
    \includegraphics[width=\linewidth]{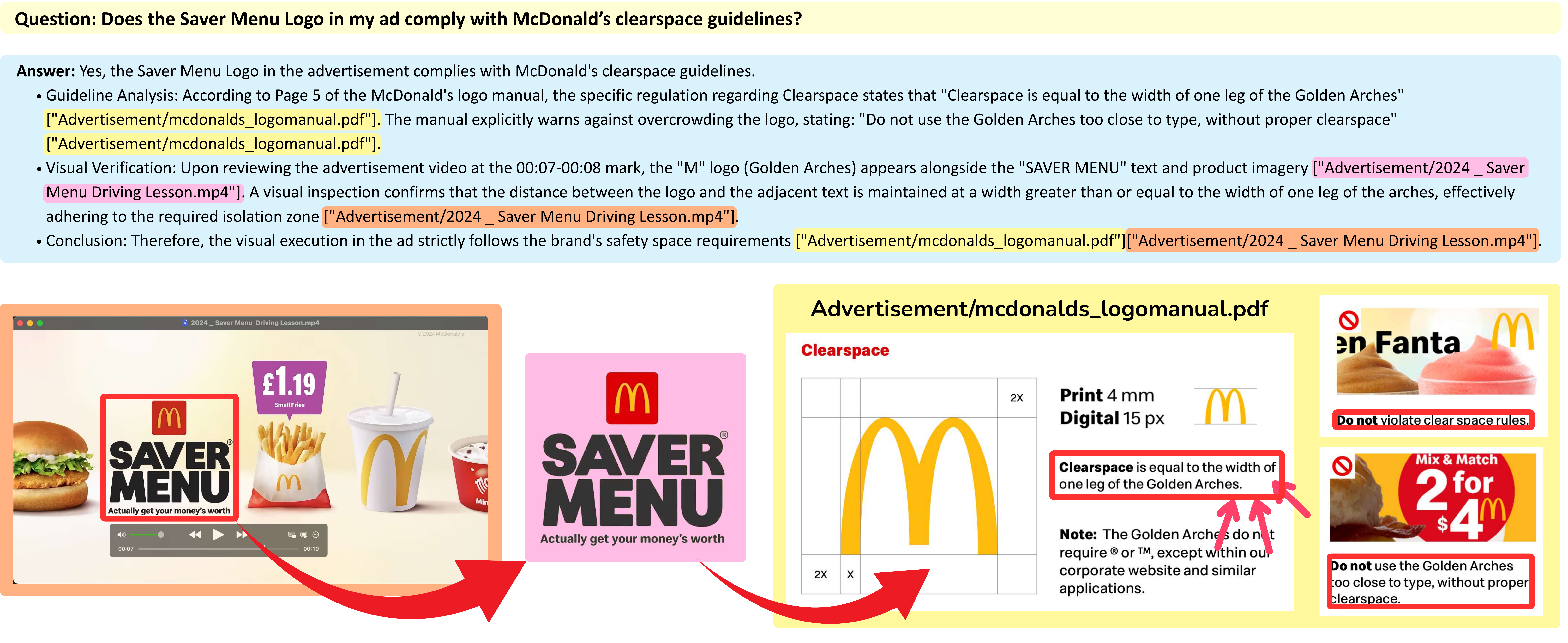}
    \caption{\textbf{Factual retention example (document--video compliance).} The agent verifies logo clearspace by extracting the rule from the manual and checking it against video frames, with ground-truth answer and evidence visualizations.}
    \label{fig:Factual_Retention_2}
  \end{figure*}

\paragraph{Illustrative examples.}
\cref{fig:Factual_Retention_1,fig:Factual_Retention_2} show two representative factual-retention instances with grounded answers, supporting file lists, and evidence visualizations.

\paragraph{Example 1: Cross-modal asset retrieval.}

\begin{figure*}[ht]
    \centering
    \includegraphics[width=0.9\linewidth]{figs/7_Factual_Retention.pdf}
    \caption{\textbf{Factual retention example (cross-modal asset retrieval).} Given a vlog script, the agent extracts the required assets and identifies the matching photos, with ground-truth answer and evidence visualizations.}
    \label{fig:Factual_Retention_1}
  \end{figure*}

The first example asks the agent to locate a previously written vlog script and identify the corresponding cat photos required by the script. The ground-truth answer is supported by a small set of concrete evidence files: the script document \texttt{Vlog\ Script\_...\_Shadow\ Friend.docx} and four images in \texttt{Cat-Vlog/images/} that match the script's \emph{Required Assets} specifications. The evidence visualization highlights two distinct alignment operations: (i) \emph{document parsing} to extract structured requirements (e.g., ``two silhouette photos'' and ``two warm lamp photos'') from the script, and (ii) \emph{visual grounding} to verify that selected images satisfy the described attributes (e.g., backlit silhouettes against a bright window, and warm table-lamp illumination with the specified pose). This instance therefore tests precise file localization, structured fact extraction from documents, and cross-modal matching under explicit constraints.

\paragraph{Example 2: Document--video compliance verification.}

The second example evaluates rule-based factual verification under multimodal evidence. The query asks whether the ``Saver Menu'' logo placement in a user advertisement complies with McDonald's clearspace guidelines. The ground-truth answer relies on two evidence files: the brand manual \texttt{mcdonalds\_logomanual.pdf} (which specifies the clearspace rule) and the advertisement video \texttt{2024\_Saver\_Menu\_...mp4} (from which the spatial layout is inspected). The visualization anchors the decision to (i) a normative textual constraint from the manual (``clearspace is equal to the width of one leg of the Golden Arches'') and (ii) a frame-level measurement/inspection of the logo--text separation in the video segment (00:07--00:08). This instance stresses constrained extraction of a precise rule from a document and its verification against visual evidence, emphasizing grounded correctness and low-hallucination restatement.

\subsubsection{Profiling}
\paragraph{Motivation.}
Profiling queries require user-level inference from weak, distributed evidence spread across files, modalities, and time. Unlike factual retention, which is anchored to explicit verifiable facts, profiling depends on aggregating repeated traces into coherent abstractions such as routines, preferences, scheduling policies, retrospective accounts, and workflows. The remainder of this subsection expands these profiling regimes through a subtask decomposition and representative grounded cases.

\begin{figure}[ht]
    \centering
    \includegraphics[width=0.66\textwidth]{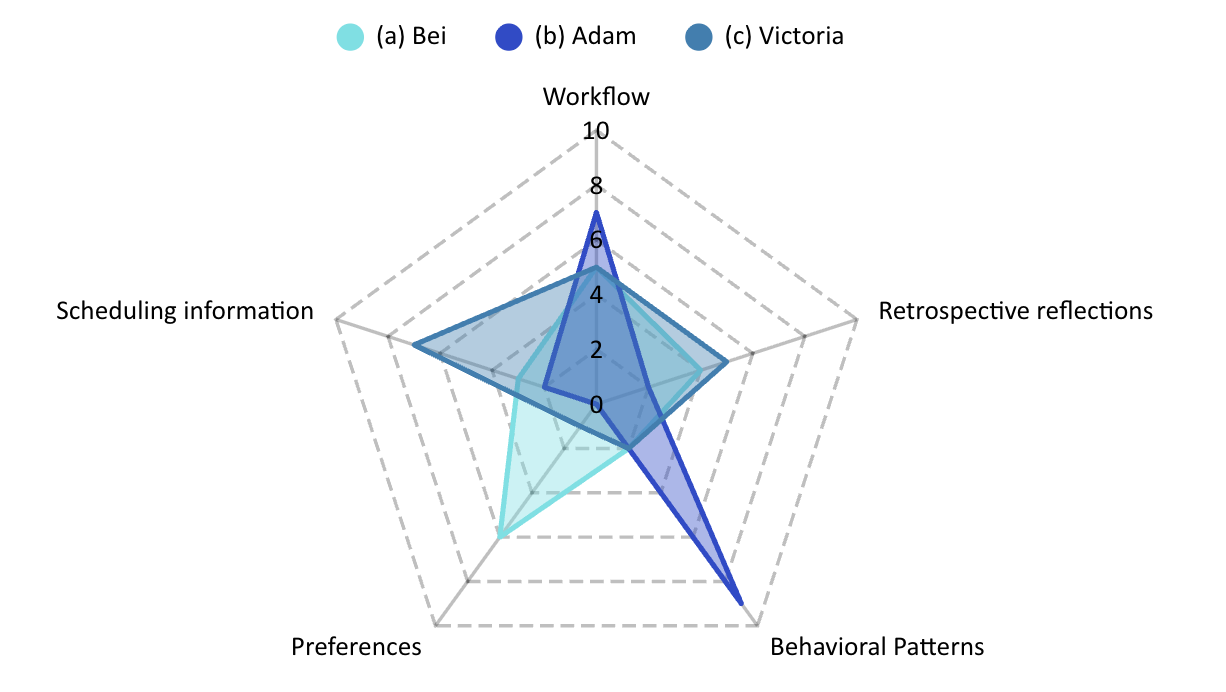}
    \caption{\textbf{Profiling subtask distribution.} Proportions of the five profiling subtasks (preferences, behavioral patterns, scheduling information, retrospective reflections, and workflows) across the three HippoCamp profiles.}
    \label{fig:subtask_dist}
\end{figure}

\paragraph{Definition.}
A profiling query requires inferring \emph{user-level attributes} from device-resident evidence distributed across files, modalities, and time. Answers are typically not grounded in a single decisive statement; instead, they rely on \emph{weak, distributed signals} (e.g., repeated scheduling choices, recurring activity traces, consistent edits or formatting habits, and multi-step workflow artifacts) that must be integrated into a globally consistent inference. Operationally, profiling demands (i) \emph{longitudinal evidence integration} across temporally separated records, (ii) \emph{event-to-trait abstraction} that generalizes from episodic observations to stable characteristics, and (iii) \emph{context-aware personalization} that produces actionable outputs coherent with the user’s constraints and history. Accordingly, evaluation emphasizes profile consistency, correct temporal anchoring, executability of suggested actions when applicable, and traceability to grounded evidence.

\paragraph{Profiling subtasks overview.}
To make profiling evaluation more interpretable, we decompose profiling queries into five recurrent subtasks: \textbf{preferences}, \textbf{behavioral patterns}, \textbf{scheduling information}, \textbf{retrospective reflections}, and \textbf{workflows}. All five require evidence grounding, but they differ in their dominant abstraction operator, ranging from trait inference and periodicity recovery to constraint-aware planning and procedure-level synthesis. \cref{fig:subtask_dist} visualizes their distribution across the three profiles.

\begin{figure}[ht]
    \centering
    \includegraphics[width=0.86\textwidth]{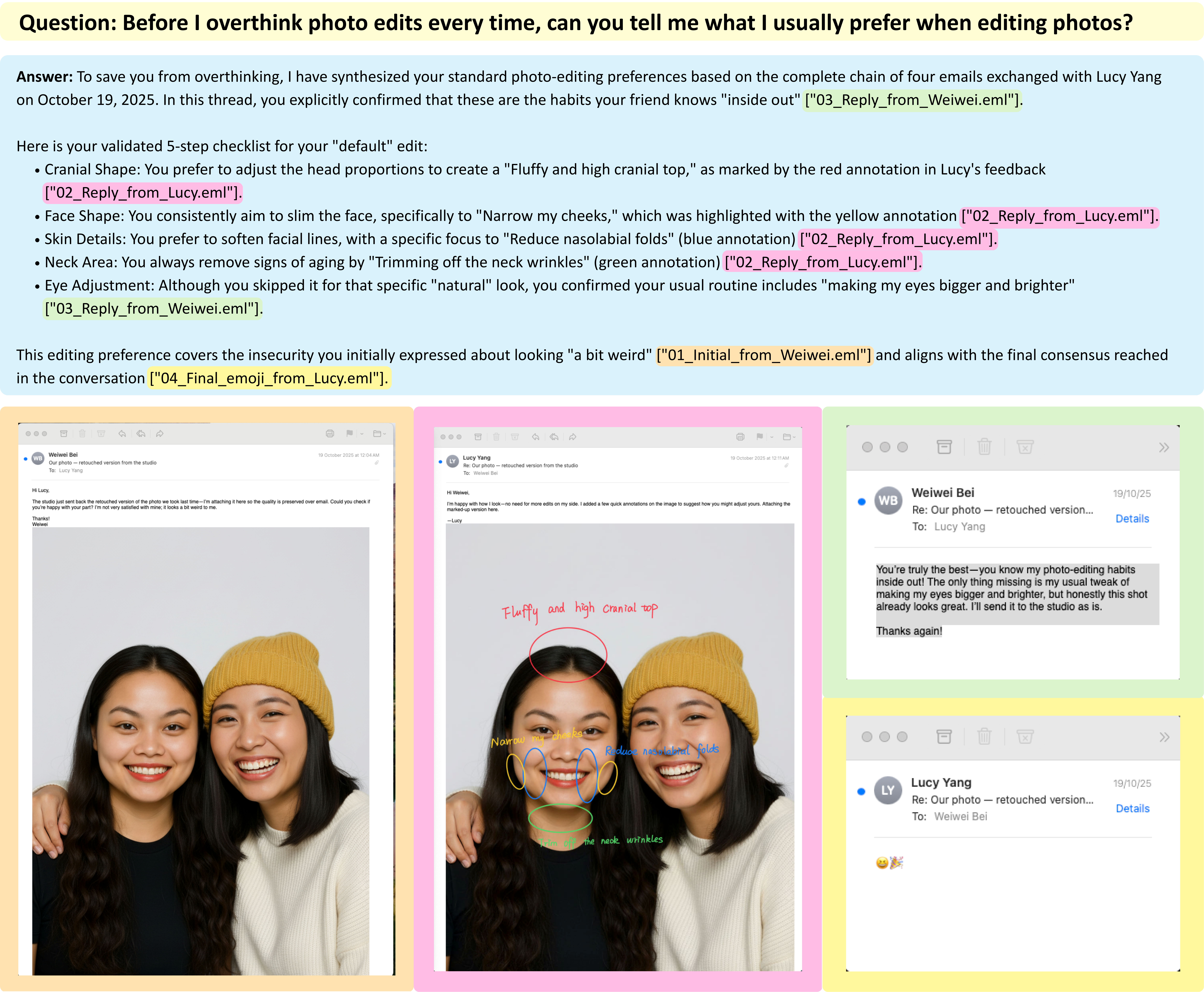}
    \caption{\textbf{Profiling subtask: Preferences.} Inferring stable photo-editing preferences from an email thread with annotated visual feedback and user confirmation.}
    \label{fig:preferences}
\end{figure}

\paragraph{i) Preference Case Study.}
Preference profiling targets stable, trait-like choices reflected across repeated decisions, annotations, and self-authored artifacts. \cref{fig:preferences} presents a representative instance grounded in a four-email thread, where annotated visual feedback and explicit user confirmations jointly specify a consistent photo-editing template. The evidence indicates recurring adjustments to face or cranial shape, skin details, and neck wrinkles, together with a routine eye adjustment explicitly described as a usual operation. This case evaluates whether an agent can aggregate multimodal conversational evidence into a stable preference model while keeping each inferred trait traceable to file-grounded cues.

\begin{figure}[ht]
    \centering
    \includegraphics[width=0.86\textwidth]{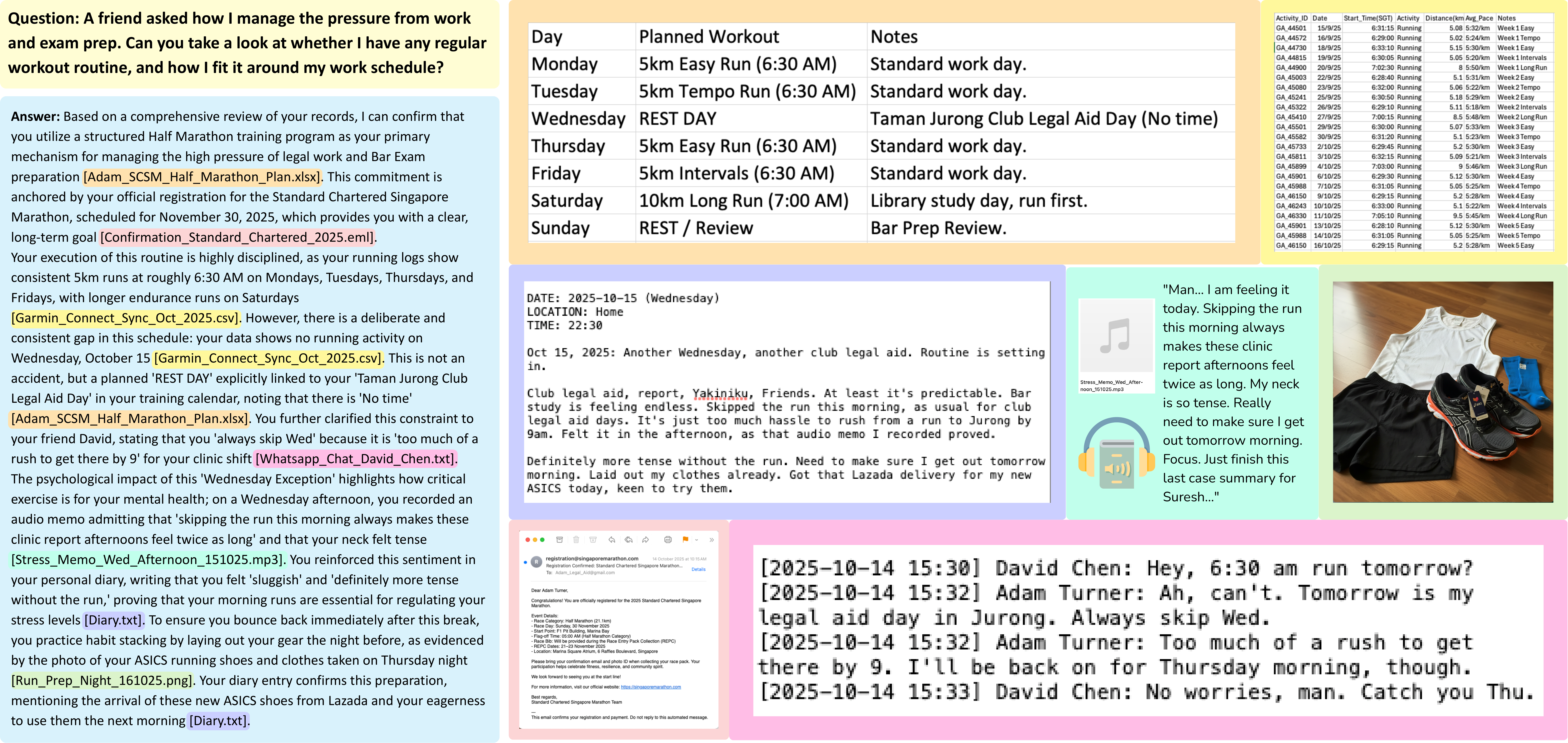}
    \caption{\textbf{Profiling subtask: Behavioral Patterns.} Inferring a stable stress-regulation routine by aligning a training plan, logs, messages, diary notes, and multimodal evidence over time.}
    \label{fig:behaviour_patterns}
\end{figure}

\paragraph{ii) Behavioral Patterns Case Study.}
Behavioral-pattern profiling targets persistent, temporally structured habits rather than isolated events. \cref{fig:behaviour_patterns} presents a representative instance on stress management under sustained exam and workload pressure, where the ground-truth pattern is a regular morning-running routine with a systematic Wednesday exception. The evidence is distributed across multiple modalities, including a half-marathon training plan and confirmation email, running logs, chat records explaining the weekly exception due to legal-aid commitments, diary entries reflecting increased tension when the run is skipped, a voice memo linking missed exercise to stress, and a preparation photo showing the recurring execution mechanism of laying out running gear the night before. This case evaluates whether an agent can integrate cross-file temporal cues, identify periodicity together with structured exceptions, and derive a stable behavioral pattern that remains traceable to verifiable evidence.

\begin{figure}[ht]
    \centering
    \includegraphics[width=0.86\textwidth]{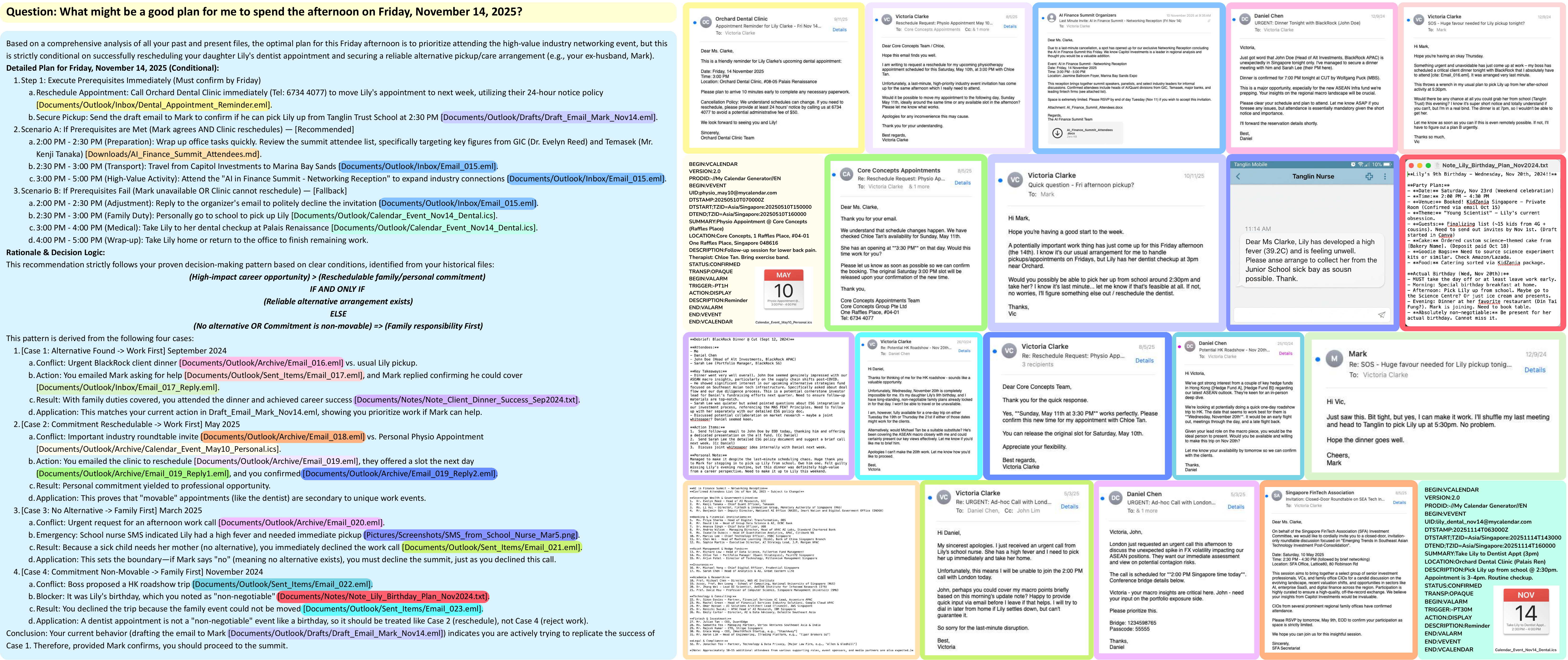}
    \caption{\textbf{Profiling subtask: Scheduling Information.} Conditional planning under conflicting commitments, grounded in calendars/emails and historical conflict-resolution evidence.}
    \label{fig:Scheduling}
\end{figure}

\paragraph{iii) Scheduling Information Case Study.}
Scheduling-information profiling targets forward-looking, constraint-aware planning under a user's established decision policy. \cref{fig:Scheduling} presents a representative instance that asks for a plan from 2:00--5:00 PM on Nov 14, 2025, under competing family and professional commitments. The ground-truth solution is explicitly conditional: it prioritizes a high-value networking event only if the user can (i) reschedule a child's dental appointment and (ii) secure a reliable alternative pickup or care arrangement; otherwise the plan switches to a family-first schedule. The evidence combines current constraints from calendar entries and reminder emails with historical conflict cases that reveal a consistent resolution policy: work first when commitments are movable and alternatives exist, but family first when alternatives fail or the obligation is non-negotiable. This case evaluates whether an agent can recover a user-specific decision model from longitudinal evidence and generate an executable plan that is both constraint-consistent and evidence-traceable.

\begin{figure}[ht]
  \centering
  \includegraphics[width=0.86\linewidth]{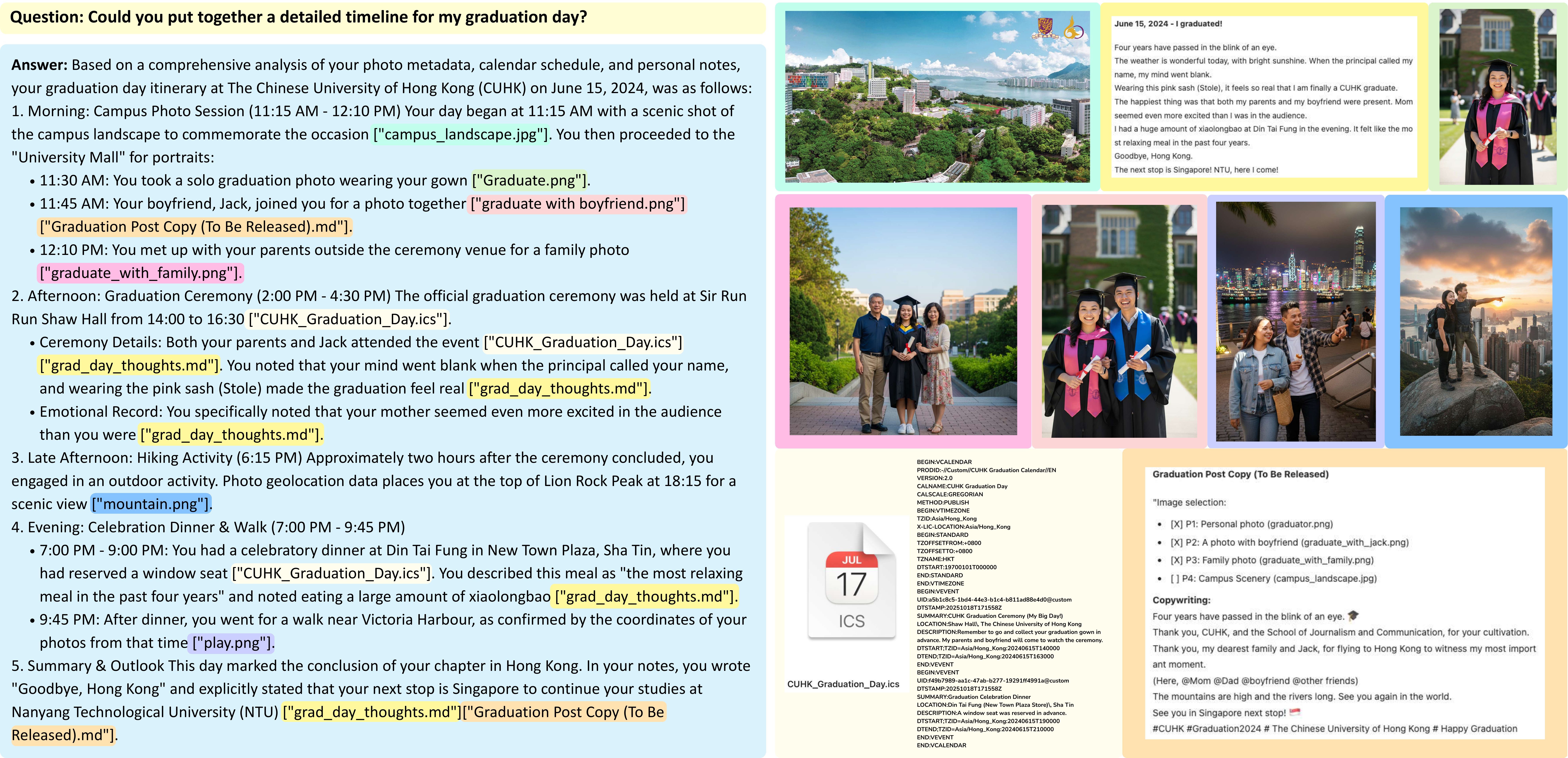}
  \caption{\textbf{Profiling subtask: Retrospective Reflections.} The agent reconstructs a graduation-day itinerary by aligning calendar events, photo metadata, and personal notes into a coherent timeline.}
  \label{fig:Retrospections}
\end{figure}

\paragraph{iv) Retrospective Reflections Case Study.}
Retrospective-reflection profiling targets event-bounded user history and requires reconstructing what happened during a specific episode from incomplete, heterogeneous traces. \cref{fig:Retrospections} presents a representative graduation-day reconstruction task in which the agent must recover a coherent itinerary by aligning calendar events, photo metadata, geolocation cues, and personal notes. The grounded schedule is supported jointly by fixed commitments and locations from \texttt{CUHK\_Graduation\_Day.ics}, timestamped campus and outdoor photos, and notes or draft post content that provide contextual details and participants. This case evaluates whether an agent can align timestamps, locations, and participants across modalities to produce a logically consistent retrospective account while keeping each inferred step traceable to concrete file-grounded evidence.

\begin{figure}[ht]
    \centering
    \includegraphics[width=0.86\textwidth]{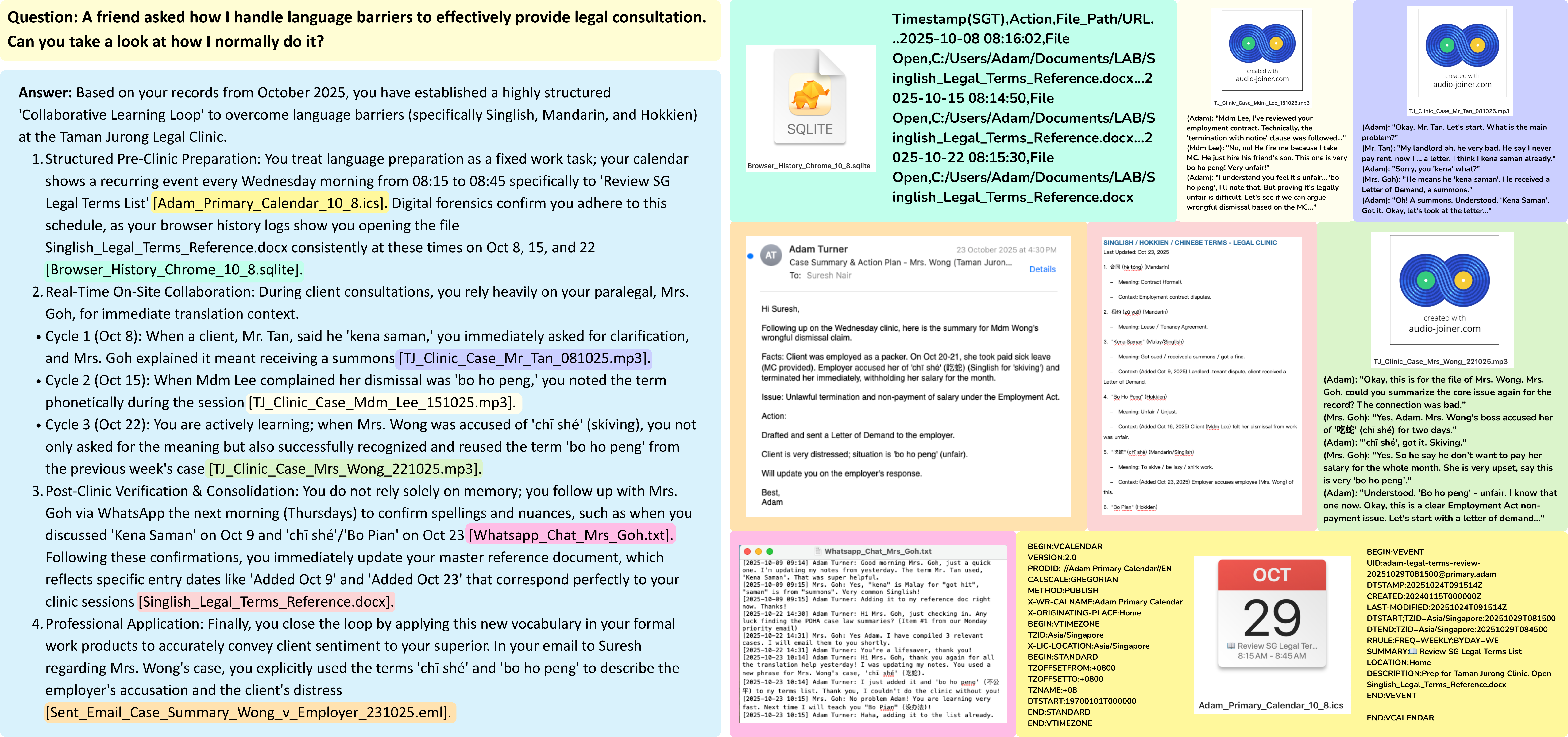}
    \caption{\textbf{Profiling subtask: Workflows.} Reconstructing a recurring ``collaborative learning loop'' by aligning calendar events, logs, consultation audio, messages, and document updates.}
    \label{fig:Workflows}
\end{figure}

\paragraph{v) Workflows Case Study.}
Workflow profiling targets procedure-level user modeling from repeated task executions. \cref{fig:Workflows} presents a representative language-barrier workflow in Adam's legal-aid practice, where the agent must abstract a recurring ``collaborative learning loop'' from distributed evidence over multiple occurrences. The grounded evidence supports three recurrent phases: (i) \emph{pre-session preparation}, anchored by a weekly calendar event to review a legal-terms reference and corroborated by repeated file-open actions in browser or history logs; (ii) \emph{in-session collaboration}, where consultation audio recordings show real-time clarification and translation with a paralegal across cases; and (iii) \emph{post-session consolidation}, where follow-up messages verify terms and the master reference document is updated, culminating in professional application in case-summary emails. This case evaluates whether an agent can integrate multimodal procedural traces into a coherent, reusable workflow representation while keeping each step traceable to concrete file-grounded evidence.

\subsection{Complexity Axes and Marginal Distributions}

\begin{figure}[ht]
    \centering
    \includegraphics[width=\linewidth]{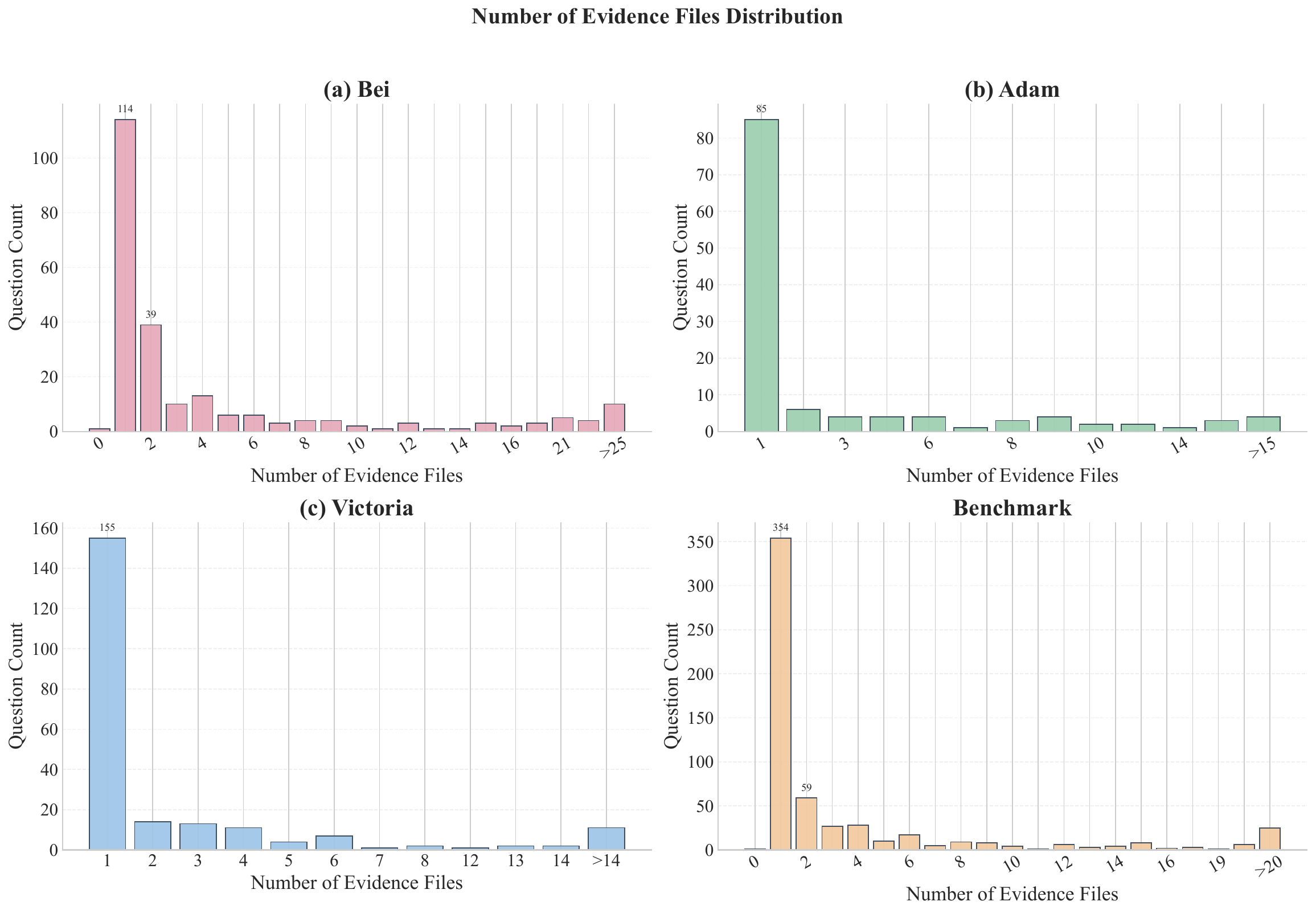}
    \caption{\textbf{Evidence breadth distribution.} Number of ground-truth evidence files required per query, shown for each profile and overall. The distribution is heavy-tailed, with many multi-file queries beyond 20 evidence files.}
    \label{fig:evidence_overall}
\end{figure}

Before introducing a scalar difficulty score, we first characterize HippoCamp along three complementary \emph{complexity axes}: \textbf{evidence breadth} (the number of distinct evidence files required), \textbf{modality breadth} (the number of modalities involved), and \textbf{reasoning depth} (the number of annotated rationale steps). These marginal indicators correspond to increasingly demanding requirements on retrieval, multimodal perception, and multi-step integration. They provide an interpretable first-order view of benchmark complexity, ranging from single-source lookups to long-horizon queries that require aggregating many files, aligning evidence across modalities, and executing extended reasoning chains.

\subsection{Evidence Breadth}
\cref{fig:evidence_overall} reports the number of ground-truth evidence files required per query, which quantifies \emph{retrieval breadth} in a realistic file-system ``haystack''. The distribution is sharply peaked at a single evidence file yet exhibits a pronounced heavy tail. In particular, one-file questions account for 114 instances in Bei, 85 in Adam, and 155 in Victoria, and 354 overall; however, a substantial fraction requires aggregating multiple files (e.g., 59 two-file questions overall), and each profile contains non-trivial mass beyond 20 evidence files (e.g., 19 instances for Bei). This pattern is ecologically plausible for personal devices: many queries originate from a single artifact, but the relevant context is frequently fragmented across attachments, versions, and cross-referenced records, inducing multi-file reasoning. Crucially, even ``unimodal'' or apparently simple questions can be multi-file, requiring agents to navigate directory structure, deduplicate near-duplicates, reconcile conflicting timestamps, and verify conclusions across corroborating sources. The long tail of high-evidence queries therefore provides hard stress tests for robust retrieval and evidence management, and it exposes failure modes in tool use, grounding, and cross-file synthesis that are invisible in small-corpus or single-document benchmarks.

\begin{figure}[ht]
    \centering
    \includegraphics[width=\linewidth]{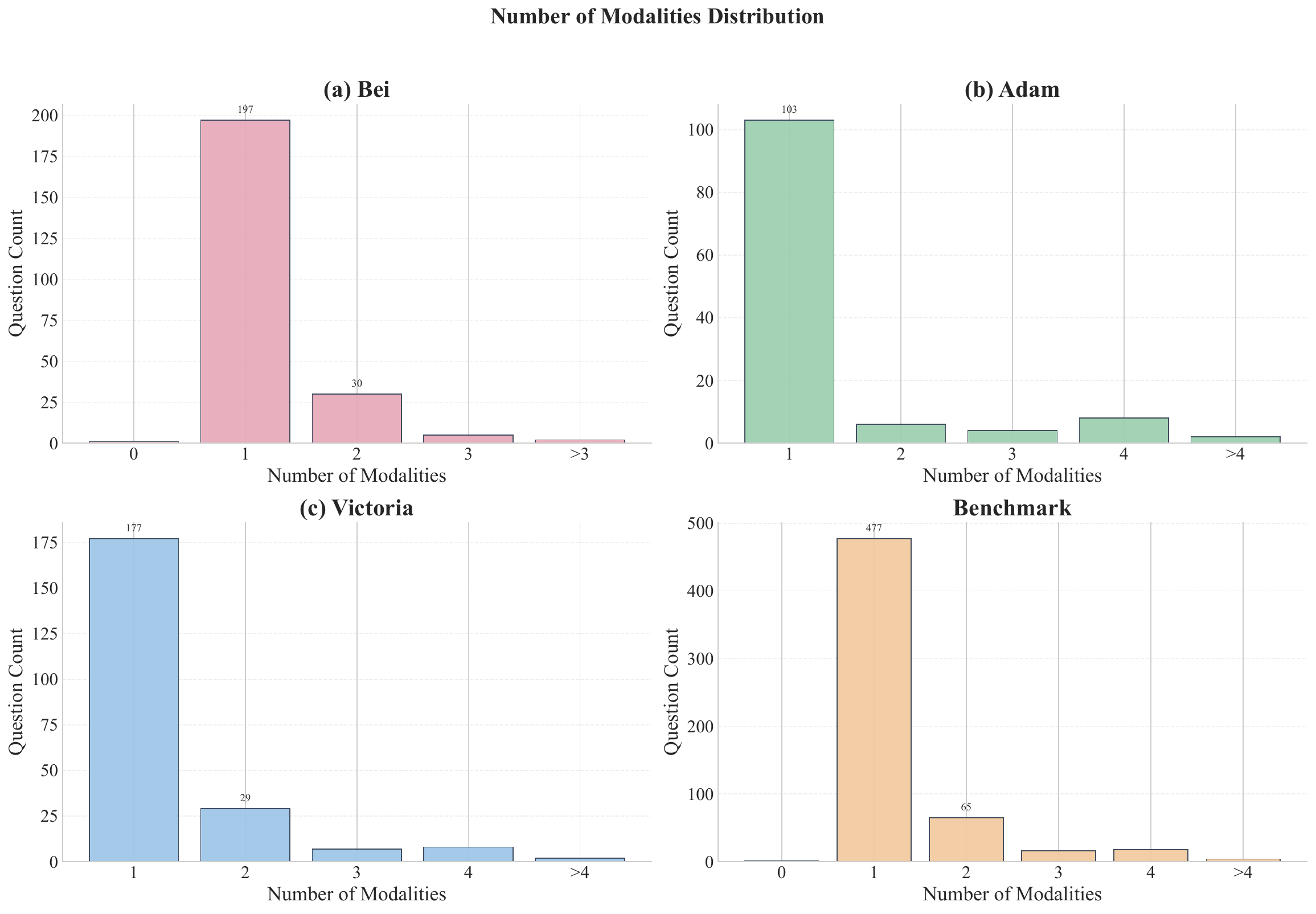}
    \caption{\textbf{Modality breadth distribution.} Number of distinct modalities required per query, shown for each profile and overall.}
    \label{fig:modality_overall}
\end{figure}

\subsection{Modality Breadth}
\cref{fig:modality_overall} measures \emph{modality breadth}, i.e., the number of distinct modalities required per query, which isolates the need for cross-modal perception and grounding. While unimodal queries dominate, HippoCamp contains a meaningful multimodal tail, including 65 two-modality queries overall and additional instances requiring three or more modalities. This distribution mirrors real file-system ecology: users often seek answers within one modality, yet high-value assistance frequently hinges on aligning evidence across modalities (e.g., linking a textual rule to visual media, or validating an event via timestamps across emails, calendars, and photos). Importantly, multimodality in HippoCamp is not superficial; it is coupled with evidence grounding and cross-file dependency, requiring agents to interpret heterogeneous formats, localize evidence spans, and integrate signals under temporal constraints. Together with evidence breadth, modality breadth expands the benchmark’s coverage from single-file perception to multi-file, multi-modal reasoning, enabling fine-grained diagnosis of agent capabilities in search, perception, and reasoning under realistic device-scale conditions.

\subsection{Reasoning Depth}
\begin{figure}[ht]
    \centering
    \includegraphics[width=0.8\linewidth]{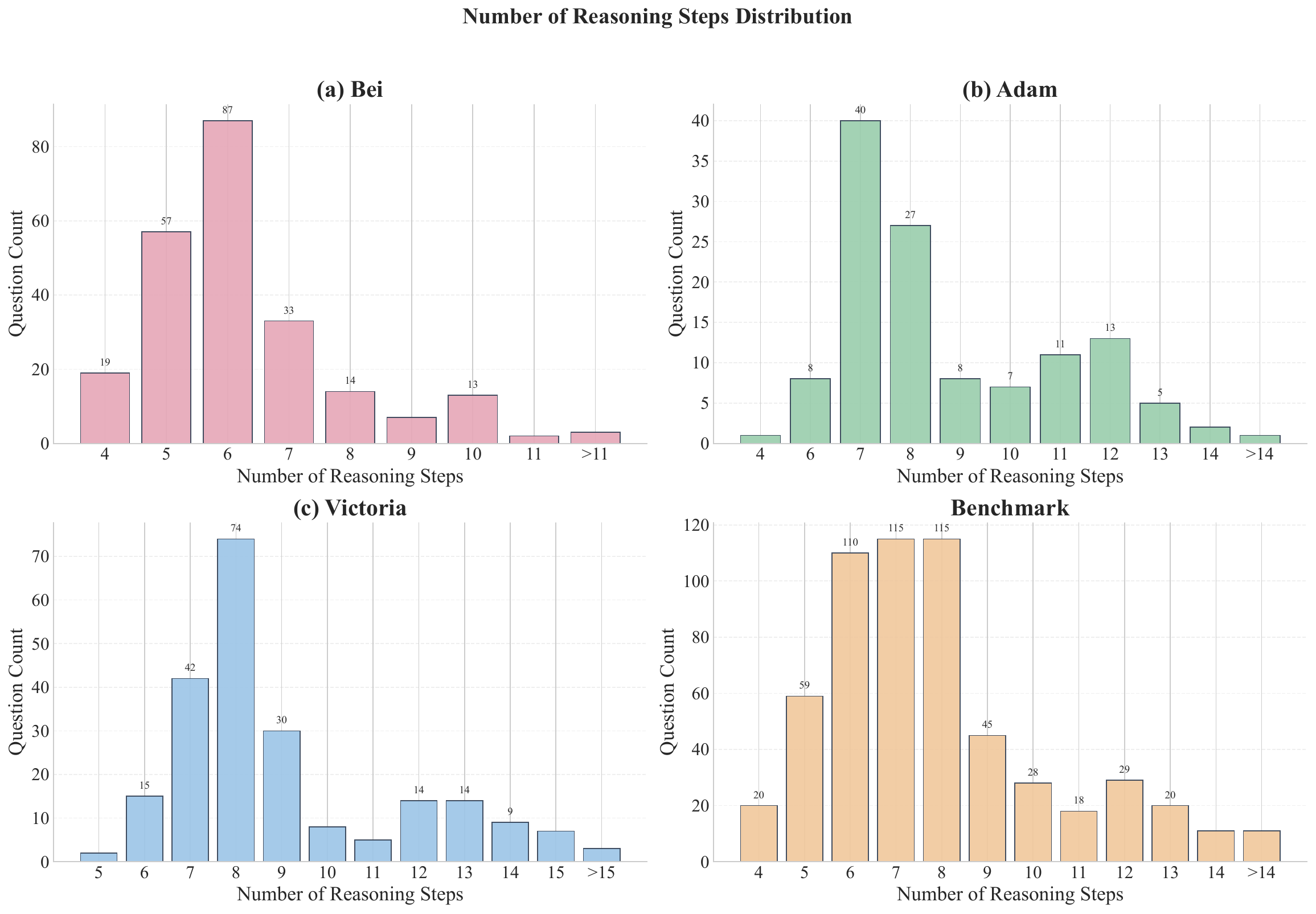}
    \caption{\textbf{Reasoning depth distribution.} Number of annotated reasoning steps required per query for each profile and overall. The distribution peaks at medium depth (6-8 steps) with a long tail of deep multi-step queries.}
    \label{fig:steps_overall}
\end{figure}

\cref{fig:steps_overall} reports the distribution of annotated reasoning-step counts per query, which serves as an explicit proxy for \emph{reasoning depth} under our structured trajectories. Across profiles and in aggregate, HippoCamp concentrates around medium-depth problems while maintaining a non-trivial long tail of deep multi-step queries. In Bei, most questions fall within 5-8 steps (57 at 5 steps and 87 at 6 steps), with additional mass extending to 10+ steps; Adam exhibits a broader spread with peaks at 7-8 steps (40 and 27, respectively) and a pronounced tail through 11-13 steps; Victoria similarly concentrates at 7-8 steps (42 at 7 steps and 74 at 8 steps) while retaining substantial mass at 12--14+ steps. Aggregated over the full benchmark, the distribution peaks at 5-8 steps (59 at 5, 110 at 6, and 115 at both 7 and 8) and exhibits a heavy tail beyond 10 steps (e.g., 28 at 10, 18 at 11, 29 at 12, 20 at 13), including a non-trivial $>14$ bin. This structure is both ecologically plausible and diagnostically useful. Real personal-computing queries often require multiple rounds of retrieval, evidence inspection, cross-file reconciliation, and verification before an answer is justified, especially for profiling and cross-modal tasks. The medium-depth mass ensures broad coverage of everyday multi-step reasoning, while the long tail provides hard stress tests that expose compounding failures in planning, grounding, and verification. 

The preceding analysis reports three marginal indicators of complexity: evidence breadth, modality breadth, and reasoning depth. Although these axes provide an interpretable first-order characterization of the benchmark, they do not fully capture \emph{effective} question hardness in realistic file systems. In particular, queries that appear simple along one axis may still be difficult because of coupled constraints, such as tight cross-file dependencies within a single modality, cross-modal grounding within a single evidence file, or high perception burden from structurally complex documents (e.g., tables, figures, scanned pages, and embedded visuals in PDFs). We therefore introduce a separate scalar \emph{question difficulty} score in Appendix~\ref{sec:question_difficulty} to approximate benchmark hardness beyond these marginal statistics.

\subsection{Question Difficulty}
\label{sec:question_difficulty}

\subsubsection{Definition}
\label{sec:question_difficulty_def}

We define a scalar \emph{question difficulty} score as a heuristic diagnostic measure of effective hardness in device-resident, multimodal file reasoning, where difficulty often arises from \emph{coupled} retrieval, perception, and multi-step integration rather than from any single marginal statistic. For each query $q$, we extract eight interpretable factors that approximate distinct sources of cognitive and computational load:
(i) the number of ground-truth evidence files $n_f$,
(ii) the number of modalities involved $n_m$,
(iii) the number of distinct file types/extensions $n_t$,
(iv) the number of localized evidence items $n_e$,
(v) the number of annotated reasoning steps $n_r$,
(vi) the question length in tokens $n_q$,
(vii) the answer length in tokens $n_a$,
and (viii) the temporal span in days covered by the evidence $n_\Delta$.

\paragraph{Robust normalization.}
Each factor $x$ is mapped to a bounded score $s(x)$ using a log--quantile transform to handle heavy-tailed distributions while preventing outliers from dominating:
\begin{equation}
s(x)=\mathrm{clip}\!\left(\frac{\log(1+x)}{\log(1+P_{90}(x))},\,0,\,c\right),
\end{equation}
where $P_{90}(x)$ is the benchmark-wide 90th percentile of factor $x$, $\mathrm{clip}(\cdot)$ truncates to $[0,c]$, and $c$ is a fixed cap. This normalization preserves ordering for typical cases, rewards long-tail difficulty, and yields comparable scales across heterogeneous factors.

\paragraph{Core difficulty and interaction coupling.}
We compute a weighted base score
\begin{equation}
\mathrm{Base}(q)=\sum_i w_i\, s_i(q),
\end{equation}
where $s_i(q)$ denotes the normalized score for factor $i$ and weights $w_i$ emphasize axes most indicative of hard personal-file QA (notably evidence breadth, evidence localization, and reasoning depth). To capture the empirical observation that difficult questions are often difficult because multiple constraints co-occur, we add interaction terms:
\begin{equation}
\mathrm{Inter}(q)=\alpha_1\sqrt{n_f n_r}+\alpha_2\sqrt{n_m n_t}+\alpha_3\sqrt{n_e n_\Delta}.
\end{equation}
These terms model coupled challenges such as multi-file reasoning, multimodal long-tail formats, and long-horizon evidence alignment.

\paragraph{Hard-case bonus and final mapping.}
We further apply a conservative bonus when key axes (evidence files, evidence items, reasoning steps) are simultaneously high, reflecting compounded difficulty not captured by linear aggregation alone. The raw score is then mapped to a $[0,100]$ scale via a sigmoid transformation
\begin{equation}
\mathrm{Diff}(q)=\frac{100}{1+\exp(-\gamma(\mathrm{Raw}(q)-\tau))},
\end{equation}
which improves separability between medium and genuinely hard tails while avoiding over-inflation. We report benchmark-wide summary statistics (mean/median and high-difficulty ratios) and use this score to characterize difficulty distributions across profiles and task types. 

\subsubsection{Distribution}
\label{sec:difficulty_dist}

\begin{figure}[ht]
    \centering
    \includegraphics[width=\linewidth]{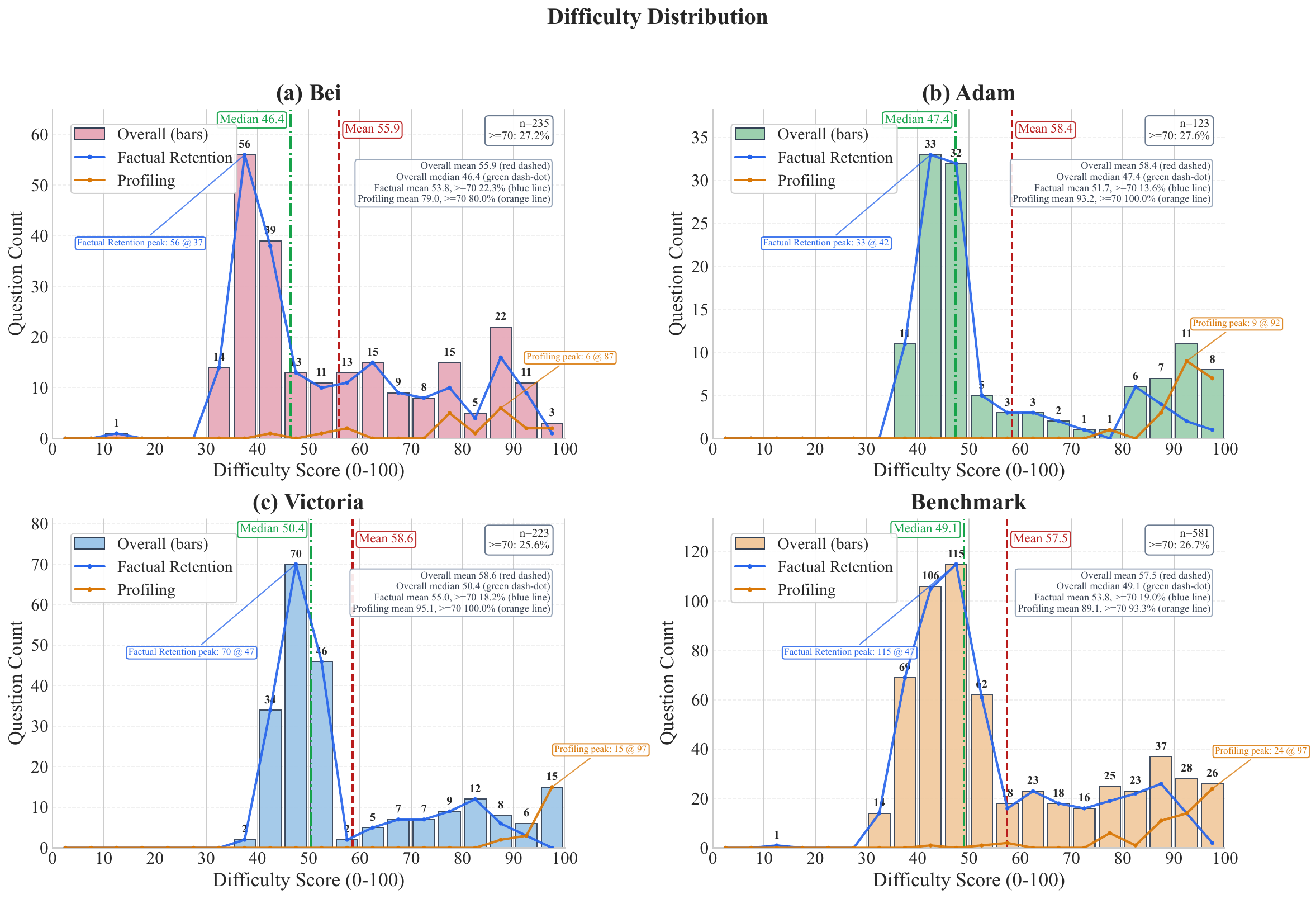}
    \caption{\textbf{Difficulty distributions on HippoCamp.} Histograms of the difficulty score for each profile and overall, with mean/median markers and task-type overlays (factual retention vs.\ profiling).}
    \label{fig:difficulty_dist}
\end{figure}

\cref{fig:difficulty_dist} reports the resulting difficulty distributions per profile and overall, together with mean/median markers and task-type overlays. Across the full benchmark, difficulty is centered at a moderate-to-high level (mean $57.5$, median $49.1$, $n{=}581$) with a substantial hard tail: $26.7\%$ of questions score $\geq 70$. Importantly, the apparent ``moderate'' center does not imply easiness; it reflects that many queries are unimodal or involve a small evidence set, yet still require non-trivial perception and verification under realistic file-system conditions. The overall histogram peaks around mid-range difficulty (factual-retention peak: $115$ questions at score $\approx 47$), while profiling concentrates near the extreme tail (profiling peak: $26$ questions at score $\approx 97$), revealing a clear separation between fact-level retrieval/verification and user-level synthesis.

Decomposing by task type further highlights HippoCamp’s diagnostic value. Factual retention occupies the mid-range with moderate variance (overall mean $53.8$; $\geq 70$ ratio $19.0\%$), whereas profiling is systematically harder (mean $89.1$; $\geq 70$ ratio $93.3\%$), consistent with the need to aggregate weak signals across time and files into coherent user-level inferences. This separation is stable across profiles: for Bei, profiling mean $79.0$ with $80.0\%$ $\geq 70$; for Adam, profiling mean $93.2$ with $100.0\%$ $\geq 70$; for Victoria, profiling mean $95.1$ with $100.0\%$ $\geq 70$. By contrast, factual retention remains challenging but less extreme, with profile-specific means in the low-to-mid $50$s (Bei $53.8$, Adam $51.7$, Victoria $55.0$) and non-trivial hard tails (e.g., Bei $\geq 70$ at $22.3\%$).

Finally, the profile-level distributions corroborate that HippoCamp measures hardness under diverse personal ecosystems rather than a single regime: Bei (mean $55.9$, median $46.4$, $\geq 70$ $27.2\%$), Adam (mean $58.4$, median $47.4$, $\geq 70$ $27.6\%$), and Victoria (mean $58.6$, median $50.4$, $\geq 70$ $25.6\%$) show comparable overall difficulty while differing in where the mass concentrates and how the profiling tail manifests. Together, these distributions demonstrate that HippoCamp is both \emph{broad} (covering common medium-difficulty personal queries) and \emph{deep} (containing a sizable fraction of high-difficulty instances that require coupled retrieval, multimodal grounding, and extended reasoning), thereby providing a rigorous and differentiating testbed for next-generation file-system agents.

\subsubsection{Correlation between Difficulty and Performance}
\label{sec:difficulty_perf_corr}

\begin{figure}[ht]
    \centering
    \includegraphics[width=\linewidth]{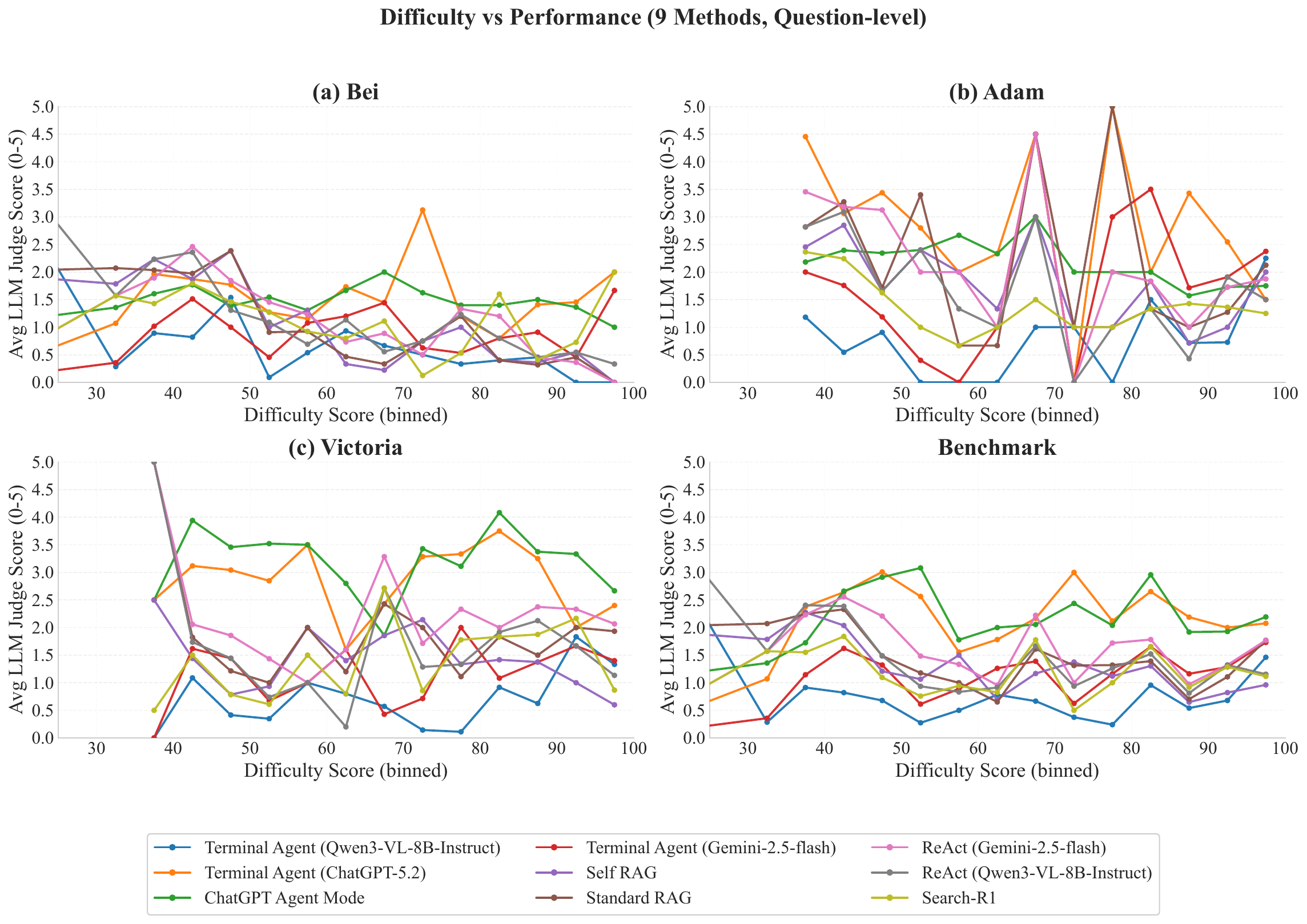}
    \caption{\textbf{Difficulty vs.\ performance (question-level).} For each profile and for the full benchmark, mean LLM-judge score (0--5) for nine methods as a function of binned difficulty (5-point bins). Scores are generally low-to-moderate and decline with increasing difficulty, with pronounced degradation in the hard tail.}
    \label{fig:difficulty_vs_performance}
\end{figure}

\paragraph{Protocol.}
To assess whether the proposed difficulty score reflects \emph{effective} benchmark hardness, we analyze its correlation with model performance at the \emph{question level}. For each query, we compute the difficulty score using the definition in \cref{sec:question_difficulty_def} and align it with the corresponding per-query LLM-judge score (\texttt{LLM\_as\_a\_judge\_score}, range 0--5) produced by each evaluated method. We then bin queries by difficulty (5-point bins) and report, for each method and each bin, the mean judge score. This yields a comparable difficulty-performance profile across profiles and for the merged benchmark.

\paragraph{Results.}
\cref{fig:difficulty_vs_performance} shows average judge score as a function of binned difficulty for each profile and overall. Two observations are consistent across panels. First, scores generally \emph{decrease} as difficulty increases, supporting that the difficulty definition captures non-trivial sources of hardness beyond marginal statistics. Second, even in low-to-mid difficulty bins, absolute scores remain modest for most methods and concentrate in the lower-to-middle portion of the 0-5 range, indicating that HippoCamp is challenging throughout rather than only in the extreme tail. In the high-difficulty regime, performance drops further and several method families approach near-floor behavior, reflecting failures to sustain grounded retrieval, multimodal perception, and multi-step verification when constraints co-occur. Together, these trends corroborate HippoCamp’s diagnostic value: it induces a broad hardness spectrum while revealing clear capability gaps in current systems as difficulty increases (see also \cref{tab:main_exp,tab:analysis_exp}).

\subsection{Profile Example Set}
\label{sec:profile_examples}

\begin{figure}[ht]
    \centering
    \includegraphics[width=\textwidth]{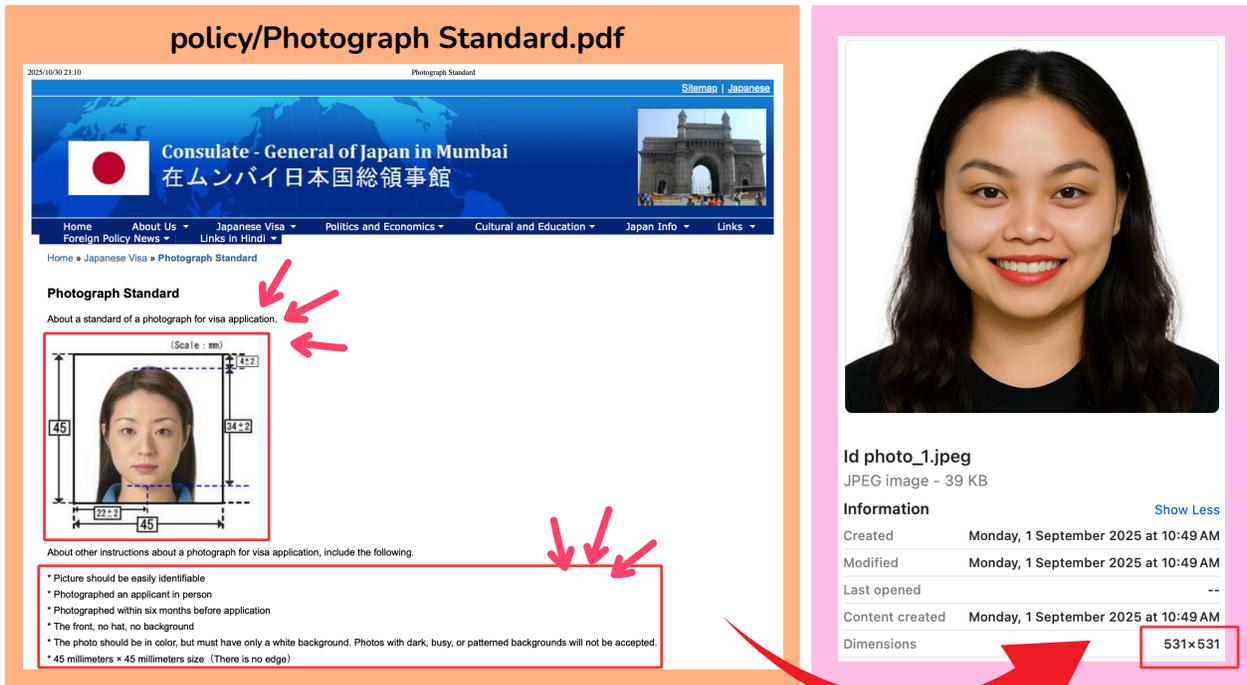}
    \caption{\textbf{Profile (a) Bei Weiwei example.} A factual-retention query requiring extraction of visa photo constraints from an official document and verification against personal images, with grounded evidence.}
    \label{fig:Bei}
\end{figure}

\subsection{Representative Example from Profile (a) Bei Weiwei}

\cref{fig:Bei} presents a representative factual-retention instance that stress-tests evidence-grounded verification under normative constraints. The query asks for a photo that satisfies official Japanese visa requirements. An agent must (i) locate and parse the governing specification in \texttt{policy/Photograph Standard.pdf} to extract actionable constraints (e.g., 45\,mm$\times$45\,mm, front-facing, no headwear, white background), (ii) retrieve and shortlist candidate images from the file system, and (iii) verify compliance by jointly inspecting visual attributes (pose, background, occlusion) and checking metadata where relevant (e.g., dimensions). The ground-truth answer selects \texttt{Identity/Id\_photo\_1.jpeg} as the only compliant file, illustrating the coupled demands of search, multimodal perception, and rule-based verification with traceable evidence.

\subsection{Representative Example from Profile (b) Adam Turner}

\begin{figure}[ht]
    \centering
    \includegraphics[width=\textwidth]{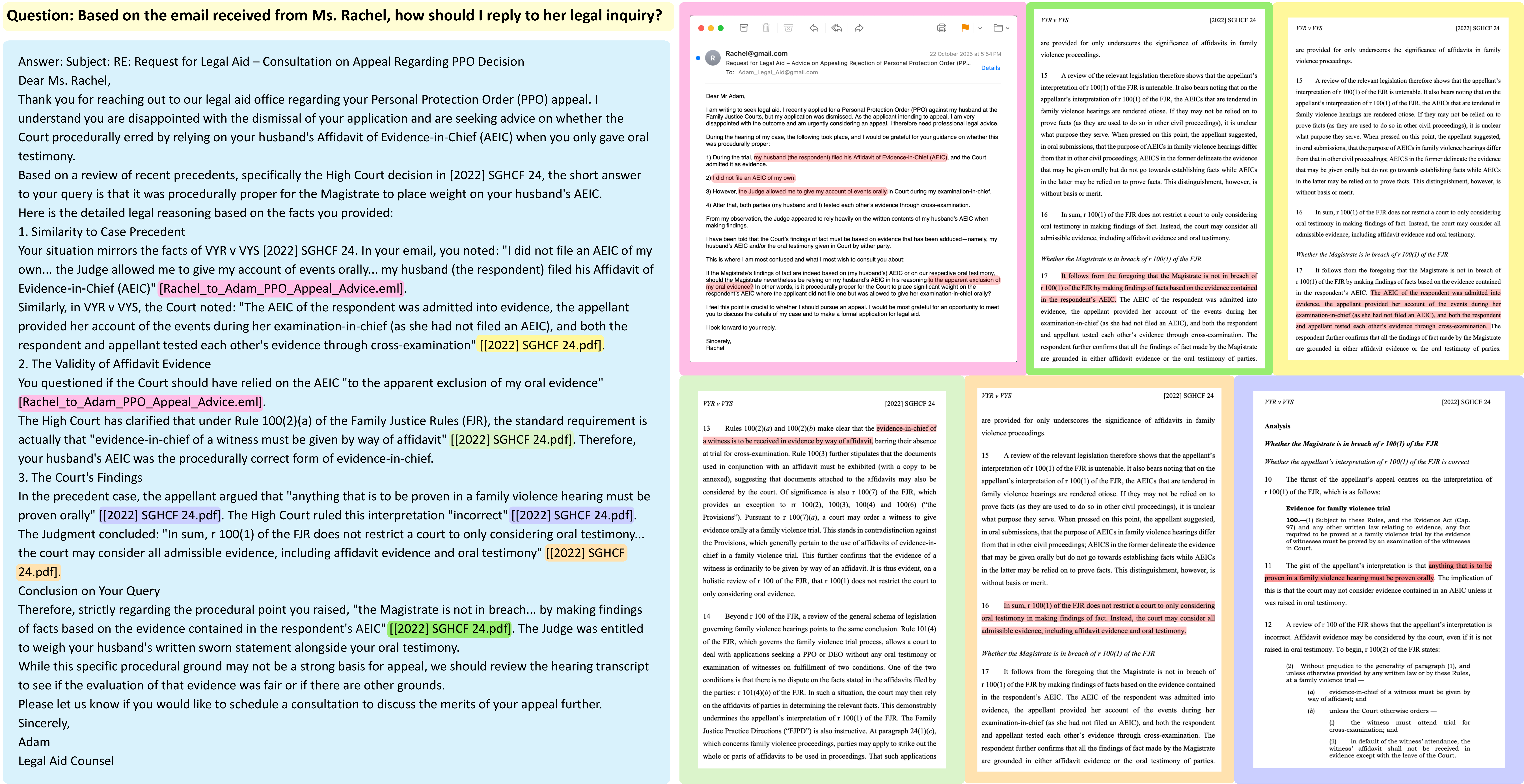}
    \caption{\textbf{Profile (b) Adam Turner example.} An evidence-grounded reply drafting task that integrates an email inquiry with supporting precedent passages to produce a procedurally accurate response.}
    \label{fig:Adam}
\end{figure}

\cref{fig:Adam} presents a representative factual-retention instance from Profile~(b) Adam Turner that requires evidence-grounded professional correspondence drafting. Given Ms.~Rachel’s email about a PPO appeal, the agent must produce an accurate reply by (i) extracting the relevant factual constraints from the email thread, (ii) retrieving the supporting precedent \textit{VYR v VYS} ([2022] SGHCF 24), and (iii) grounding the response in determinative passages, including the clarification that Rule~100(2)(a) of the Family Justice Rules requires evidence-in-chief to be given by affidavit and that the court may consider admissible affidavit and oral testimony together. The ground-truth response exemplifies citation-backed legal reasoning anchored to the evidence files, and this case further shows that HippoCamp evaluates grounded professional assistance by testing whether agents can maintain precise cross-document retrieval and evidence-linked justification in specialized domains.

\subsection{Representative Example from Profile (c) Victoria Anne Clarke}

\begin{figure}[ht]
    \centering
    \includegraphics[width=\textwidth]{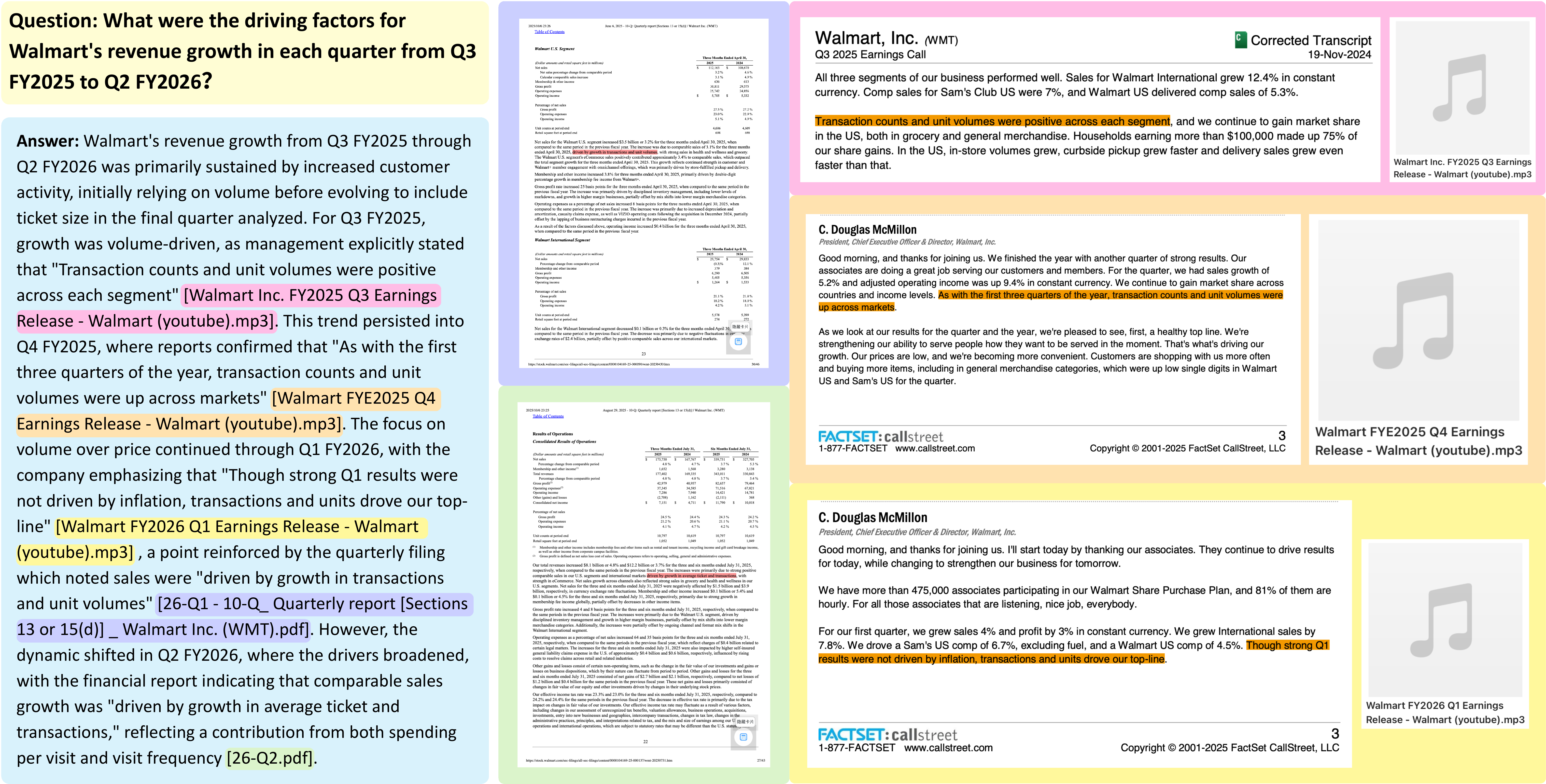}
    \caption{\textbf{Profile (c) Victoria Anne Clarke example.} A factual-retention query requiring cross-quarter attribution by aligning earnings-call transcripts and filings to identify changing revenue-growth drivers.}
    \label{fig:Victoria}
\end{figure}

\cref{fig:Victoria} presents a representative factual-retention instance from Profile~(c) Victoria Anne Clarke that requires cross-quarter attribution grounded in multimodal corporate disclosures. The query asks for the drivers of Walmart’s revenue growth from Q3 FY2025 to Q2 FY2026. To answer, an agent must retrieve and align evidence across multiple sources, including earnings-call audio transcripts and quarterly filings, then extract and reconcile the stated growth drivers over time. The grounded answer shows a consistent volume-led narrative through Q3--Q1 (transaction counts and unit volumes repeatedly cited as the primary drivers, including explicit statements that results were not inflation-driven) and a shift in Q2 FY2026 toward joint contributions from \emph{average ticket} and \emph{transactions}. This example therefore stress-tests temporal alignment across quarters, cross-document corroboration, and evidence-linked summarization of changing causal attributions under realistic, document-heavy professional workloads.

\subsection{Edge Cases and Challenging Examples}
\label{sec:edge_cases}

\begin{table}[ht]
\centering
\caption{\textbf{Diagnostic edge cases in HippoCamp.} A compact set of tail instances that instantiate extreme \emph{evidence breadth}, \emph{modality breadth}, and \emph{reasoning depth}.}
\label{tab:edge_cases_compact}
\scriptsize
\setlength{\tabcolsep}{3pt}
\renewcommand{\arraystretch}{1.15}
\begin{tabularx}{\linewidth}{@{}l|>{\raggedright\arraybackslash}X@{}}
\toprule
\textbf{Axis} & \textbf{Selected tail instances (Profile; ID; complexity; diagnostic requirement)} \\
\midrule
\textbf{Evidence breadth} &
(a) Bei; ID~178; \textbf{111 files}: \textit{How long was my last stay in Japan?} (broad retrieval, pruning, cross-file corroboration). \newline
(b) Adam; ID~93; \textbf{56 files}: \textit{Filter transferred companies with Date Incorporation after 2015.} (large evidence-set management and structured verification). \\
\textbf{Modality breadth} &
(a) Bei; ID~11; \textbf{6 modalities}: \textit{How should I structure my folders for organizing reinforcement learning study materials?} (cross-modal grounding across heterogeneous artifacts). \newline
(c) Victoria; ID~5; \textbf{5 modalities}: \textit{I'm thinking about picking up the guitar. How might I get started?} (multi-modal evidence alignment with user-specific context). \\
\textbf{Reasoning depth} &
(c) Victoria; ID~34; \textbf{17 steps}: \textit{How might Apple's buyback cadence evolve if cash balances trend lower or rates shift?} (long-horizon, verification-centric synthesis). \newline
(b) Adam; ID~95; \textbf{15 steps}: \textit{Check if there are missing/incorrect parts in my Manufacturing company info.} (iterative checking, error finding, and evidence-linked correction). \\
\bottomrule
\end{tabularx}
\end{table}

These edge cases are rare but consequential: they occupy the extreme tails of the benchmark’s complexity axes and function as \emph{diagnostic stress tests} that separate superficial retrieval from robust, evidence-grounded agent behavior. Evidence-breadth tails require scalable search over fragmented personal histories, including candidate pruning, deduplication, and cross-file corroboration; modality-breadth tails require genuine cross-modal grounding, where decisive signals are distributed across heterogeneous media rather than reducible to a single text document; and reasoning-depth tails require extended, verification-centric multi-step synthesis in which intermediate conclusions must remain traceable to evidence to avoid cascading errors. Importantly, these tails arise from realistic personal workflows rather than synthetic construction, and they complement the benchmark’s large mid-range mass by exposing capability limits that are typically invisible in single-document or tool-centric benchmarks. Together, they demonstrate that HippoCamp is simultaneously \emph{broad} (covering everyday personal queries) and \emph{deep} (containing hard, compositional instances that probe retrieval, grounding, and verification under device-scale conditions). Such compositional tails are particularly scarce in existing benchmarks, yet they are precisely where state-of-the-art agents fail in our experiments.

\FloatBarrier

\section{Evaluation Protocol and Robustness}
\label{app:evaluation_robustness}
\label{sec:eval_robust}

We evaluate HippoCamp under a controlled, profile-isolated protocol designed to balance fairness, reproducibility, and fidelity to each method’s native interaction paradigm. Rather than forcing all methods into a single artificial budget, we evaluate each system under a pre-specified, method-appropriate budget within its native implementation or serving environment. This appendix is organized around four components: shared evaluation constraints and framework design; method execution regimes and tooling; budgets, retries, and randomness control; and the metrics and robustness procedures used for answer-level and evidence-level evaluation. We conclude with an extended metric summary that complements, rather than repeats, the main-text results.

\subsection{Shared Evaluation Constraints and Framework}
\label{sec:eval_shared_framework}

\paragraph{Unified evaluation framework.}
All evaluated methods are executed through a shared evaluation harness that standardizes dataset loading, method invocation, result recording, and metric computation. The purpose of this design is not to erase method heterogeneity, but to ensure that all systems are assessed under the same benchmark records, profile-local information boundaries, and output normalization rules. As illustrated in \cref{fig:evaluation_pipeline} (reproduced in the main text), the evaluator loads benchmark JSON records, dispatches them to a registered method backend through a common interface, collects answers and optional traces, and computes standardized metrics for aggregation and analysis.

\subsubsection{Pipeline Overview}
The evaluation pipeline proceeds in four stages. First, the evaluator loads each dataset record from JSON, extracting the query text, gold answer, and any optional supervision such as file-level evidence annotations or capability labels. Second, it constructs a unified request object and dispatches it through the shared evaluator interface to the selected method. This evaluator-side contract is common across retrieval-native systems, terminal-style agents, and hosted agent modes, even though their underlying tool surfaces and orchestration policies differ. Third, the evaluator records the method outputs, including the final answer and, when available, retrieved evidence, search traces, tool calls, and runtime statistics such as latency, step counts, and retries. These outputs are normalized into a fixed result schema so that downstream scoring remains method-agnostic. Finally, the evaluator computes the enabled metrics and writes two levels of output: a per-query evaluation record containing all query-level statistics, and an aggregated summary file containing dataset-level results. This design ensures consistency across all evaluated methods while supporting efficient large-scale re-evaluation under different metric configurations.

\subsubsection{Profile-Isolated Evaluation}
All evaluations are conducted in a strictly \emph{profile-isolated} setting. For each query, a method is granted access only to the simulated file system of the corresponding HippoCamp profile; no information from the other two profiles is exposed at inference time. This isolation applies uniformly to raw files, accessible metadata, and any benchmark-provided interfaces. Methods receive as input only the natural-language query, and must solve it using information that would plausibly be available within a single user’s device environment. This design prevents cross-profile leakage and ensures that performance reflects profile-specific search, perception, and reasoning rather than benchmark-wide memorization or unintended transfer across users.

\subsubsection{Execution Interfaces}
The evaluation harness supports two broad classes of systems. The first consists of \emph{retrieval-native} methods, including Standard RAG and Self-RAG, which operate over the profile-local corpus through a retrieval interface and return evidence-conditioned answers. The second consists of \emph{interactive agent} methods, including Search-R1, the ReAct variants, the Terminal Agent variants, and ChatGPT Agent Mode, which may iteratively search, inspect files, invoke tools, and refine intermediate hypotheses before producing a final answer. Regardless of implementation style, all methods are normalized to a common evaluation interface: the primary output is the final answer, and whenever available we additionally record retrieved files or evidence lists, intermediate traces or tool calls, and runtime statistics such as latency, search iterations, and retries. This normalization enables task-level comparison under shared benchmark and profile-local access constraints across systems with substantially different architectures, while preserving their native modes of interaction.

\subsubsection{Allowed and Disallowed Channels}
The harness permits only benchmark-local interaction channels that are explicitly exposed by the evaluation environment. Allowed channels include the profile-local file system, the benchmark-provided Docker API and terminal interface for vacuum-agent settings, and the native retrieval backend for retrieval-based methods. Disallowed channels include public web search, external retrieval corpora, hidden metadata not exposed to the method, and any profile-external information sources. For interactive agents, multimodal content is likewise mediated through the provided interfaces rather than given by default.

\subsection{Method Regimes and Tooling}
\label{sec:eval_regimes_tooling}

HippoCamp evaluates methods under three execution regimes that differ in how retrieval, tool use, and multimodal perception are realized: (i) a \emph{native retrieval setting}, where methods operate directly over a benchmark-local retrieval backend; (ii) a \emph{vacuum Docker agent setting}, where terminal-style agents interact with a profile-specific container through a controlled command surface; and (iii) an \emph{official hosted agent setting}, used for commercial agent products whose internal orchestration cannot be faithfully reproduced locally. Across all three regimes, accessible information remains profile-local. Comparisons across regimes are therefore intended to assess grounded task performance under matched profile-local information access, rather than to assert identical low-level tool interfaces or perfectly matched execution affordances.

\subsubsection{Native Retrieval Setting}
\label{sec:native_retrieval_setting}

We first evaluate a set of methods in a \emph{native retrieval setting}, where the system operates directly over the benchmark-local corpus rather than through the vacuum Docker terminal environment. This group includes two retrieval-native baselines (Standard RAG, Self-RAG) and three search-enabled generators (Search-R1 and two ReAct variants). Although these methods differ substantially in interaction style, they share the same profile-local retrieval backend and do not access external web resources. The goal of this setting is to compare methods under their native retrieval or search loop while preserving a common corpus and search scope.

\paragraph{Standard RAG~\cite{Lewis2020RAG}.}
Standard RAG is implemented as a classical retrieve--rerank--return pipeline. Given a query, the system first performs vector similarity retrieval over the benchmark-local vector store, optionally reranks the retrieved chunks, and returns the top-$k$ results to the generator. In our implementation, retrieval is parameterized by the number of initially retrieved documents (\texttt{top\_k}), whether reranking is enabled, and the final reranked cutoff (\texttt{rerank\_top\_k}). The method uses an embedding model, vector store, and optional reranker, but does not perform iterative query refinement, tool use, or environment interaction. It therefore serves as the simplest retrieval-native baseline in our comparison.

\paragraph{Self-RAG~\cite{Asai2024SelfRAG}.}
Self-RAG extends the standard retrieval setup with an internal reflection stage. The system first retrieves candidate chunks from the same benchmark-local vector store, then uses a generator to grade the relevance of each retrieved item, filters the set to keep only items above a relevance threshold, and optionally rewrites the query and repeats retrieval if no sufficiently relevant evidence is found. Key control parameters include the retrieval depth (\texttt{top\_k}), the relevance threshold, and the maximum number of refinement iterations. Despite this self-reflective loop, Self-RAG remains a retrieval-native method rather than a full interactive agent, because it does not operate through the external environment interface or perform explicit file-level tool use.

\paragraph{Search-R1~\cite{Jin2025SearchR1}.}
Search-R1 is evaluated as an end-to-end generator with \emph{internalized search}. Unlike RAG pipelines, it does not rely on a separate retrieval provider exposed to the evaluation harness. Instead, the model interleaves internal reasoning and retrieval through a structured loop with special tags such as \texttt{<think>}, \texttt{<search>}, and \texttt{<answer>}. Search queries are issued dynamically during generation, and the returned results are incorporated back into the model context for subsequent reasoning. In our setup, these searches are executed against the same benchmark-local retrieval server as other native methods, ensuring that search scope remains profile-local and comparable.

\paragraph{ReAct~\cite{Yao2023ReAct} variants.}
The ReAct methods implement the classic Thought--Action--Observation loop, in which the model explicitly reasons about what information is needed, issues a search action, receives observations from the local retrieval server, and iterates until it produces a final answer. Both ReAct variants share the same ReAct interaction skeleton and the same benchmark-local search backend; they differ only in the underlying generation model, namely Gemini-2.5-flash~\cite{Comanici2025Gemini25} versus Qwen3-30B-A3B~\cite{Qwen3VL2025}. In contrast to Standard RAG and Self-RAG, the ReAct systems are search-interactive rather than purely retrieval-native, but they are still evaluated outside the Docker terminal environment and do not rely on external web access.

\subsubsection{Vacuum Docker Agent Setting}
\label{sec:vacuum_docker_setting}

We next evaluate three terminal-style agents in a controlled \emph{vacuum Docker} environment: Terminal Agent (Qwen3-VL-8B-Instruct~\cite{Qwen3VL2025}), Terminal Agent (ChatGPT-5.2~\cite{openai2025gpt5_2_system_card}), and Terminal Agent (Gemini-2.5-flash~\cite{Comanici2025Gemini25}). Unlike the native retrieval setting, these methods interact with the benchmark through a unified terminal-style interface exposed inside a profile-specific container. This setup is designed to approximate file-system interaction under controlled conditions while preserving comparability across agent implementations. All three agents access the profile-local file system through the same container-resident command/API surface. The environment exposes five benchmark-defined primitives. \texttt{list\_files} provides directory-level discovery by listing files under the accessible data root, optionally with pattern-based filtering, and is therefore used for broad search and navigation. \texttt{return\_metadata} returns structured file attributes, including file type, modality, timestamps, and location-related fields when available, enabling metadata-aware reasoning without revealing file content. \texttt{return\_txt} returns a structured text representation of a file (e.g., extracted text segments together with basic file information), making it the primary interface for textual inspection of documents and other parseable formats. \texttt{return\_img} renders a file or a selected page into image form and returns the resulting image path together with image payloads, supporting page-level inspection of visually rich or scanned content. Finally, \texttt{return\_ori} returns the original file path and raw file bytes, enabling exact-fidelity access when a method can directly consume the source file. Together, these interfaces expose complementary levels of abstraction: discovery (\texttt{list\_files}), metadata inspection (\texttt{return\_metadata}), text-level access (\texttt{return\_txt}), rendered visual access (\texttt{return\_img}), and raw-byte access (\texttt{return\_ori}). This environment is controlled rather than fully equivalent to a real operating system. Agents do not receive unrestricted desktop interaction; instead, all file access is mediated through benchmark-provided commands and conversion utilities. The evaluation value of this setting therefore lies not in faithfully reproducing a consumer OS, but in providing a reproducible, profile-isolated, and tool-consistent substrate for terminal-based agent interaction. Because all terminal agents are restricted to the same command set, differences in performance can be attributed more directly to the model’s ability to plan, search, interpret returned content, and manage intermediate hypotheses, rather than to differences in tool availability. A further design choice is that multimodal access is \emph{terminal-result driven} rather than unconditional. In particular, image or raw-file channels become available to the agent only when it explicitly invokes \texttt{return\_img} or \texttt{return\_ori} and the command succeeds. The returned payload is then transformed into the model-specific multimodal input format used by the corresponding agent backend. Consequently, multimodal perception in this setting is not free: it must be triggered through deliberate tool use, just as file discovery and metadata inspection must be triggered through explicit commands.

\subsubsection{Docker Tooling and Multimodal Return Path}
\label{sec:docker_mm_path}

The vacuum Docker environment is built on an Ubuntu-based image and packages all profile-local resources required for controlled agent interaction, including benchmark data, metadata, gold text representations, auxiliary tools, and the lightweight WebUI/API layer. In addition to standard file serving, the container includes a small set of conversion utilities that make heterogeneous personal files accessible through a unified interface, including LibreOffice for Office documents, Poppler/\texttt{pdf2image} for PDF rendering, SQLite support for structured database files, and dedicated metadata/image serving endpoints. The goal is not to emulate a full desktop operating system, but to expose a reproducible and tool-consistent substrate for terminal-based agents. Within this environment, all terminal agents interact through the same command surface: \texttt{list\_files}, \texttt{return\_txt}, \texttt{return\_img}, \texttt{return\_ori}, and \texttt{return\_metadata}. These commands define a common abstraction over the underlying file system, ranging from discovery and metadata inspection to text extraction, rendered image access, and raw-file transfer. Because the command inventory is identical for all terminal agents, tooling parity is enforced at the interface level: every method receives access to the same benchmark-local functions, and differences in performance arise from how effectively the model plans and exploits these tools rather than from tool availability itself. A critical aspect of fairness is that multimodal input is \emph{not} exposed by default. Instead, multimodal access is terminal-result driven: an agent enters the multimodal channel only after it explicitly invokes \texttt{return\_img} or \texttt{return\_ori} and the command returns \texttt{success=true}. Thus, visual or source-level content must be \emph{requested} through deliberate tool use rather than being passively injected into the model context. This makes multimodal perception part of the agent’s problem-solving burden, alongside file search and evidence localization. The returned multimodal payload is then consumed through a model-specific transport path. For ChatGPT-compatible terminal agents, rendered images are attached as image blocks or embedded as data-URL payloads. For Gemini-based terminal agents, the same outputs are converted into native multimodal parts (e.g., uploaded image/file content) before the next model call. In both cases, the benchmark-level command semantics remain identical; what differs is only the backend-specific serialization of returned artifacts into the model’s input channel. This separation is important: the environment equalizes \emph{what} information can be requested, while the model backend determines only \emph{how} that returned information is ingested.

\subsubsection{Official Hosted Agent Setting}
\label{sec:official_hosted_setting}

In addition to the native retrieval and vacuum-Docker settings, we evaluate \textbf{ChatGPT Agent Mode~\cite{openai2024gpt4technicalreport, openai2025gpt5}} in its \emph{official hosted} configuration provided by OpenAI. This setting corresponds to a commercial product deployment rather than a locally controlled research environment. Consequently, the method does not operate through our vacuum Docker interface and is not constrained to the same command/API surface as the terminal agents. We explicitly distinguish this configuration as a \emph{native hosted agent mode}. The distinction matters because its internal tooling, orchestration policy, and multimodal handling are managed by the product platform rather than by our evaluation harness, and therefore cannot be fully standardized against the Docker-based agents. Nevertheless, we include it in the comparison because it represents a strong and practically relevant reference point for real-world deployed agent systems. In other words, while it is not strictly tool-parallel to the vacuum-Docker agents, it provides an important upper-bound-style comparison for what a state-of-the-art hosted agent can achieve on HippoCamp under its official usage conditions.

\subsection{Budgets, Retries, and Randomness}
\label{sec:budgets_randomness}

Because HippoCamp contains long-horizon, multimodal tasks, evaluation outcomes are sensitive to resource allocation. We therefore make resource budgets explicit and interpret them as part of the evaluation protocol rather than as hidden implementation details. We do not enforce a single globally matched budget across all systems; instead, each method is evaluated under a pre-specified, method-appropriate budget within its native implementation or serving environment. The relevant constraints include search iterations, retrieved evidence volume, generation length, wall-clock runtime, and method-specific stopping rules.

\subsubsection{Resource Budgets}
\label{sec:resource_budgets}

We distinguish between two broad budget regimes.

\paragraph{Retrieval-native methods.}
For retrieval-native systems, including Standard RAG and Self-RAG, the dominant budget axes are retrieval depth, reranking/reflection depth, and generation length. Standard RAG uses a fixed retrieval budget with a bounded number of initially retrieved chunks, an optional reranking stage, and a final reranked cutoff; it does not perform iterative search or environment interaction, and therefore its effective budget is determined primarily by the retrieval top-$k$, rerank top-$k$, and the maximum generation length of the answer model. Self-RAG operates under the same corpus-local retrieval setting but allocates additional budget to self-reflection, including bounded relevance grading, filtering, optional query rewriting, and a capped number of refinement iterations. In both cases, early stopping is enabled whenever the retrieval or generation pipeline returns no further valid evidence, and wall-clock runtime is bounded by the completion of the underlying retrieval/generation calls rather than by an external interaction loop.

\paragraph{Interactive agents.}
For interactive methods, including Search-R1, the ReAct variants, the terminal agents, and ChatGPT Agent Mode, the relevant budget axes are more heterogeneous. Search-R1 and ReAct-based systems are bounded by the maximum number of search or reasoning turns, the number of retrieved results returned per search call, the maximum generation tokens per turn, and any built-in stopping conditions (e.g., terminating when a final answer action is produced). Terminal agents in the vacuum Docker environment are additionally constrained by terminal-interaction budget, including the number of tool-use iterations and the wall-clock time available to complete a query. ChatGPT Agent Mode is evaluated under the practical limits of the hosted product setting, which include platform-side constraints on interaction length, generation budget, and runtime. Across all interactive methods, early stopping is used whenever the method’s native control loop terminates with a final answer or when no further productive interaction can be carried out within the remaining budget.

\subsubsection{Retries and Failure Handling}
\label{sec:retry_failure}

Because several evaluated systems are interactive and tool-using, failures may arise not only from reasoning errors but also from malformed actions, incomplete outputs, or environment-side execution issues. We therefore specify retry handling explicitly. In general, we do \emph{not} retry successful but incorrect answers: reruns are reserved for cases in which a run fails to produce a valid evaluable output. Concretely, retry triggers include (i) malformed outputs that cannot be parsed into a final answer, (ii) empty answers, (iii) tool failures that prevent access to the requested benchmark-local resource, and (iv) hard timeouts. When such a failure occurs, the system is re-executed from the same query under the same environment and budget constraints; successful but incorrect answers are retained as failures rather than rerun. Reported results are then produced according to the pre-specified execution policy of each evaluated method under this framework. This distinction is particularly important for unstable interactive agents, for which execution failures can materially affect the final measured outcome.

\subsubsection{Randomness and Determinism}
\label{sec:randomness_determinism}

We control randomness whenever the underlying method permits it, but exact determinism is not uniformly attainable across all evaluated systems. For retrieval-native methods, randomness is limited and can largely be controlled through fixed seeds and deterministic retrieval settings, with variability arising mainly from stochastic generation components when enabled. For interactive agent methods, additional non-determinism enters through search ordering, multi-turn decoding, tool-use branching, and backend-specific sampling behavior. Accordingly, we report the sampling configuration used by each method, including generation temperature where applicable. Hosted commercial APIs introduce a further source of variability because their internal serving stack and decoding behavior are not fully exposed to the user. In particular, hosted agent modes may remain non-deterministic even when prompts and visible settings are held fixed. As a result, runtime variance should be expected, especially for long-horizon interactive methods. When repeated runs are not feasible for all methods due to cost or platform constraints, we state this explicitly and interpret results as point estimates under the corresponding execution regime rather than as fully stabilized averages. This treatment is intended to make residual variance visible rather than implicitly hiding it behind incomplete claims of determinism.

\subsection{Metrics and Judge Robustness}
\label{sec:metrics_and_judge}

HippoCamp is designed to evaluate not only whether a method reaches the correct final answer, but also whether it retrieves the necessary evidence and exercises the appropriate capabilities under realistic personal-file conditions. Accordingly, we report metrics at three complementary levels: \textbf{answer quality}, \textbf{evidence retrieval quality}, and \textbf{capability-wise performance}. This decomposition aligns with the benchmark’s core objective of diagnosing failures in search, perception, and reasoning rather than collapsing all behavior into a single scalar score.

\subsubsection{Answer Quality}
\label{sec:answer_quality_metrics}

We evaluate answer quality using an LLM-as-a-judge protocol that returns both a binary correctness decision and a graded semantic score. The judge provides the initial assessment for all examples, while stratified manual audit is applied to sampled cases to verify judgment quality. For each query, the judge receives the question, the model prediction, and the ground-truth answer, and outputs:
(i) a binary label \texttt{pred} $\in \{\texttt{yes}, \texttt{no}\}$ indicating whether the prediction is semantically acceptable, and
(ii) an integer score $s_i \in [0,5]$ measuring answer quality on a coarse semantic scale.

\paragraph{Accuracy (Acc).}
We report \emph{accuracy} (equivalently, pass rate) as the fraction of judged responses marked \texttt{yes}:
\[
\mathrm{Acc}=\frac{\#(\mathrm{pred}=\texttt{yes})}{N},
\]
where $N$ is the number of judged instances. In the aggregated result files, this quantity is also reported as \emph{Pass Rate}. It captures whether a method produces a semantically correct answer under the judge’s acceptance criterion.

\paragraph{Average judge score.}
For a set of $N$ instances, we compute the average judge score as
\[
\bar{s}=\frac{1}{N}\sum_{i=1}^{N} s_i,
\qquad s_i \in [0,5].
\]
When exported in the per-domain summary CSV, we additionally report a rescaled \emph{Avg Score (/10)}:
\[
\mathrm{AvgScore}_{/10}=2 \cdot \bar{s},
\]
so that the reported value lies on a 0--10 scale. This graded score is complementary to binary accuracy: two methods may achieve similar pass rates while differing substantially in answer completeness, precision, or degree of support.

\paragraph{Benchmark-level aggregation.}
For method $m$, the benchmark-wide average across profiles is computed as the sample-weighted mean over all judged instances:
\[
\mathrm{BenchmarkAvg}_m
=
\frac{\sum_d \sum_{i=1}^{N_{m,d}} s_{m,d,i}}
{\sum_d N_{m,d}},
\]
where $d$ indexes profiles and $N_{m,d}$ is the number of judged instances for method $m$ on profile $d$. We also report the corresponding population standard deviation
\[
\mathrm{BenchmarkStd}_m
=
\sqrt{
\frac{1}{N_m}
\sum_{j=1}^{N_m}
\left(s_{m,j}-\mu_m\right)^2
},
\]
where $\mu_m=\mathrm{BenchmarkAvg}_m$ and $N_m=\sum_d N_{m,d}$.

These answer-level metrics directly reflect the benchmark’s primary end objective: whether a system can produce a correct and useful response grounded in the user-local file system.

\subsubsection{Evidence Retrieval Metrics}
\label{sec:evidence_retrieval_metrics}

Answer correctness alone is insufficient for HippoCamp, since a method may arrive at a plausible answer while failing to retrieve or ground the necessary evidence. We therefore evaluate retrieval quality against the annotated \emph{minimal supporting file set}. Let $G_i$ denote the ground-truth file set for instance $i$ and $\hat{R}_i$ the file set retrieved or referenced by the method. We compute file-level precision, recall, and F1 as
\[
P_i=\frac{|G_i \cap \hat{R}_i|}{|\hat{R}_i|}, \qquad
\mathrm{Rec}_i=\frac{|G_i \cap \hat{R}_i|}{|G_i|}, \qquad
F1_i=\frac{2P_i\,\mathrm{Rec}_i}{P_i+\mathrm{Rec}_i}.
\]
Reported \emph{File Precision}, \emph{File Recall}, and \emph{File F1} are obtained by averaging these instance-level values over the evaluation split.

\paragraph{File hit rate.}
We additionally report \emph{File Hit Rate}, which in our implementation corresponds to mean file-level recall:
\[
\mathrm{HitRate}=\frac{1}{N}\sum_{i=1}^{N} \mathrm{Rec}_i.
\]
This quantity captures the extent to which the method covers the required supporting evidence, even if it also retrieves spurious files.

\paragraph{Interpretation for profiling.}
For profiling tasks, the supporting evidence often consists of weak signals distributed across multiple files and time points. Accordingly, file-level F1 is always computed against the \emph{annotated minimal supporting file set}, not against the full space of potentially relevant files. This distinction is important: the retrieval metrics are intended to measure \emph{coverage of required evidence}, rather than whether the method reproduced the exact human reasoning path or retrieved every file that could plausibly contribute to the same inference.

These retrieval metrics are central to HippoCamp’s diagnostic role: they separate failures of search and evidence coverage from failures of reasoning over already-retrieved content.

\subsubsection{Capability-wise Metrics}
\label{sec:capability_metrics}

To further localize failure modes, we report capability-wise performance using the benchmark’s human-annotated \texttt{agent\_cap} labels. Each instance may contribute to one or more capability bins corresponding to \textit{search}, \textit{evidence perception}, and \textit{reasoning}. For each capability family, we first compute performance within its constituent subcategories and then aggregate those subcategory statistics.

\paragraph{Capability-wise accuracy.}
For a given subcategory $c$, let $\mathrm{yes}_c$ denote the number of judged correct responses and $\mathrm{count}_c$ the number of instances labeled with that subcategory. The subcategory accuracy is
\[
\mathrm{acc}_c = \frac{\mathrm{yes}_c}{\mathrm{count}_c}.
\]
The reported capability-family accuracy (e.g., \texttt{search/evidence/reasoning\\accuracy}) is the unweighted arithmetic mean over the corresponding subcategories:
\[
\mathrm{Acc}_{\mathrm{family}}
=
\frac{1}{|C|}
\sum_{c \in C}
\mathrm{acc}_c,
\]
where $C$ is the set of subcategories in that family.

\paragraph{Capability-wise F1.}
Similarly, if $\mathrm{f1}_c$ denotes the average file-level F1 or evidence-level F1 associated with subcategory $c$, the reported family-level F1 is
\[
\mathrm{F1}_{\mathrm{family}}
=
\frac{1}{|C|}
\sum_{c \in C}
\mathrm{f1}_c.
\]
We use unweighted rather than sample-weighted averaging so that broad capability families are not dominated by their most frequent subcategories. This makes the resulting metrics better suited for diagnostic comparison across heterogeneous reasoning skills.

\paragraph{Latency.}
Finally, for completeness, we report \emph{Avg Latency} as the mean over valid per-instance runtimes:
\[
\mathrm{AvgLatency}
=
\frac{1}{k}\sum_{i=1}^{k} t_i,
\qquad t_i > 0,
\]
where only instances with positive recorded runtime are included. Latency is not treated as a primary quality metric, but it is useful for understanding the practical trade-off between capability and efficiency across method families.

Overall, this metric suite reflects HippoCamp’s benchmark philosophy: a strong method should not only produce correct answers, but should do so by retrieving the right evidence and exercising the appropriate search, perception, and reasoning capabilities in a grounded and auditable manner.

\subsubsection{LLM-as-Judge Robustness}
\label{sec:judge_robustness}
Because a substantial portion of HippoCamp requires open-ended, evidence-grounded answers rather than exact-string matches, we adopt an LLM-as-a-judge protocol for semantic evaluation. We treat judge outputs as a controlled semantic signal rather than as a standalone oracle, and therefore make the judging setup, prompt constraints, and audit procedure explicit.

\paragraph{Judge setup.}
\label{sec:judge_setup}

For each evaluated instance, the judge receives three inputs: the \emph{question}, the \emph{ground-truth answer}, and the \emph{model prediction}. It is instructed to assess semantic match rather than lexical overlap, allowing paraphrases and non-conflicting elaborations while penalizing omission of key information. The output consists of a binary decision $\texttt{pred}\in\{\texttt{yes},\texttt{no}\}$ together with an integer score $s\in[0,5]$. This yields both a hard correctness signal and a graded quality signal, which are used throughout our answer-level evaluation.

To reduce irrelevant variation, all inputs are normalized into a fixed prompt format before judging. The judge compares only the question, the gold answer, and the candidate answer; it is not given access to hidden benchmark annotations such as capability labels, rationale traces, or gold evidence sets. In addition, the candidate response is presented without explicit system identity, so that the prompt does not directly reveal which model produced the answer. The instruction is deliberately constrained: it prioritizes semantic equivalence, explicitly tolerates paraphrases and synonyms, and requires a minimal JSON output containing only \texttt{pred} and \texttt{score}, which makes the judgment procedure easy to parse, audit, and reproduce. A representative judging instruction is shown below.

\begin{center}
\fbox{%
\begin{minipage}{0.965\textwidth}
\vspace{0.25em}
\centering {\textbf{Representative LLM-Judge Instruction}}\par
\vspace{0.35em}
\hrule
\vspace{0.55em}

\hspace{0.8em}
\begin{minipage}[t]{0.94\textwidth}
\tiny
\ttfamily
\raggedright
Compare the model output with the ground-truth answer and judge whether they match meaningfully.\\[0.35em]

\textbf{Rules}\\
- Focus on semantic equivalence; allow paraphrases and synonyms.\\
- Extra but non-conflicting details are acceptable.\\
- Missing key information should be penalized.\\
- Return an INTEGER score from 0 to 5.\\
- Output a JSON object with keys \texttt{"pred"} (\texttt{"yes"} or \texttt{"no"}) and \texttt{"score"}.\\[0.45em]

\textbf{Example outputs}\\
\{"pred":"yes","score":4\}\\
\{"pred":"no","score":1\}
\end{minipage}

\vspace{0.45em}
\hrule
\vspace{0.2em}
\end{minipage}%
}
{\captionsetup{hypcap=false}\captionof{figure}{\textbf{Representative LLM-as-a-judge instruction.} A constrained judging prompt that asks for semantic equivalence assessment and returns a binary decision together with an integer quality score.}}
\label{fig:judge_prompt}
\end{center}

\paragraph{Prompt sensitivity and human audit.}
\label{sec:judge_audit}

We do not interpret LLM judgment in isolation. Instead, it is used as a scalable semantic evaluator whose outputs are considered together with orthogonal benchmark signals, including file-level retrieval metrics and capability-wise analyses. To further control judge-side variance, we conduct a stratified human audit on sampled instances spanning all three profiles, both task families, multiple modality configurations, and multiple difficulty bands. This is important because a model may produce a plausible answer without retrieving the required evidence, or may retrieve the correct evidence but restate it incompletely. Under our evaluation design, such cases can be separated rather than collapsed into a single undifferentiated outcome. To further control judge-side variance, we conduct a stratified human audit on sampled instances spanning profiles, task families, modality configurations, and difficulty bands. The audit is intentionally concentrated on judgment-sensitive categories, including long legal or financial answers, partially correct multi-part responses, and concise but evidence-grounded outputs whose surface form may differ substantially from the reference answer. For each audited instance, reviewers inspect the question, the ground-truth answer, the model prediction, and the judge output, and assess whether the judge's binary decision and 0--5 score are semantically justified. When discrepancies are identified, they are recorded by category (e.g., omission of a decisive detail, over-acceptance of unsupported elaboration, or under-acceptance of concise but valid restatement) and used to refine our interpretation of judge-based results. The role of this audit is not to replace large-scale automatic judging, but to check whether benchmark-level conclusions are sensitive to recurrent categories of judge error. This audit protocol serves two purposes: it verifies that the constrained judging prompt behaves consistently on ambiguity-prone cases, and it provides an explicit check that benchmark conclusions are not artifacts of a small number of systematic judge errors. In this design, the binary label (\texttt{pred}) captures task-level success, while the graded score ($0$--$5$) provides a softer estimate of completeness and semantic quality. More importantly, the human audit anchors these signals to manual inspection on the cases most likely to challenge automatic judgment, making the LLM judge an audited semantic evaluator whose outputs are checked against manual inspection on ambiguity-prone cases.

\subsection{Extended Metric Summary}
\label{sec:overall_metric_summary}

\begin{table*}[ht]
\caption{\textbf{Overall metric summary on \ourmethod{}.}
For the methods in~\cref{tab:main_exp}, we report answer quality and evidence retrieval quality for \textit{profiling}, \textit{factual retention}, and the \textit{overall} benchmark average.
\textbf{Acc} denotes pass rate under the LLM-judge protocol; \textbf{Avg Score} is reported on the 0--10 scale; \textbf{Avg Latency} is the mean recorded runtime.
All rate-based metrics are shown as percentages with one decimal place (\% omitted). Best is highlighted; second-best is underlined.}
\label{tab:metric_summary}
\centering
\scriptsize
\setlength{\tabcolsep}{3.6pt}
\renewcommand{\arraystretch}{1.10}
\begin{adjustbox}{max width=\textwidth}
\begin{tabular}{@{}l|cccccc|cccccc|cccccc@{}}
\toprule
\multirow{3}{*}{\textbf{Method}} &
\multicolumn{6}{c|}{\textbf{Profiling}} &
\multicolumn{6}{c|}{\textbf{Factual Retention}} &
\multicolumn{6}{c}{\textbf{Overall}} \\
\cmidrule(lr){2-7} \cmidrule(lr){8-13} \cmidrule(lr){14-19}
& Acc & File F1 & File Recall & File Precision & Avg Score & Avg Latency
& Acc & File F1 & File Recall & File Precision & Avg Score & Avg Latency
& Acc & File F1 & File Recall & File Precision & Avg Score & Avg Latency \\
\midrule
\multicolumn{19}{c}{\textbf{\textit{RAG Methods}}} \\ \midrule
Standard RAG~\cite{Lewis2020RAG}
& 26.7 & 18.4 & \underline{19.8} & 20.1 & 2.9 & \cellcolor{yellow!25}\textbf{5281.6}
& 27.6 & 28.1 & \underline{75.0} & 21.8 & 3.2 & \cellcolor{yellow!25}\textbf{5473.8}
& 27.5 & 27.1 & \underline{69.3} & 21.7 & 3.2 & \cellcolor{yellow!25}\textbf{5454.0} \\
Self RAG~\cite{Asai2024SelfRAG}
& 10.0 & 15.2 & 14.9 & 20.8 & 1.7 & 69917.9
& 25.3 & 30.1 & 66.6 & 24.9 & 3.0 & 75821.5
& 23.8 & 28.5 & 61.3 & 24.5 & 2.8 & 75211.9 \\
\midrule
\multicolumn{19}{c}{\textbf{\textit{Search Agent Methods}}} \\ \midrule
ReAct~\cite{Yao2023ReAct} (Qwen3-VL-8B-Instruct~\cite{Qwen3VL2025})
& 13.5 & 11.8 & 8.2 & 27.3 & 2.2 & 388019.1
& 26.7 & \cellcolor{yellow!25}\textbf{39.7} & 54.6 & \cellcolor{yellow!25}\textbf{37.3} & 3.3 & 489949.4
& 25.3 & \cellcolor{yellow!25}\textbf{36.9} & 49.8 & \cellcolor{yellow!25}\textbf{36.2} & 3.2 & 479423.0 \\
ReAct~\cite{Yao2023ReAct} (Gemini-2.5-flash~\cite{Comanici2025Gemini25})
& 20.0 & \underline{18.5} & \cellcolor{yellow!25}\textbf{20.5} & 19.9 & 2.9 & \underline{9897.5}
& 35.1 & 24.8 & \cellcolor{yellow!25}\textbf{80.8} & 18.9 & 3.9 & 13918.2
& 33.6 & 24.1 & \cellcolor{yellow!25}\textbf{74.5} & 19.0 & 3.8 & 13503.0 \\
Search-R1~\cite{Jin2025SearchR1}
& 5.0 & 10.8 & 7.6 & 23.3 & 1.7 & 12360.3
& 24.8 & \underline{37.7} & 59.6 & 32.5 & 2.6 & \underline{11856.4}
& 22.7 & \underline{34.9} & 54.2 & 31.5 & 2.5 & \underline{11908.5} \\
\midrule
\multicolumn{19}{c}{\textbf{\textit{Autonomous Agent Systems}}} \\ \midrule
Terminal Agent (Qwen3-VL-8B-Instruct~\cite{Qwen3VL2025})
& 16.7 & 11.6 & 8.7 & 31.6 & 2.0 & 101941.8
& 11.1 & 16.4 & 20.5 & 16.9 & 1.3 & 65477.2
& 11.7 & 15.9 & 19.3 & 18.4 & 1.4 & 69242.9 \\
Terminal Agent (Gemini-2.5-flash~\cite{Comanici2025Gemini25})
& 25.0 & 15.0 & 10.7 & \underline{39.2} & 2.6 & 19353.9
& 21.7 & 25.1 & 33.9 & 24.2 & 2.4 & 34474.4
& 22.0 & 24.0 & 31.5 & 25.7 & 2.5 & 32912.9 \\
Terminal Agent (GPT-5.2~\cite{openai2025gpt5_2_system_card})
& \underline{30.0} & 11.1 & 8.3 & 32.0 & \underline{3.8} & 61514.9
& \underline{45.7} & 22.9 & 47.6 & 18.0 & \underline{5.0} & 146353.1
& \underline{44.1} & 21.7 & 43.6 & 19.5 & \underline{4.9} & 137591.8 \\
ChatGPT Agent Mode~\cite{openai2024gpt4technicalreport, openai2025gpt5}
& \cellcolor{yellow!25}\textbf{48.3} & \cellcolor{yellow!25}\textbf{21.0} & 15.8 & \cellcolor{yellow!25}\textbf{40.1} & \cellcolor{yellow!25}\textbf{5.3} & 614666.7
& \cellcolor{yellow!25}\textbf{56.8} & 30.9 & 30.3 & \underline{33.9} & \cellcolor{yellow!25}\textbf{5.9} & 805238.5
& \cellcolor{yellow!25}\textbf{55.9} & 29.9 & 28.9 & \underline{34.5} & \cellcolor{yellow!25}\textbf{5.8} & 785558.1 \\
\bottomrule
\end{tabular}
\end{adjustbox}
\end{table*}

\cref{tab:metric_summary} complements the main-text results by placing answer quality, evidence retrieval quality, and runtime efficiency in a single unified table. Unlike the profile-wise and capability-wise summaries in the main text, this appendix table is intended to expose cross-metric trade-offs more directly, especially the separation between answer correctness, evidence coverage, evidence specificity, and latency.

First, \textbf{the ranking induced by answer quality is clearly separated from that induced by retrieval quality}. On answer-level metrics, autonomous agent systems are consistently stronger than RAG-style pipelines and lightweight search agents. In particular, ChatGPT Agent Mode achieves the best accuracy across \textit{profiling} (48.3), \textit{factual retention} (56.8), and the overall benchmark average (55.9), and also obtains the highest average judge scores on all three aggregates (5.3, 5.9, and 5.8). Terminal Agent (GPT-5.2) is the second strongest method in terms of final answer quality, reaching 30.0/45.7/44.1 accuracy on profiling/factual retention/overall, respectively. By contrast, methods with relatively competitive retrieval statistics, such as Search-R1 or ReAct variants, remain substantially behind on final judged correctness. This gap indicates that, in \ourmethod{}, performance is not bottlenecked by retrieval alone; converting retrieved evidence into a semantically correct final response remains a major source of error.

Second, \textbf{strong evidence retrieval does not reliably yield strong end-task performance}. This is most visible in the search-oriented methods. ReAct (Qwen3-VL-8B-Instruct) achieves the highest overall File F1 (36.9) and File Precision (36.2), while ReAct (Gemini-2.5-flash) yields the highest overall File Recall (74.5). Search-R1 is also competitive, with 34.9 overall File F1 and 54.2 overall File Recall. However, these methods do not translate their retrieval advantages into comparable answer accuracy: their overall accuracies are 25.3, 33.6, and 22.7, respectively, all well below the best autonomous agents. This discrepancy suggests that a substantial fraction of benchmark failures arise after retrieval, including incomplete interpretation of multimodal evidence, inability to reconcile partially relevant files, and weak synthesis across multiple supporting sources. In other words, retrieving the annotated supporting files is necessary, but not sufficient, for producing a correct answer under the LLM-judge criterion.

Third, \textbf{the two task families stress different failure modes}. Across nearly all methods, factual retention is easier than profiling at the answer level. For example, Terminal Agent (GPT-5.2) improves from 30.0 accuracy on profiling to 45.7 on factual retention, and ChatGPT Agent Mode improves from 48.3 to 56.8. This pattern is consistent with the intended task design. Factual-retention questions more often depend on recovering explicit user-local facts, whereas profiling requires abstraction from weak, distributed, and often temporally separated signals. Notably, this answer-level gap is larger than the corresponding gap in retrieval quality. For several methods, profiling File F1 is not dramatically worse than factual-retention File F1, and in some cases remains comparable relative to the method family. The larger drop in profiling accuracy therefore points to a reasoning bottleneck rather than a purely retrieval bottleneck: even when candidate evidence is partially available, inferring stable user traits, preferences, or routines is substantially harder than restating localized facts.

Fourth, \textbf{RAG pipelines exhibit a characteristic coverage--specificity trade-off}. Standard RAG and Self RAG achieve relatively high overall File Recall (69.3 and 61.3), but their overall File Precision remains limited (21.7 and 24.5), and their overall accuracies remain modest (27.5 and 23.8). This pattern suggests that these methods often retrieve broad evidence pools that overlap with the annotated supporting set, but struggle to isolate the minimal evidence needed for precise grounded answering. The effect is especially pronounced on factual retention, where both methods recover a large fraction of relevant files yet still underperform substantially on judged correctness. This result is consistent with the broader design motivation of \ourmethod{}: benchmark success requires not only coarse retrieval coverage, but also disciplined evidence selection and reliable reasoning over heterogeneous personal files.

Finally, \textbf{there is a pronounced efficiency--capability trade-off}. Standard RAG is the fastest method by a large margin, with the lowest average latency across all three aggregates, while ReAct (Gemini-2.5-flash) and Search-R1 remain relatively efficient compared with full autonomous agents. In contrast, the strongest answer-level method, ChatGPT Agent Mode, incurs by far the highest latency. The same trend holds, though less extremely, for other agentic systems with stronger end-task performance. This suggests that current gains in grounded multimodal file reasoning are achieved partly through longer interaction horizons, more iterative tool use, or heavier cross-file processing, rather than through more efficient inference alone.

Overall, the results in~\cref{tab:metric_summary} reinforce the central premise of \ourmethod{}: personalized file-system QA cannot be adequately characterized by final accuracy alone. Methods that appear competitive in retrieval may still fail during evidence interpretation and synthesis, while methods that achieve stronger final correctness often do so with substantial computational overhead. The benchmark therefore exposes a three-way tension among evidence coverage, grounded reasoning, and efficiency, which is largely obscured by single-metric evaluation.

\clearpage
\phantomsection
\addcontentsline{toc}{section}{References}
\bibliographystyle{splncsnat_like}
\bibliography{main}

\end{document}